\documentclass[]{article}

\let\today\relax
\makeatletter
\def\ps@pprintTitle{%
    \let\@oddhead\@empty
    \let\@evenhead\@empty
    \def\@oddfoot{\footnotesize\itshape
         {} \hfill\today}%
    \let\@evenfoot\@oddfoot
    }
\makeatother
\usepackage{authblk}
\usepackage{amsmath,amsfonts,amssymb}
\usepackage{listings}
\usepackage{algorithm}
\usepackage{algorithmic}
\usepackage{eqparbox}

\usepackage{capt-of}
\usepackage{orcidlink}
\usepackage[T1]{fontenc}
\usepackage[latin1]{inputenc}
\usepackage{graphics}
\usepackage{multirow}
\usepackage{subcaption}
\usepackage{hyperref} 
\usepackage{hypcap} 
\usepackage{multirow}
\usepackage{multicol}
\usepackage{mathtools}
\usepackage{xcolor}
\usepackage[normalem]{ulem}
\usepackage[backend=bibtex,giveninits=true,doi=false,url=false]{biblatex}
\newcommand{\norm}[1]{\left\lVert#1\right\rVert}

\newcommand{\omegaVector}{\boldsymbol{\omega}}
\newcommand{\omegaRV}{\boldsymbol{\Omega}}
\newcommand{\xVector}{\boldsymbol{x}}
\newcommand{\xRV}{\boldsymbol{X}}

\newcommand{\ampl}{a}
\def\BFBH{\boldsymbol{a}}
\newcommand{\rwInc}{\boldsymbol{\nu}}
\newcommand{\cset}{\mathbb{C}}
\newcommand{\C}{\mathbb{C}}
\def\BARS{\mathbf{S}}
\newcommand{\nset}{\mathbb{N}}
\newcommand{\rset}{\mathbb{R}}
\def\R{\mathbb{R}}
\def\EXPECT{\mathbb{E}}

\def\DATAM{{\rho}}

\def\ID{\mathbf{I}}

\def\ALPHAEX{\gamma}             
\providecommand{\algAM}{\ensuremath{\mathrm{AM}}}
\providecommand{\algAMR}{\ensuremath{\mathrm{AMR}}}
\providecommand{\algAMRmod}{\ensuremath{\mathrm{AMRv2}}}
\providecommand{\algRWR}{\ensuremath{\mathrm{RWR}}}

\title{Adaptive Random Fourier Features Training Stabilized By Resampling With Applications in Image Regression}
\author[1]{Aku Kammonen\,\orcidlink{0000-0002-8458-0852}}
\author[2]{Anamika Pandey\,\orcidlink{0000-0001-8644-8540}}
\author[1]{Erik~von~Schwerin\,\orcidlink{0000-0002-2964-7225}}
\author[1,2,3]{Ra\'{u}l~Tempone\,\orcidlink{0000-0003-1967-4446}}
\affil[1]{Computer, Electrical and Mathematical Sciences and Engineering,
4700 King Abdullah University of Science and Technology (KAUST),
Thuwal 23955-6900, Kingdom of Saudi Arabia.}
\affil[2]{Chair of Mathematics for Uncertainty Quantification, RWTH Aachen University, 52062 Aachen, Germany.} 
\affil[3]{Alexander von Humboldt Professor in Mathematics for Uncertainty Quantification, RWTH Aachen University, 52062 Aachen, Germany.}

\newcommand{\keywords}[1]{\par\noindent\textbf{Keywords:} #1}
\newcommand{\subjclass}[2][2020]{\par\noindent\textbf{Mathematics Subject Classification #1:} #2}
\date{\today}

\begin{document}

\maketitle
\begin{abstract}
This paper presents an enhanced adaptive random Fourier features (ARFF) training algorithm for shallow neural networks, 
building upon the work introduced in 
"Adaptive Random Fourier Features with Metropolis Sampling", Kammonen et al., \emph{Foundations of Data Science}, 2(3):309--332, 2020.
This improved method uses a particle filter-type resampling technique to stabilize the training process and reduce the sensitivity to parameter choices. 
The Metropolis test can also be omitted when resampling is used, reducing the number of hyperparameters by one and reducing the computational cost per iteration compared to the ARFF method.
We present comprehensive numerical experiments demonstrating the efficacy of the proposed algorithm in function regression tasks 
as a stand-alone method and as a pretraining step before gradient-based optimization, using the Adam optimizer. 
Furthermore, we apply the proposed algorithm to a simple image regression problem, illustrating its utility in sampling frequencies for the 
random Fourier features (RFF) layer of coordinate-based multilayer perceptrons. 
In this context, we use the proposed algorithm to sample the parameters of the RFF layer in an automated manner.
\end{abstract}
\subjclass[2020]{68T05, 65D15, 68T20, 65D40, 65C05}
\keywords{Adaptive Random Fourier Features, Shallow Neural Networks, Multilayer Perceptron, Particle Filter Resampling, Function Regression, Image Regression}
\section{Introduction}
\label{sec:Intro}
More than a decade ago, randomized feature weighting methods were proposed in the machine learning community to accelerate kernel machine training. 
The central idea is based on combining the robustness of kernel machines with the scalability of neural network structures. 
Rahimi and Recht introduced random Fourier features~\cite{RR2007RandomFF} as a computationally fast and accurate method to approximate the kernels of support vector machines~\cite{Vladimir1998, SCSpringer2008}. 
They presented a pointwise convergence proof and offered empirical evidence of the competitiveness of random Fourier features (RFF) considering accuracy, training, and evaluation time compared to state-of-the-art kernel machines~\cite{JohnNIPS1998SVM}.
In follow-up work~\cite{NIPS2008RahimiRecht}, they introduced randomization in learning feature weights instead of minimization for shallow neural networks. 
Training algorithms based on randomizing the feature weights have been resurfacing in the machine learning community, 
primarily because randomization is computationally cheaper than optimization in many cases. 
Many interesting studies have been published that highlight the idea of randomizing feature weights instead of training them, demonstrating  
numerical success~\cite{Tancik2020FourierFL, lisi_paper, huang2024convergenceratesrandomfeature} and providing 
theoretical understanding~\cite{Yehudai2019OnTP, Sonoda2020OnTA, Li2017InsightsIR, huang2024convergenceratesrandomfeature, BarronSpaceWeinan2022}.

The core idea in the mentioned literature is to assign random weights, which are kept fixed during training, for the hidden layer and only optimize the weights of the output layer or of the nonlinearities. 
In~\cite{kammonen2020adaptive}, the authors presented an adaptive approach to sampling the random weights for the hidden layer based on a Metropolis-like sampling algorithm, stated there as 
Algorithm~1 "Adaptive random Fourier features with Metropolis sampling," herein referred to as ARFF. 
They also derived an error bound for two-layer (one hidden layer) neural networks following the well-cited results for sigmoidal functions by Barron~\cite{Barron_1993}. 
Further, in~\cite{kammonen2020adaptive}, the optimal distribution of the random weights, in the sense of minimizing the constant in the error bound, was derived for a trigonometric activation function. 
Following~\cite{kammonen2020adaptive}, the two-layer neural networks treated in the present work are
\begin{equation*}
    \beta(\xVector) = \sum_{k=1}^K \ampl_k \exp{(\mathrm{i}\omegaVector_k^T \xVector)}
\end{equation*}
where $\ampl_k\in \mathbb{C}$ and $\xVector, \omegaVector_{k} \in \mathbb{R}^d$ are column vectors. The weights $\ampl_k$ are amplitudes, and the weights $\omegaVector_{k}$ are frequencies.
In addition, ARFF has successfully been used for training in different settings~\cite{kiessling2021, huang2024convergenceratesrandomfeature, lisi_paper, kammonen2022deepVSshallow}. 
For instance, an experimental comparative study between stochastic gradient descent and the ARFF algorithm in the context of spectral bias~\cite{kiessling2022computable} was conducted in~\cite{lisi_paper}.
The theoretical analysis in~\cite{kammonen2022deepVSshallow} provided quantitative results of when deep residual neural networks 
can give smaller error than shallow neural networks. The same paper also describes a method to pretrain deep residual neural networks 
layer-by-layer with ARFF, which could also be applied using the algorithm proposed in this paper.

This paper proposes a novel approach to sample the frequencies in neural networks optimally using the resampling technique widely used in particle filtering. 
Gordon et al.~\cite{Gordon1993} first introduced resampling, a crucial step in particle filters, to address the problem of weight degeneracy.
Since its introduction, resampling has become an integral component of particle filtering, with numerous developments and variations proposed in the literature~\cite{Beadle1997, Tiancheng2015, LiuAndChen1998}.

The primary objective of resampling is to mitigate the scenario in which a small subset of particles carries significant weight values while the majority have negligible weight values. 
This phenomenon, known as weight degeneracy, compromises the ability of the particle filter to represent the posterior distribution accurately. Drawing parallels to this concept, 
with neurons corresponding to particles, we explore the application of resampling to sample the feature weights in neural networks.
Specifically, we investigate the resampling of feature weights based on the modulus of their corresponding amplitudes $|\ampl_k|$.
To the best of our knowledge, this marks the first application of resampling techniques to features sampling in neural networks. 
We investigate the effect of resampling in ARFF and also in the context of a simple random walk method. 

In~\cite{UnifiedAnalysisRFF2021}, the authors proposed a data-dependent approach for sampling feature weights 
(corresponding to the number of nodes in the hidden layer of a neural network) based on ridge leverage scores. 
Their method controls the effective number of feature weights to use in RFF. 
In contrast, in this paper, resampling is introduced to improve the properties of the iterative ARFF method, while keeping 
the number of feature weights fixed.

We summarize the proposed RFF algorithm with and without resampling and with and without a Metropolis test, where ARFF is a particular case, in Algorithm~\ref{alg:AMRS_E}. 
The resampling step is executed when the effective sample size (ESS) falls below a predetermined fraction of the total number of samples, corresponding to the width of the neural network. 
Including the resampling step in the training algorithm offers the following three significant benefits: 
\begin{enumerate}
    \item Resampling increases the training stability and makes the algorithm less sensitive to parameter choices.
    \item Resampling tends to reduce errors faster in the first iterations, which is of particular importance when the algorithm is employed as a pretrainer for a gradient-based minimization method.
    \item The proposed algorithm can be run without the Metropolis step if resampling is employed, reducing the work per iteration of ARFF and removing one of the hyperparameters ($\gamma$).
\end{enumerate}

We first illustrate these benefits on a manufactured function approximation problem. 
Then, building upon the work of Lazzari et al.~\cite{lazzari2023understanding} and Tancik et al.~\cite{Tancik2020FourierFL}, 
we demonstrate the application of our proposed algorithm for adaptive sampling of random weights. 
Lazzari et al.~\cite{lazzari2023understanding} presented a study on the spectral bias of rectified linear unit (ReLU) multilayer perceptrons (MLPs) in coordinate-based image regression, 
whereas Tancik et al.~\cite{Tancik2020FourierFL} demonstrated how RFF can enhance the ability of MLPs to learn high-frequency image content. 
In some experiments, Tancik et al.~\cite{Tancik2020FourierFL} sampled the weights of the RFF layer from a normal distribution with a manually tuned standard deviation. 
The proposed approach employs the proposed algorithm with a resampling strategy to sample these weights adaptively. We applied this method to a dataset of 92 images, observing promising results.

The contributions of this work are summarized as follows:
\begin{itemize}
    \item We present an improved ARFF training algorithm based on resampling that stabilizes the training process and reduces the sensitivity of the parameter choices. Furthermore, the algorithm can be run without the Metropolis test used in ARFF, reducing the number of hyperparameters by one.
    \item We also present numerical experiments where the proposed algorithm is used alone and for pretraining before training with the Adam optimizer in the context of function regression.
    \item Moreover, we demonstrate how the proposed algorithm can be employed to sample frequencies for the RFF layer of coordinate-based MLPs in image regression, as discussed in~\cite{Tancik2020FourierFL}.
\end{itemize}
\section{Adaptive Random Features With Resampling}

First, Section~\ref{sec:ARFF} recalls concepts related to the ARFF algorithm.
Then, Section~\ref{sec:resampling} describes the modified algorithm.

\subsection{Problem Statement and Adaptive Random Fourier Features}
\label{sec:ARFF}

We have a data-set $\left(\xRV_{m}, Y_{m}\right) \in \R^d \times \R$, for $m = 1,\dots,M$, 
and assume that $\xRV_m$ are independent and identically distributed (i.i.d.) random variables from an unknown distribution
with density $\rho$ and that $Y_m=f\left(\xRV_m\right)$ for some unknown function $f:\R^d \rightarrow \R$.
We may also consider 
noisy data, see~\cite{kammonen2020adaptive}, but for simplicity, this paper assumes exact function evaluations.

The aim is to approximate the target function $f$ using a two-layer neural network
with a trigonometric activation function. The neural network is defined as the parameterized function 
$\beta_{\boldsymbol{\theta}}: \R^d \rightarrow \C$, given by
\begin{equation}
    \label{eq:NN-Def}
    \beta_{\boldsymbol{\theta}}(\xVector) = 
    \sum_{k = 1}^{K} \ampl_{k}\exp(\mathrm{i}\omegaVector_{k}^T \xVector).
\end{equation}
The parameters are the weights, $\ampl_{k}\in\C$ and $\omegaVector_{k}\in\R^d$ for $k=1,\dots,K$, that we call amplitudes and frequencies, respectively. 
We let 
$\boldsymbol{\theta}\in\boldsymbol{\Theta}=\boldsymbol{\Theta}_{\ampl} \times\boldsymbol{\Theta}_{\boldsymbol{\omegaVector}}\subseteq\C^K \times \R^{d\times K}$
denote all parameters.
To simplify the notation, we sometimes use $\beta(\xVector)$ to denote the neural network model, omitting the parameter dependence.

Using a quadratic loss function and introducing a penalty on the amplitudes 
leads to a regularized least--square minimization problem

\begin{equation}
    \label{eq:NonConvexOptProblem}
    \min_{\boldsymbol{\theta} \in \boldsymbol{\Theta}} 
    \left\{ \EXPECT_{\rho}\left[(Y - \beta_{\boldsymbol{\theta}}(\xRV))^{2}\right] 
    + \lambda \|\BFBH\|_2^{2} \right\},
\end{equation}
for some $\lambda\geq 0$, where $\BFBH = (\ampl_{1}, \ampl_{2}, \ldots, \ampl_{K})^{T}$ and $\|\BFBH\|_2^{2}=\sum_{k=1}^K{|\ampl_{k}|^{2}}$. 
The subscript $\rho$ in $\EXPECT_{\rho}$ indicates that we take the expectation w.r.t. $\xRV~\sim \rho$, with $Y=f(\xRV)$.
The data density function $\rho$ is unknown. Therefore, in practice, the expected value in~\eqref{eq:NonConvexOptProblem}
is approximated using an average based on the existing data.

This study follows~\cite{kammonen2020adaptive} and treats the frequencies $\left\{\omegaVector_k\right\}_{k=1}^K$
as i.i.d. random variables, $\left\{\omegaRV_k\right\}_{k=1}^K$, with the associated probability density function $p$.
For the given frequencies $\omegaVector_k$, this approach results in a least--square problem for linear equations in the amplitudes 
$\left\{\ampl_k\right\}_{k=1}^K$.
As shown in~\cite{kammonen2020adaptive},
\begin{align}
    \label{eq:one_over_K_bound}
    \min_{\boldsymbol{\theta}\in\boldsymbol{\Theta}} \left\{ \EXPECT_{\rho}[(Y - \beta_{\boldsymbol{\theta}}(\xRV))^{2}] + \lambda \|\BFBH\|_2^{2}\right\} 
    & \leq \frac{1+\lambda}{K}\EXPECT_{p}\left[\frac{|\hat{f}(\omegaRV)|^2}{(2\pi)^dp(\omegaRV)^2}\right],
\end{align}
where $\hat{f}(\omegaVector) = (2\pi)^{-d/2}\int_{\R^d}f(\xVector)\exp(-\mathrm{i}\omegaVector^T \xVector)\mathrm{d}\xVector$ is the Fourier transform of $f$. 
On the right-hand side of~\eqref{eq:one_over_K_bound}, $\EXPECT_{p}$ denotes the expectation
w.r.t. $\omegaRV~\sim p$. 
In~\cite{kammonen2020adaptive}, it is also shown, under some assumptions and following a classical result in importance sampling, 
that the constant of proportionality $\EXPECT_{p}\left[\frac{|\hat{f}(\omegaRV)|^2}{(2\pi)^dp(\omegaRV)^2}\right]$ 
in~\eqref{eq:one_over_K_bound} is minimized by
\begin{equation*}
    p(\omegaVector) = p_{*}(\omegaVector) = \frac{|\hat{f}(\omegaVector)|}{\int_{\R^{d}} |\hat{f} (\omegaVector^{'})|\mathrm{d}\omegaVector'}.
\end{equation*}
Using the minimizer $p_{*}$ in~\eqref{eq:one_over_K_bound} yields
\begin{align}
	\label{eq:error_bound}
    \min_{\boldsymbol{\theta}\in\boldsymbol{\Theta}}
    \left\{\EXPECT_\DATAM\left[|Y-\beta(\xRV)|^2\right]+\lambda\|\BFBH\|^2\right\}
	& \leq
	\frac{1+\lambda}{(2\pi)^{d}K}\left\|\hat{f}\right\|_{L^1\left(\mathbb{R}^d\right)}^2.
\end{align}
An estimate of the approximation error that considers the finite data-set requires an
additional error term that goes to zero as $M\to\infty$ (see Theorem~2.1 in~\cite{huang2024convergenceratesrandomfeature}), where such an error bound
is stated for a related problem.

The ARFF algorithm is an iterative process that aims to asymptotically sample from $p_*$. 
Given the iterand $\left[\omegaVector_1,\dots,\omegaVector_K\right]$, each iteration involves selecting a batch, $D_{M_B}$, 
consisting of $M_B$ distinct\footnote{That is, selecting $M_B$ out of $M$ indices $\left\{m_j\right\}_{j=1}^{M_B}\subset\{1,\dots,M\}$ such that $j\neq l\implies m_j\neq m_l$.} 
points $\left\{\left(\xRV_{m_j},Y_{m_j}\right)\right\}_{j=1}^{M_B}$, with $K< M_B\leq M$, from the data-set, and solving 
\begin{equation}\label{eq:num_normal_eq}
    (\BARS^H\BARS+\lambda M_B \ID)\BFBH = \BARS^H {\mathbf{y}}
\end{equation}
for $\BFBH$, where $\BARS\in\mathbb{C}^{M_B\times K}$ represents the matrix with elements $\BARS_{j,k} = \exp(\mathrm{i}\omegaVector_{k}^{T} \xVector_{m_j})$, 
$\ID$ denotes an identity matrix, 
${\mathbf{y}}$ denotes the column vector with elements ${\mathbf{y}}_j={Y_{m_j}}$, and 
the superscript $H$ denotes the Hermitian transpose. 
The idea behind ARFF is that the normalized modulus of the coefficients, $\vert \ampl_k \vert/\norm{\BFBH}_1$, in a certain sense, 
approximates the optimal density $p_*\left(\omegaVector_{k}\right)$. Based on this idea, a modified Metropolis sampling
step is used to update the frequencies. 
In the ARFF proposal step, $\omegaVector'_{k}=\omegaVector_{k}+\delta \rwInc_k$ are sampled with $\rwInc_k$ 
independent $d$-dimensional standard Gaussian random variables, with $\delta>0$.
The proposal $\omegaVector'_{k}$ is accepted with probability 
$\min{\left(1, \frac{\vert \ampl'_k\vert^\gamma}{\vert a_k\vert^\gamma}\right)}$, 
where the amplitude $\ampl'_k$ is obtained by solving~\eqref{eq:num_normal_eq} with the proposal frequencies, and 
$\gamma$ represents an algorithm parameter tuned to the problem.

The ARFF algorithm, as presented in~\cite{kammonen2020adaptive}, also has the update frequency of $\BFBH$ as a parameter, 
unlike Algorithm~\ref{alg:AMRS_E}, where $\BFBH$ is updated in every iteration. With that exception, 
ARFF is included in Algorithm~\ref{alg:AMRS_E} as a special case where 
$R(n)\equiv 0$, $A(n)\equiv\mathbf{true}$, and $M_B \equiv M$.

\subsection{Training Algorithm With Resampling}
\label{sec:resampling}

In particle filters for partially observed diffusions, the filtering probability distribution is approximated 
by a collection of particles associated with weights updated with successive observations. 
There, the issue of degeneracy when the probability mass concentrates on a small number of particles is 
counteracted by resampling (e.g., see the tutorial~\cite{PF_Smoothing_tutorial}).

In ARFF, the relative accuracy of $\vert \ampl_k \vert/\norm{\BFBH}_1$ as an approximation of $p_*\left(\omegaVector_{k}\right)$
is poor when $p_*\left(\omegaVector_{k}\right)$ is small enough for a given $K$, reducing the effectiveness of ARFF, 
motivating the enhancement of ARFF by including occasional resamplings. 
Hence, we extended the analogy between $\vert \ampl_k \vert/\norm{\BFBH}_1$ and a probability mass function and applied the concept of 
ESS from particle filters and related methods:
\begin{align}
	\label{eq:ESS}
	K_\mathrm{ESS} & := \frac{1}{\sum_{k=1}^K \check{p}_k^2}, 
\end{align}
where $\mathbf{\check{p}}:=\left\vert\BFBH\right\vert/\norm{\BFBH}_1$. 
Then, $1\leq K_\mathrm{ESS}\leq K$ where the two extremes correspond to only one non-zero 
amplitude, and $\vert \ampl_k \vert=\vert \ampl_j \vert$ for all $j$ and $k$.
Interpreting a low value of $K_\mathrm{ESS}$ as an indication of many wasted degrees of freedom in the neural network, we
introduced a threshold,\footnote{This threshold may dependent on the iteration (e.g. $R=1$ for the 
the first iterations, and later $R<1$).} $R\in[0,1]$, monitored $K_\mathrm{ESS}$, and resampled the 
frequencies $\left\{\omegaVector_k\right\}_{k=1}^K$ according to the probability mass function $\mathbf{\check{p}}$
when $K_\mathrm{ESS}\leq RK$. This enhanced version of ARFF is described in Algorithm~\ref{alg:AMRS_E}, 
which takes the initial frequencies and problem data as inputs.

The statement of Algorithm~\ref{alg:AMRS_E} also allows for bypassing the Metropolis-type test 
completely, with each iteration only combining a random walk step with a solution to~\eqref{eq:num_normal_eq}
and a resampling step. With that approach $R\equiv 1$ is natural, 
and intuitively it may work because, for very large $K$, each iteration can be viewed as approximately 
drawing i.i.d. samples from a distribution that is a convolution between the (unknown) target density $p_*$
and the distribution of a centered $d$-dimensional Gaussian with variance $\delta^2\ID$.

Constructing and solving the least--square systems dominates the computational cost 
of Algorithm~\ref{alg:AMRS_E}, performed as follows:
\begin{itemize}
\item once per iteration for random walk with resampling in each iteration,
\item twice per iteration for Adaptive Metropolis without resampling, and 
\item two or three times per iteration (depending on $K_\mathrm{ESS}$ and $R(n)$) for Adaptive Metropolis with resampling.
\end{itemize}

\begin{algorithm}[h!]
  \caption{Adaptive RFF with Metropolis sampling and resampling}
  \label{alg:AMRS_E}
  \begin{algorithmic}
    \STATE {\bfseries Input:} $\{(\xVector_m, y_m)\}_{m=1}^M$, $\left\{\omegaVector_k\right\}_{k=1}^K$
    \STATE {\bfseries Output:} $\xVector\mapsto\sum_{k=1}^K a_k \exp{\left(\mathrm{i}\omegaVector_k^T \xVector \right)}$, \hspace{\algorithmicindent} $\ampl_k\in\cset$, $\omegaVector_k\in\rset^d$
    \STATE {\bfseries Algorithm Parameters:} 
      \\ \hspace{\algorithmicindent} $N$ \hspace{\algorithmicindent}(number of iterations),
      \\ \hspace{\algorithmicindent} $\delta>0$ \hspace{\algorithmicindent}(standard deviation of the proposal step),
      \\ \hspace{\algorithmicindent} $\gamma\geq 1$ \hspace{\algorithmicindent}(exponent in the acceptance condition),
      \\ \hspace{\algorithmicindent} $K< M_B\leq M$ \hspace{\algorithmicindent}(batch size),
      \\ \hspace{\algorithmicindent} $\lambda>0$ \hspace{\algorithmicindent}(Tikhonov parameter),
      \\ \hspace{\algorithmicindent} $R$ \hspace{\algorithmicindent}(resampling rule, $R:\nset\mapsto[0,1]$; e.g. a constant $R\in[0,1]$),
      \\ \hspace{\algorithmicindent} $A$ \hspace{\algorithmicindent}(Adaptive-Metropolis rule $A:\nset\mapsto\{\TRUE,\FALSE\}$)
    \STATE~ 
    \STATE $D_{M_B} \gets  \text{batch of $M_B$ data points from $\{(\xVector_m, y_m)\}_{m=1}^M$}$
    \STATE $\BFBH \gets \text{solution to~\eqref{eq:num_normal_eq} given $\left\{\omegaVector_k\right\}_{k=1}^K$ and $D_{M_B}$}$
    \FOR{$n = 1$ {\bfseries to} $N$}
      \STATE $\overline{\ampl} \gets \sum_{k=1}^K\left\vert \ampl_k\right\vert$
      \STATE $\mathbf{\check{p}} \gets \left\vert\BFBH\right\vert / \overline{\ampl}$
      \STATE $D_{M_B} \gets  \text{batch of $M_B$ data points from $\{(\xVector_m, y_m)\}_{m=1}^M$}$
      \STATE $K_\mathrm{ESS} \gets \left(\sum_{k=1}^K \check{p}_k^2\right)^{-1}$
      \IF[Resampling]{$K_\mathrm{ESS}\leq R(n)K$}
        \STATE $\mathbf{j} \gets \text{vector of $K$ independent samples from $1,\dots,K$ with PMF $\mathbf{\check{p}}$}$
        \STATE $\omegaVector_{1,\dots,K} \gets \omegaVector_\mathbf{j}$
        \IF{A(n)}
            \STATE $\BFBH \gets \text{solution to~\eqref{eq:num_normal_eq} given $\left\{\omegaVector_k\right\}_{k=1}^K$ and $D_{M_B}$}$
        \ENDIF
      \ENDIF
      \STATE $\rwInc_{1,\dots,K} \gets \text{$K$ independent $d$-dim. Standard Multivariate Normals}$
      \IF[Adaptive Metropolis]{A(n)}
        \STATE $\omegaVector'_{1,\dots,K} \gets \omegaVector_{1,\dots,K} + \delta \rwInc_{1,\dots,K}$
        \STATE $\BFBH' \gets \text{solution to~\eqref{eq:num_normal_eq} given $\left\{\omegaVector'_k\right\}_{k=1}^K$ and $D_{M_B}$}$
        \FOR{$k = 1$ {\bfseries to} $K$}
          \STATE  $r_{\mathcal{U}} \gets \text{sample from $U[0,1]$ independent from all other r.v.}$
          \IF {$|\ampl'_k|^\gamma/|\ampl_k|^\gamma>r_{\mathcal{U}}$}
            \STATE $\omegaVector_{k} \gets \omegaVector'_k$
          \ENDIF
        \ENDFOR
      \ELSE[Random walk]
        \STATE $\omegaVector_{1,\dots,K} \gets \omegaVector_{1,\dots,K} + \delta \rwInc_{1,\dots,K}$
      \ENDIF
      \STATE $\BFBH \gets \text{solution to~\eqref{eq:num_normal_eq} given $\left\{\omegaVector_k\right\}_{k=1}^K$ and $D_{M_B}$}$
    \ENDFOR
  \end{algorithmic}
\end{algorithm}
\section{Numerical Experiments}
\label{sec:num}

\newcounter{numtestcounter}
\DeclareRobustCommand{\testnr}[1]{%
   \refstepcounter{numtestcounter}%
   \thenumtestcounter\label{#1}}

This section contains two numerical examples: one manufactured function approximation problem in Section~\ref{sec:numex_reg_disc} 
and an image regression problem in Section~\ref{sec:num_exp_image}.
For the function approximation problem, where neural networks are trained to approximate a function $f:\R^{d} \rightarrow \R$, 
we tested Algorithm~\ref{alg:AMRS_E} as a training algorithm in its own right (Section~\ref{sec:num_exp_stand_alone}) 
and as a method for pretraining before applying a gradient-based algorithm (Section~\ref{sec:num_exp_pre_training}).
In 
the second numerical experiment, we considered the problem of image regression. 
This test is based on an example from~\cite{Tancik2020FourierFL}, which we modified by changing the method 
to sample the frequencies in an RFF layer that is part of an MLP.

\subsection{Regularized Discontinuity in Four Dimensions}
\label{sec:numex_reg_disc}

The problem concerns approximating a function $f:\rset^{d} \rightarrow \rset$ based on noiseless evaluations, 
where $f$ denotes a regularized discontinuity (also used in~\cite{kammonen2020adaptive,kammonen2022deepVSshallow}), 
defined in terms of the sine integral, $\mathrm{Si}(x) = \int_0^x \frac{\sin{(t)}}{t}dt$, as
\begin{subequations}
	\label{eq:reg_disc_data_set}
	\begin{align}
		\xRV_m & \sim N\left(\mathbf{0},\ID\right),\quad\text{independent, $m=1,\dots,M$,}  \\
		Y_m & = f\left(\xRV_m\right), \\
        \label{eq:prop_Sinint_target}
		f\left(\xVector\right) & = \mathrm{Si}\left(\frac{[B^{-1}\xVector]_1}{\alpha}\right)\exp{\left(-\frac{\|B^{-1}\xVector\|_2^2}{2}\right)},
	\end{align}
\end{subequations}
for a rotation matrix $B$. 
The notation $[B^{-1}\xVector]_1$ denotes the first component of the vector $B^{-1}\xVector$.
In all tests, $d=4$ and $\alpha=0.01$.

The primary motivation for choosing the target function~\eqref{eq:prop_Sinint_target}
(as in~\cite{kammonen2020adaptive}) is 
to test the capability of the training algorithm to sample frequencies from a difficult distribution. 
In this manufactured test problem, the Fourier transform of $f$ decays slowly along the direction influenced by $\mathrm{Si}$ up to $\norm{\omegaVector}_2 \approx 1/\alpha$, but decays very quickly after that. 
Hence, the high-frequency content is relatively significant and is known. 
This Fourier transform can be approximated numerically, allowing for explicit comparisons 
between the actual errors and the bound~\eqref{eq:error_bound} 
for the best error in the case of noiseless function evaluations with unlimited data.

\subsubsection{Algorithm~\ref{alg:AMRS_E} as a Stand-alone Training Algorithm}
\label{sec:num_exp_stand_alone}

We ran Algorithm~\ref{alg:AMRS_E} for a fixed number of iterations, $N=10^4$, and 
compared adaptive Metropolis \emph{without} resampling (\algAM) to adaptive Metropolis \emph{with} resampling (\algAMR) and
random walk with resampling (\algRWR). 
The choices of $R(n)$ and $A(n)$ in Algorithm~\ref{alg:AMRS_E}
control which alternative is used (Table~\ref{tab:Alg_defs_1}). 
\begin{table}[ht]
	\centering
	\begin{tabular}{|c|l|c|c|}
	\hline
	& & $R(n)$ & $A(n)$ \\ 
	\hline
	\algAM & Adaptive Metropolis & 0 & \textbf{true} \\ 
    \algAMR & Adaptive Metropolis with resampling & 0.75 & \textbf{true} \\ 
    \algRWR & Random walk with resampling & 1 & \textbf{false} \\ 
	\hline
	\end{tabular}
	\caption{Flow control parameters in Algorithm~\ref{alg:AMRS_E} defining the three versions, denoted by \algAM, \algAMR, and \algRWR, which are compared in Tests~\ref{test:statistics},~\ref{test:all_data},~\ref{test:effect_gamma},~\ref{test:effect_batch}, and~\ref{test:effect_init}.}
	\label{tab:Alg_defs_1}
\end{table}

The choice of $R(n)\equiv0.75$ in \algAMR~was not tuned 
to obtain optimal results in this problem. However, the limited tests performed with $R(n)\equiv1$ for adaptive Metropolis with 
resampling, using the complex exponential activation function, suggest that the algorithm behaves very 
similarly to \algRWR, but at a cost that is almost three times higher (due to the extra least--square solutions in every iteration).

In the experiments in Section~\ref{sec:num_exp_stand_alone}, the rotation matrix in~\eqref{eq:prop_Sinint_target} was
\begin{align}
	\label{eq:rot_mat}
	B & = 
	\begin{bmatrix} 
		0.8617 &   0.4975 &  -0.0998 &  -0.0000 \\
    0.3028 &  -0.5246 &  -0.0000 &   0.7957 \\
    0.0865 &   0.0499 &   0.9950 &   0.0000 \\
    0.3978 &  -0.6891 &  -0.0000 &  -0.6057
	\end{bmatrix}.
\end{align}
Algorithm~\ref{alg:AMRS_E} was employed on pre-processed data-sets $\left\{\left(\xVector_m,y_m\right)\right\}_{m=1}^M$, 
where $\{\xVector_m\}_{m=1}^M$ was obtained from the samples $\{\xRV_m\}_{m=1}^M$ by centering and normalizing (i.e., 
subtracting the mean and dividing by the standard deviations componentwise), and $\{y_m\}_{m=1}^M$ was obtained by normalizing $\{Y_m\}_{m=1}^M$.

All experiments in Tests~\ref{test:all_data},~\ref{test:effect_gamma},~\ref{test:effect_batch}, and~\ref{test:effect_init}
were performed by running Algorithm~\ref{alg:AMRS_E} once per $K$-value. 
This is to say that the figures depict single outcomes of a stochastic algorithm. 
However, the illustrated features are consistent over the various tests, which are independent outcomes. 
Test~\ref{test:statistics} shows results based on 100 independent 
realizations of Algorithm~\ref{alg:AMRS_E} for each $K$-value to illustrate the random effects.
When comparisons are made in Section~\ref{sec:num_exp_stand_alone} between the three versions of Algorithm~\ref{alg:AMRS_E} 
(i.e., \algAM, \algAMR, and \algRWR), the same outcome of the increments
$\rwInc_{1,\dots,K}$ was used in all 
three versions, and all three versions shared training and testing data. 

Test~\ref{test:statistics} is included to illustrate how representative the single 
outcomes are in the following tests.
Test~\ref{test:all_data} presents the performance of \algAM, \algAMR, and \algRWR~in an idealized setting, where 
the entire data-set is used in every iteration. Test~\ref{test:all_data} also serves as a reference for the following tests. 
In addition,
\begin{itemize}
    \item Test~\ref{test:effect_gamma} illustrates that resampling reduces the sensitivity w.r.t. the parameter $\ALPHAEX$ in \algAM~and \algAMR,
    \item Test~\ref{test:effect_batch} illustrates that resampling reduces the sensitivity w.r.t. the batch size, $M_B$, and
    \item Test~\ref{test:effect_init} illustrates that resampling reduces the sensitivity of \algAM~w.r.t. the initial distribution.
\end{itemize}

\paragraph{\textbf{Test~\testnr{test:statistics}:} Statistics over 100 independent realizations}
This study used the parameter values in Tables~\ref{tab:Alg_defs_1} and~\ref{tab:Sigint_statistics} 
and was based on 100 independent realizations of the stochastic algorithms for each $K$, 
with initial frequencies $\omegaVector_k^0\equiv\boldsymbol{0}$.
As in Test~\ref{test:all_data} to~\ref{test:effect_init}, the training efficiency was measured
in the decay of $\|\mathbf{r}\|_2^2/M_B$, where
\begin{align}\label{eq:residual_normal_eq}
    \mathbf{r} & = (\BARS^H\BARS)\BFBH - \BARS^H {\mathbf{y}},
\end{align}
that is, the residual of the nonregularized ($\lambda=0$) version of~\eqref{eq:num_normal_eq}. 
The batch size $M_B\leq M$ is the number of data points included in~\eqref{eq:num_normal_eq} 
and~\eqref{eq:residual_normal_eq} in each iteration.
In Test~\ref{test:statistics}, $M_B=\left\lceil M^{3/4}\right\rceil$.
Throughout Tests~\ref{test:statistics} to~\ref{test:effect_init}, the total number of iterations was 
fixed at $10^4$ instead of stopping the algorithm adaptively. 
The corresponding $\|\mathbf{r}\|_2^2/M_T$ values, with~\eqref{eq:residual_normal_eq} based on $M_T=10^3$ data points, independent 
of those used in training, are presented as test errors.

\begin{table}[ht]
	\centering
	\begin{tabular}{|c|c|c|c|c|c|c|c|c|}
	\hline
	Problem & $K$ & $M$ & $N$ & $\delta$ & $\gamma$ & $M_B$ & $\lambda$ & $\omegaVector^0$ \\ 
	\hline
	\multirow{5}{*}{\eqref{eq:NN-Def},\eqref{eq:NonConvexOptProblem},\eqref{eq:reg_disc_data_set},\eqref{eq:rot_mat}} & 
	32 & 
	\multirow{5}{*}{$K^2$} &
	\multirow{5}{*}{$10^4$} &
	$2^{-0.75}$ & 
	\multirow{5}{*}{$10$} &
	\multirow{5}{*}{$K^{3/2}$} &
	\multirow{5}{*}{$0.1$} &
	\multirow{5}{*}{$\mathbf{0}$} \\
	 & $64$ & & & $2^{-1}$ & & & & \\ 
	 & $128$ & & & $2^{-1.25}$ & & & & \\ 
	 & $256$ & & & $2^{-1.5}$ & & & & \\ 
	 & $512$ & & & $2^{-1.75}$ & & & & \\ 
	\hline
	\end{tabular}
	\caption{Algorithm parameters corresponding to Algorithm~\ref{alg:AMRS_E} as used in Test~\ref{test:statistics} (see Figure~\ref{fig:Sigint_statistics}).}
	\label{tab:Sigint_statistics}
\end{table}

\begin{figure}[ht]
    \begin{center}
	   \includegraphics[width=0.48\linewidth]{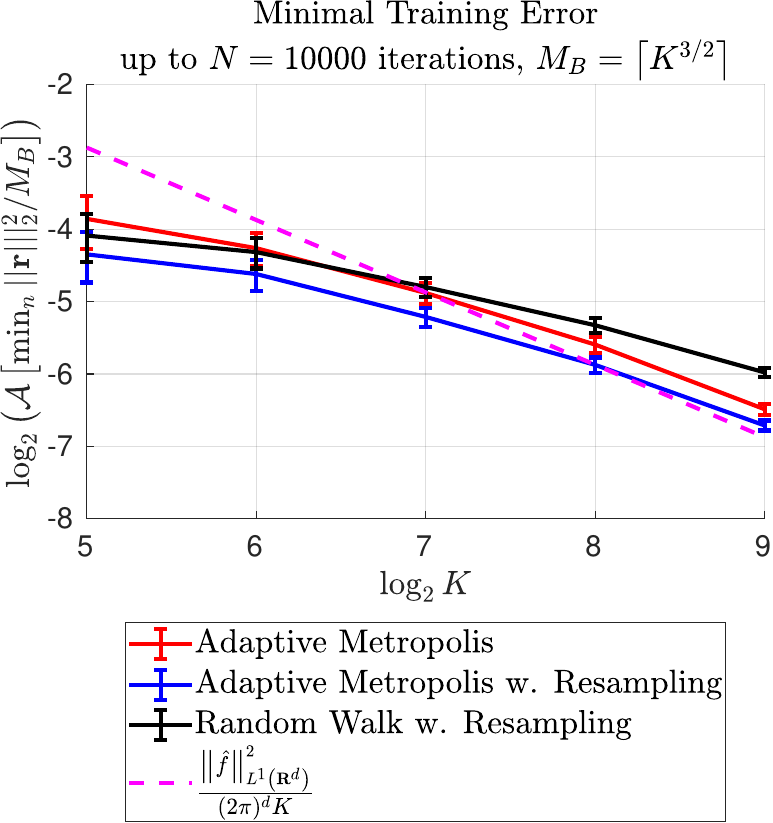}
	   \hspace{0.02\linewidth}
	   \includegraphics[width=0.48\linewidth]{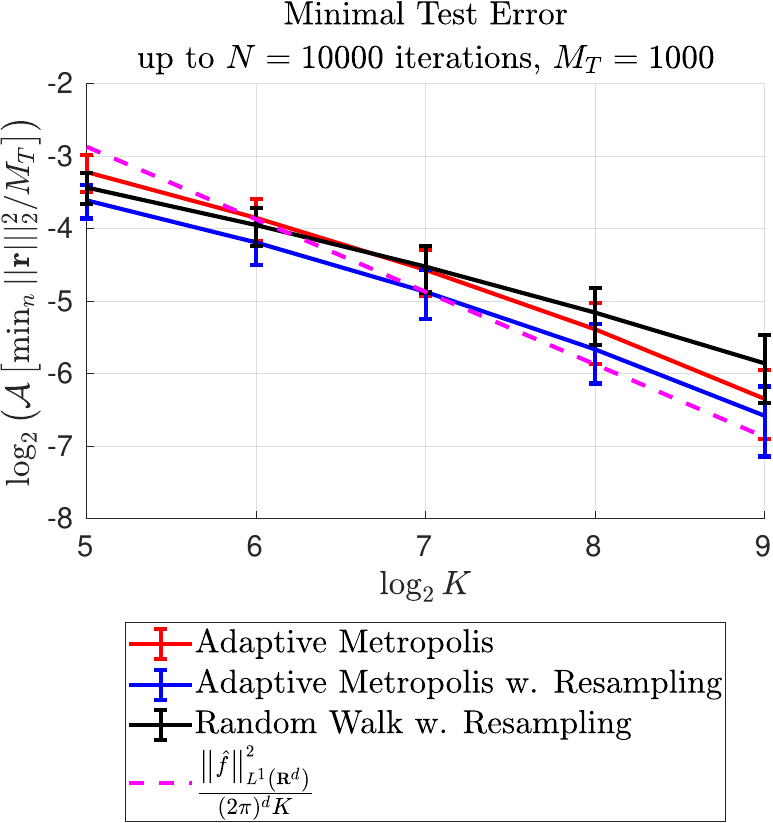}
    \end{center}

	\vspace{5mm}
    \begin{center}
	   \includegraphics[width=0.48\linewidth]{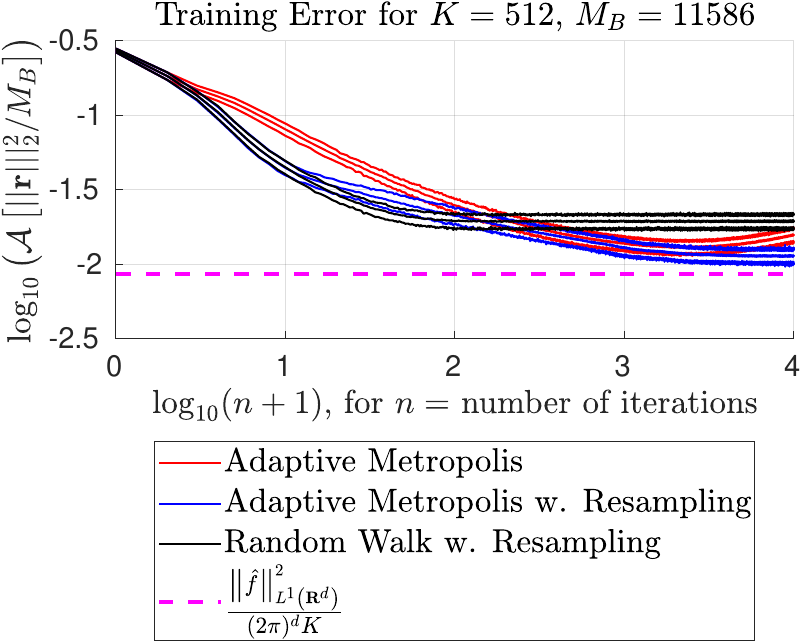}
	   \hspace{0.02\linewidth}
	   \includegraphics[width=0.48\linewidth]{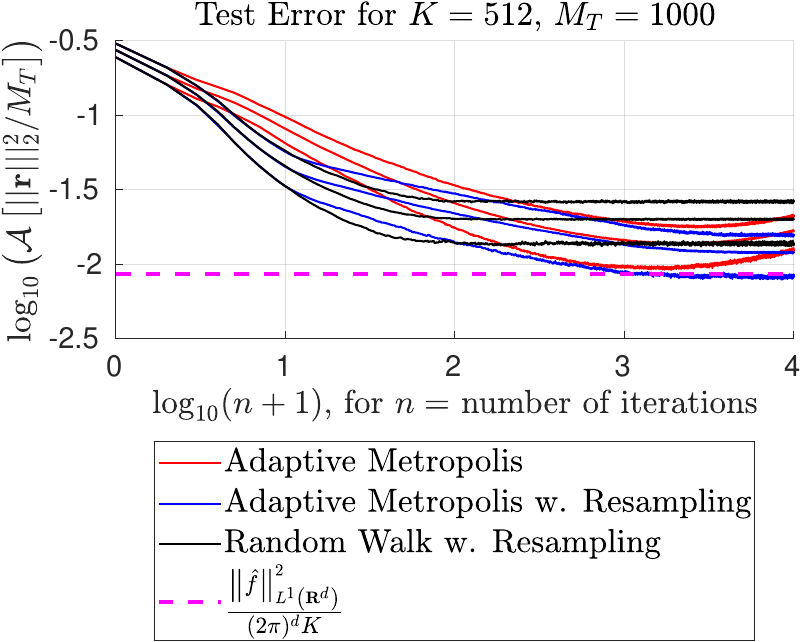}
    \end{center}
    
	\caption{Test~\ref{test:statistics} (i.e.,~\eqref{eq:reg_disc_data_set} with $B$ in~\eqref{eq:rot_mat} 
        and parameters in Table~\ref{tab:Sigint_statistics}) 
		based on 100 independent realizations of the stochastic algorithms for each $K$.\\
		Top row: Convergence of the minimal training and testing errors w.r.t. the number of nodes, $K$. 
        Sample means with the error bars that indicating a confidence interval of $\pm 2$ 
        sample standard deviations. 
        The $\lambda\to 0$ limit of the error estimate~\eqref{eq:error_bound} is included for reference.\\ 
		Bottom row: 
        Errors for $K=512$ as a function of the number of iterations. Sample means and sample means $\pm 2$ 
        sample standard deviations.
		}
	\label{fig:Sigint_statistics}
\end{figure}

The top row of Figure~\ref{fig:Sigint_statistics} depicts the convergence of the sample average of the minimal values 
of $\|\mathbf{r}\|_2^2/M_B$ and $\|\mathbf{r}\|_2^2/M_T$ over the $10^4$ iterations as a function of $K$
for \algAM, \algAMR, and \algRWR. 
The error bars are based on $\pm 2$ sample standard deviations. The confidence intervals for the minimal training 
errors are narrow compared to the difference between \algAM, \algAMR, and \algRWR.
As expected, fixing $M_T=10^3$ as $K$ and $M$
increase leads to higher statistical uncertainty in the testing errors than in the training errors. The same would hold
in Tests~\ref{test:all_data} to~\ref{test:effect_init}, where only one random outcome is presented.
The bottom row of Figure~\ref{fig:Sigint_statistics} depicts the convergence of the sample average of 
$\|\mathbf{r}\|_2^2/M_B$ and $\|\mathbf{r}\|_2^2/M_T$ with respect to the iterations. Curves corresponding to 
$\pm 2$ sample standard deviations are also included. Notably, the general trend with  \algRWR~stagnating at a 
slightly larger error and \algAM~having a slower initial convergence and increasing again as the iterations run into 
the thousands, is also evident for the confidence intervals.

\paragraph{\textbf{Test~\testnr{test:all_data}:} The full data-set in each iteration with $\gamma=10$ in the Metropolis tests.}

This convergence study for an increasing number of nodes, $K$, applied the parameter values in Tables~\ref{tab:Alg_defs_1} 
and~\ref{tab:Sigint_full} and was based on one realization of the stochastic algorithms for each $K$.
The number of data points was $M=K^2$, following the error analysis in~\cite{huang2024convergenceratesrandomfeature}.
Employing a batch size of $M_B=M$ is impractical because then the complexity of constructing least-square problems in each iteration is $M^2=K^4$, 
but an excessively large $M_B$ in this test effectively means over-killing one error source.

\begin{table}[ht]
	\centering
	\begin{tabular}{|c|c|c|c|c|c|c|c|c|}
	\hline
	Problem & $K$ & $M$ & $N$ & $\delta$ & $\gamma$ & $M_B$ & $\lambda$ & $\omegaVector^0$ \\ 
	\hline
	\multirow{6}{*}{\eqref{eq:NN-Def},\eqref{eq:NonConvexOptProblem},\eqref{eq:reg_disc_data_set},\eqref{eq:rot_mat}} & 
	32 & 
	\multirow{6}{*}{$K^2$} &
	\multirow{6}{*}{$10^4$} &
	$2^{-0.75}$ & 
	\multirow{6}{*}{$10$} &
	\multirow{6}{*}{$M$} &
	\multirow{6}{*}{$0.1$} &
	\multirow{6}{*}{$\mathbf{0}$} \\
	 & $64$ & & & $2^{-1}$ & & & & \\ 
	 & $128$ & & & $2^{-1.25}$ & & & & \\ 
	 & $256$ & & & $2^{-1.5}$ & & & & \\ 
	 & $512$ & & & $2^{-1.75}$ & & & & \\ 
	 & $1024$ & & & $2^{-2}$ & & & & \\
	\hline
	\end{tabular}
	\caption{Algorithm parameters corresponding to Algorithm~\ref{alg:AMRS_E} as used in Test~\ref{test:all_data} 
		(see Figure~\ref{fig:Sigint_full}) and as a reference in Tests~\ref{test:effect_gamma} and~\ref{test:effect_batch} 
		(see Figures~\ref{fig:rs_less_sensitive},~\ref{fig:reduced_batch_size}, and~\ref{fig:reduced_batch_size_ESS}.)}
	\label{tab:Sigint_full}
\end{table}

\begin{figure}[ht]
    \begin{center}
	   \includegraphics[width=0.48\linewidth]{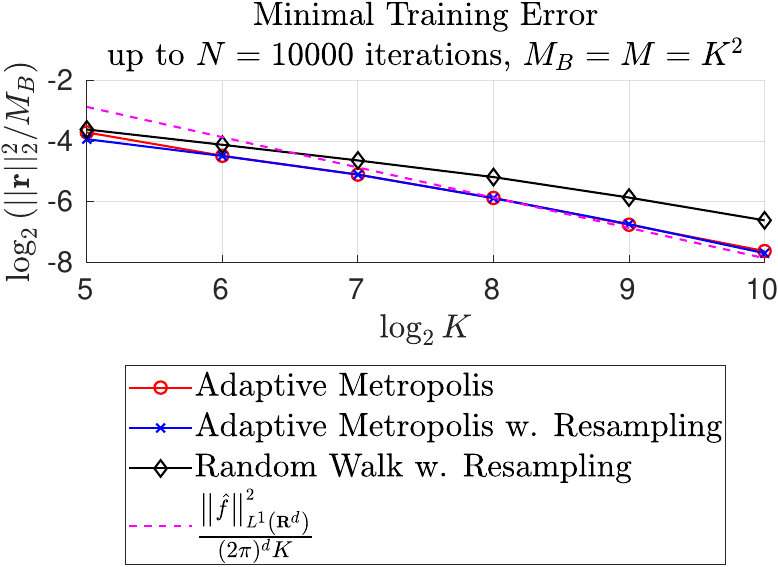}
	   \hspace{0.02\linewidth}
	   \includegraphics[width=0.48\linewidth]{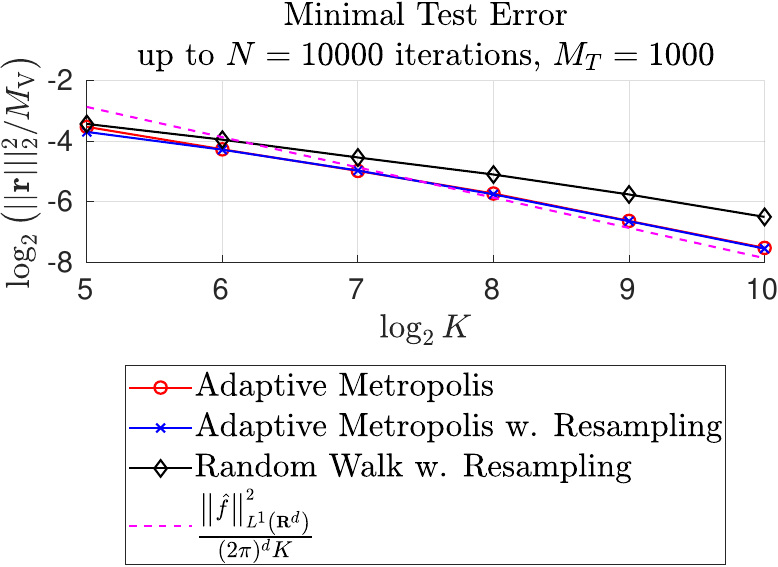}
    \end{center}

	\vspace{5mm}
    \begin{center}
	   \includegraphics[width=0.48\linewidth]{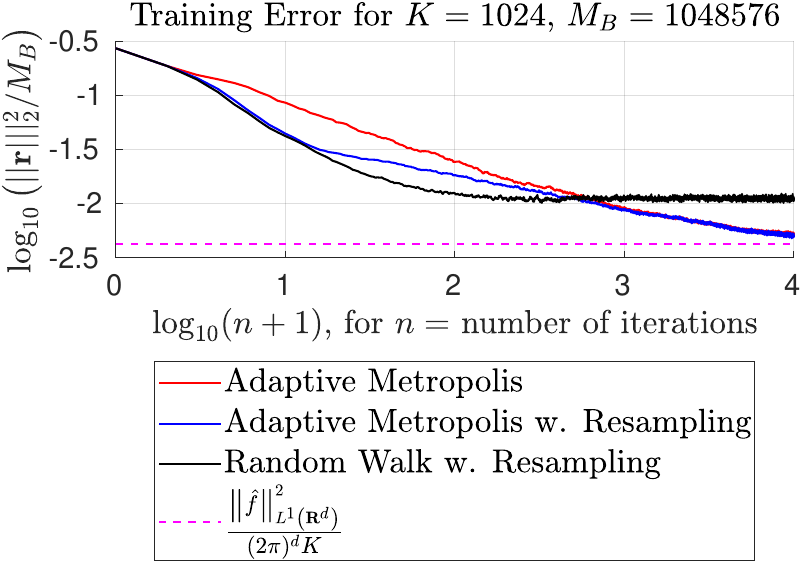}
	   \hspace{0.02\linewidth}
	   \includegraphics[width=0.48\linewidth]{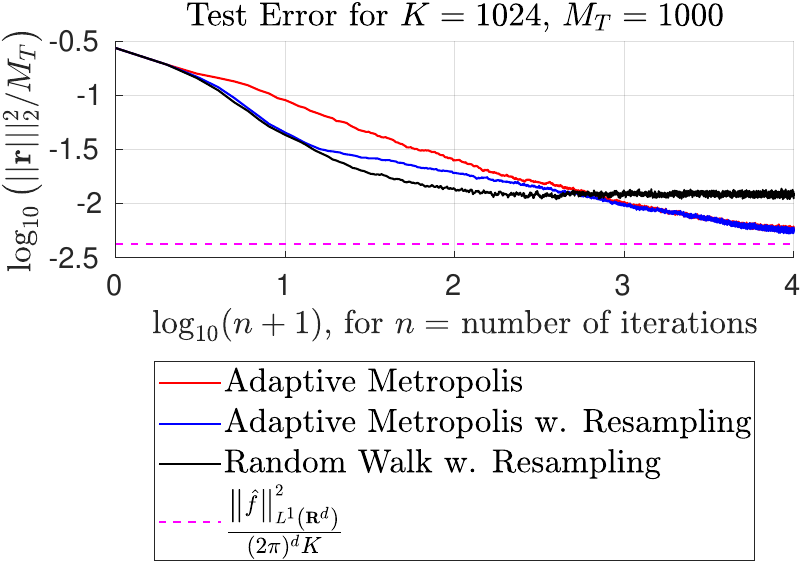}
    \end{center}
	
	\vspace{5mm}
    \begin{center}
	   \includegraphics[width=0.48\linewidth]{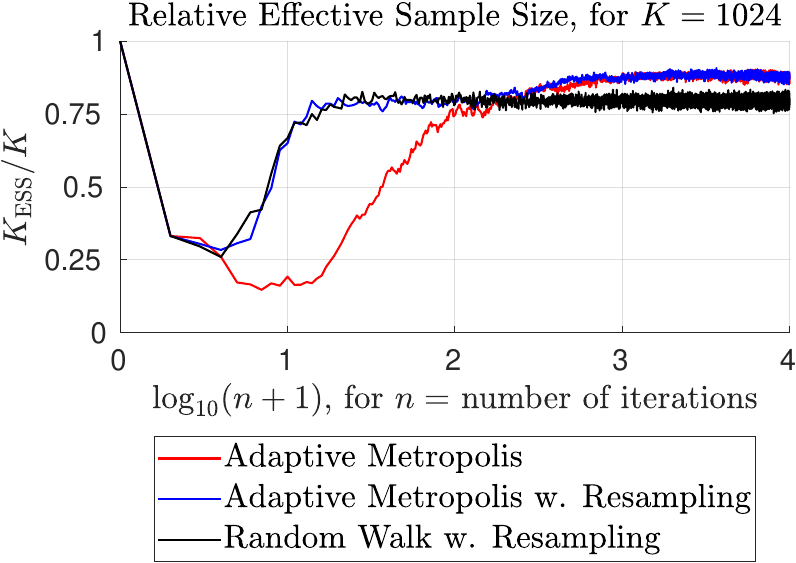}
    \end{center}
	\caption{Test~\ref{test:all_data} (i.e.,~\eqref{eq:reg_disc_data_set} with $B$ in~\eqref{eq:rot_mat} and parameters in Table~\ref{tab:Sigint_full}) 
		with one realization of the stochastic algorithms for each $K$. \\
		Top row: Convergence of the minimal training and testing errors w.r.t. the number of nodes, $K$. The $\lambda\to 0$ limit of the error estimate~\eqref{eq:error_bound} is included for reference.\\ 
		Middle row: 
        Errors for $K=1024$ as a function of the number of iterations.\\
		Bottom row: Normalized effective sample size, $K_\mathrm{ESS}/K$, with $K_\mathrm{ESS}$ defined in~\eqref{eq:ESS}, for $K=1024$ as a function of the number of iterations.
		}
	\label{fig:Sigint_full}
\end{figure}

In the example from the top row of Figure~\ref{fig:Sigint_full} 
the $\lambda\to 0$ limit of the error bound~\eqref{eq:error_bound} 
is almost attained after training with \algAM~and \algAMR, 
although, as presented in the middle row, 
the convergence has not yet quite stagnated for $K=1024$. 
The difference between the training and testing errors, computed from a set of 1000 data points independent 
from the training data, is small. The middle row of Figure~\ref{fig:Sigint_full} reveals that  
\algAMR~and \algRWR~reduce
the error faster in the first iterations; this observation holds in general in our experience. 
This finding is also reflected in the fact that it takes longer for the ESS, $K_\mathrm{ESS}$, to rebound from the initial decline when 
\algAM~is applied. 
For $K=1024$, it takes around 100 iterations for
\algAM~to
recover to a level reached after about 10 iterations 
of \algAMR~and \algRWR~(bottom row in Figure~\ref{fig:Sigint_full}). 
When training with 
\algRWR, the
error stagnates at a higher level than when training with \algAMR~and \algRWR.
In contrast, the latter two nearly attain the predicted error.

For $M_B=M=K^2$ with a large $K$ value, 
constructing and solving the least-square systems dominates the computational cost, and the number of times this is performed per iteration is one for \algRWR, two for \algAM, and either two or three (depending on $K_\mathrm{ESS}$ and $R(n)$) for \algAMR.
The bottom row in Figure~\ref{fig:Sigint_full} reveals that because the resampling threshold was set 
at $0.75 K$ and because $K_\mathrm{ESS}$ mostly remained above this threshold after the first 20 or 30 iterations, resampling was only 
used sparingly in this test after the initial phase so that the cost of 
\algAM~and \algAMR~was
approximately the same.

Tests that were not included in the figure, 
with Adaptive Metropolis with resampling in every iteration (i.e., $R(n)\equiv 1$) present results similar to those of 
\algRWR~but at a higher computational cost.

\paragraph{\textbf{Test~\testnr{test:effect_gamma}:} Full data-set in each iteration, comparing $\gamma=1$ and $\gamma=10$.}

Based on the values of the parameters in Tables~\ref{tab:Alg_defs_1} and~\ref{tab:Sigint_gamma_1}, this test used $\gamma=1$ for $K=256$ nodes
to illustrate the sensitivity of the Adaptive Metropolis algorithm w.r.t. the parameter $\gamma$.

\begin{table}[ht]
	\centering
	\begin{tabular}{|c|c|c|c|c|c|c|c|c|}
	\hline
	Problem & $K$ & $M$ & $N$ & $\delta$ & $\gamma$ & $M_B$ & $\lambda$ & $\omegaVector^0$ \\ 
	\hline
    \eqref{eq:NN-Def},\eqref{eq:NonConvexOptProblem},\eqref{eq:reg_disc_data_set},\eqref{eq:rot_mat} &
	256 & 
	$K^2$ &
	$10^4$ &
	$2^{-1.5}$ & 
	$1$ &
	$M$ &
	$0.1$ &
	$\mathbf{0}$ \\
	\hline
	\end{tabular}
	\caption{Algorithm parameters corresponding to Algorithm~\ref{alg:AMRS_E}, 
        as used in Test~\ref{test:effect_gamma} (see Figure~\ref{fig:rs_less_sensitive}). 
        The choice of $\gamma$ differs from Table~\ref{tab:Sigint_full}.}
	\label{tab:Sigint_gamma_1}
\end{table}

The choice of $\gamma=10$ in Test~\ref{test:all_data} follows the rule of thumb of choosing $\gamma=3d-2$, 
justified in~\cite{kammonen2020adaptive} for Gaussian target functions as striking a balance between the accuracy of approximating 
the optimal distribution and computational efficiency. 
Instead, using $\gamma=1$ corresponds to approximating the unknown, ideal sampling density using the modulus of the amplitudes, 
$\left\vert \ampl_k\right\vert / \overline{\ampl}$ in the Metropolis algorithm.

\begin{figure}[ht]
    \begin{center}
	   \includegraphics[width=0.48\linewidth]{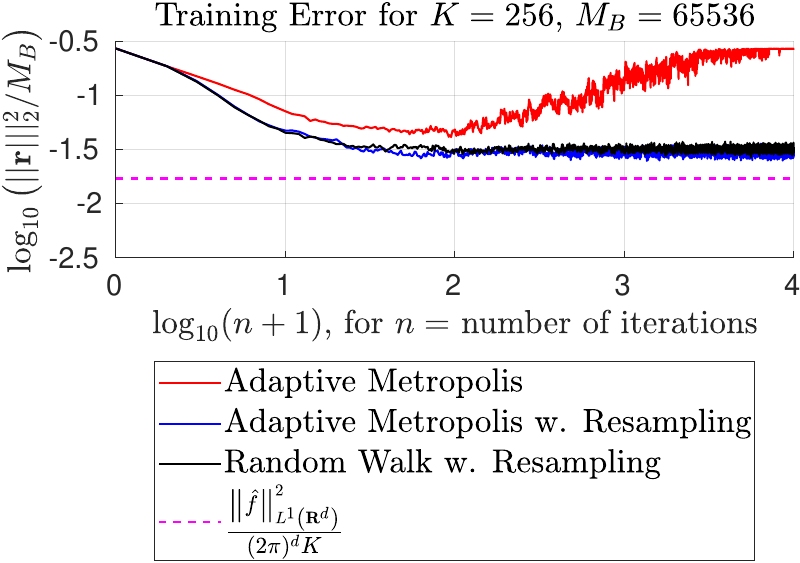}
	   \hspace{0.02\linewidth}
	   \includegraphics[width=0.48\linewidth]{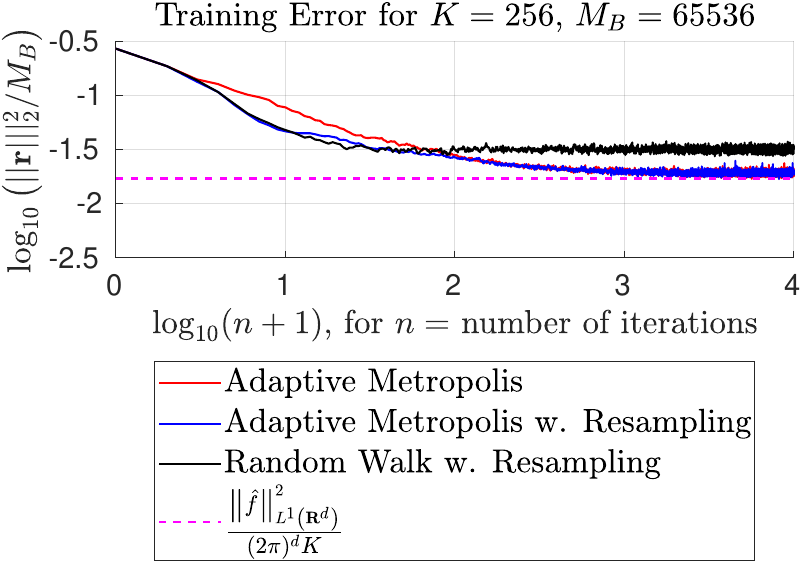}
    \end{center}

	\vspace{5mm}
    \begin{center}
	   \includegraphics[width=0.48\linewidth]{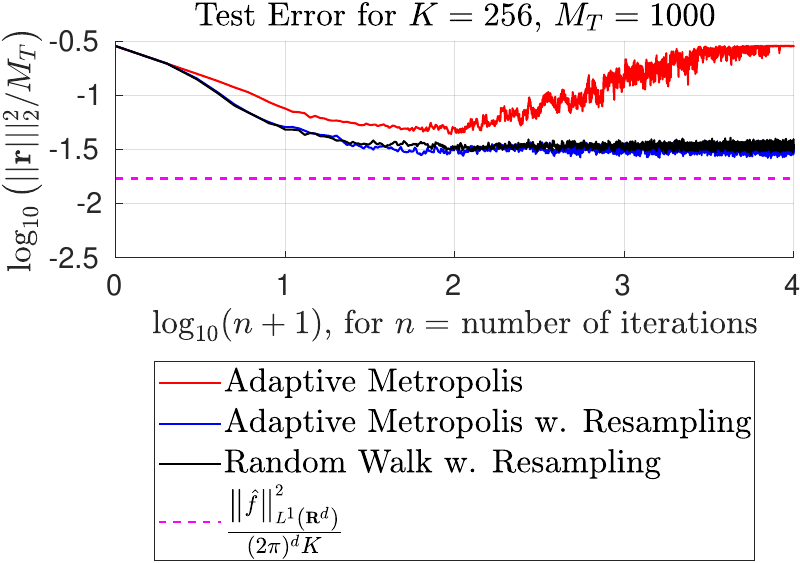}
	   \hspace{0.02\linewidth}
	   \includegraphics[width=0.48\linewidth]{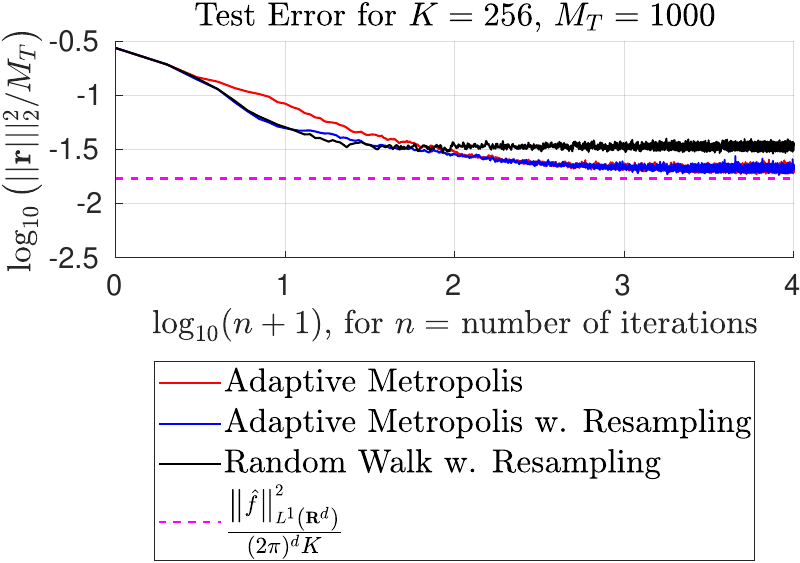}
    \end{center}
	
	\vspace{5mm}
    \begin{center}
	   \includegraphics[width=0.48\linewidth]{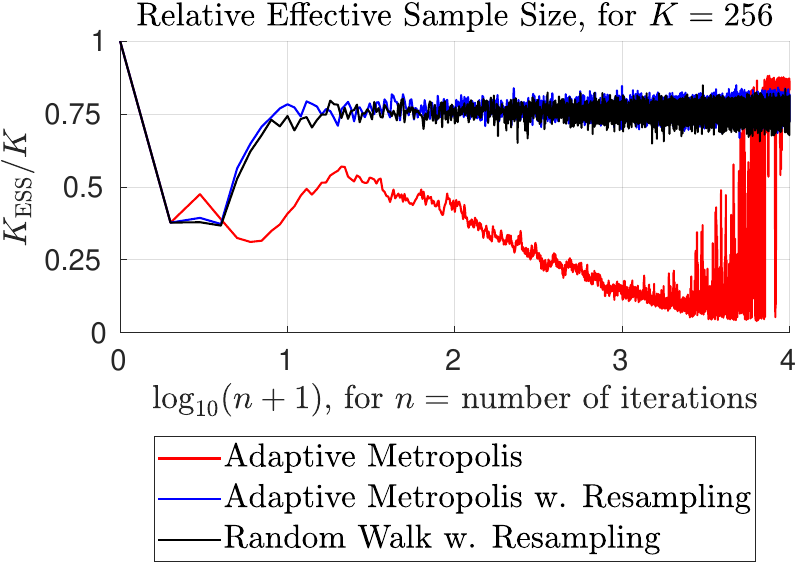}
	   \hspace{0.02\linewidth}
	   \includegraphics[width=0.48\linewidth]{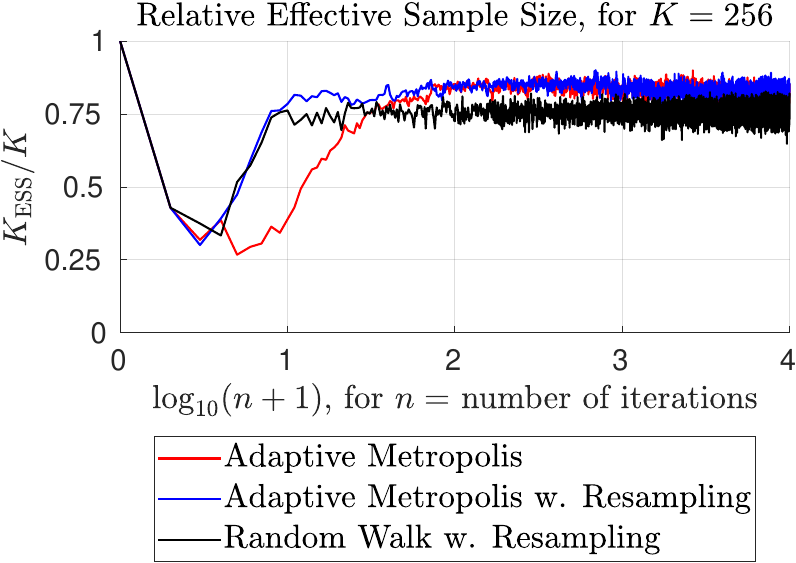}
    \end{center}
	\caption{Test~\ref{test:effect_gamma} (i.e.,~\eqref{eq:reg_disc_data_set} with $B$ in~\eqref{eq:rot_mat}) illustrating sensitivity w.r.t. $\gamma$.
		One realization of the stochastic algorithms is included.\\
		Left column: The case $\gamma=1$, with parameters in Table~\ref{tab:Sigint_gamma_1}.\\ 
		Right column: The case $\gamma=10$, with parameters in Table~\ref{tab:Sigint_full} for $K=256$.
		}
	\label{fig:rs_less_sensitive}
\end{figure}

The left column of Figure~\ref{fig:rs_less_sensitive} reveals that \algAM, with $\gamma=1$, 
fails to reach the theoretical error bound associated with sampling from the ideal distribution. 
Instead, the training and testing errors increase after fewer than 100 iterations as the ESS decreases. 
In contrast, the errors using \algAMR~stagnate before reaching the error bound but do so at a lower level.
The stabilizing effect that resampling has on the Adaptive Metropolis algorithm is illustrated by the fact 
that the behavior is comparable to that of \algRWR, where the parameter $\gamma$ does not appear.

For comparison, the right column of Figure~\ref{fig:rs_less_sensitive} provides the results from the case of $K=256$ in Test~\ref{test:all_data}. 
For 
\algRWR, 
the results in the two columns are independent realizations of the stochastic algorithm with identical parameter values.
For \algAM~and \algAMR, the difference is the value of $\gamma$. 
When using \algAM~with $\gamma=10$, the theoretical error bound is almost attained. 
When using \algAMR~with $\gamma=10$, the errors initially decay faster than for 
\algAM~and eventually stagnate around the level of the theoretical 
error bound. These findings again support the observation that resampling reduces the sensitivity of the adaptive Metropolis method w.r.t. the value of $\gamma$.

\begin{figure}[ht]
    \begin{center}
	   \includegraphics[width=0.48\linewidth]{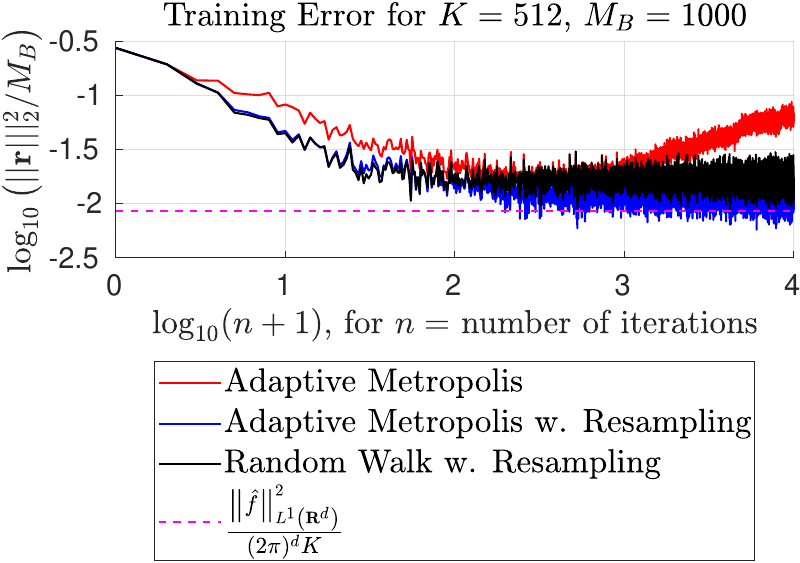}
	   \hspace{0.02\linewidth}
	   \includegraphics[width=0.48\linewidth]{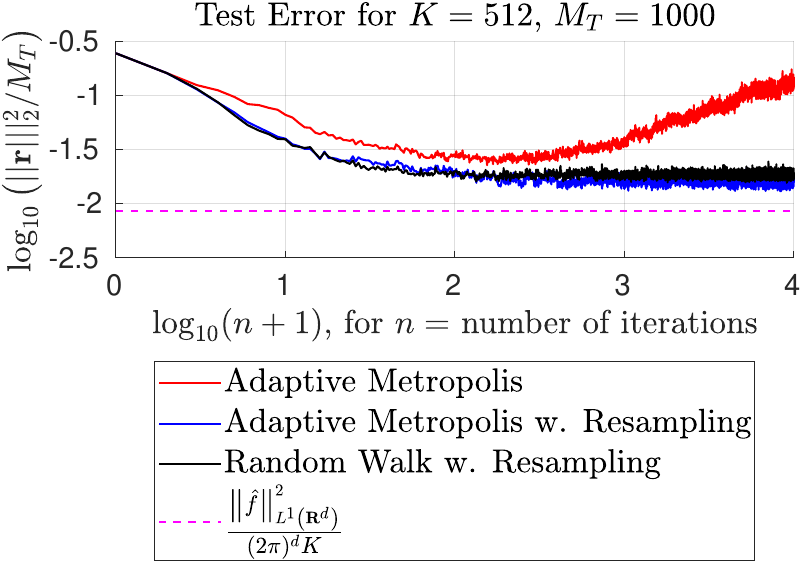}
    \end{center}

	\vspace{5mm}
    \begin{center}
	   \includegraphics[width=0.48\linewidth]{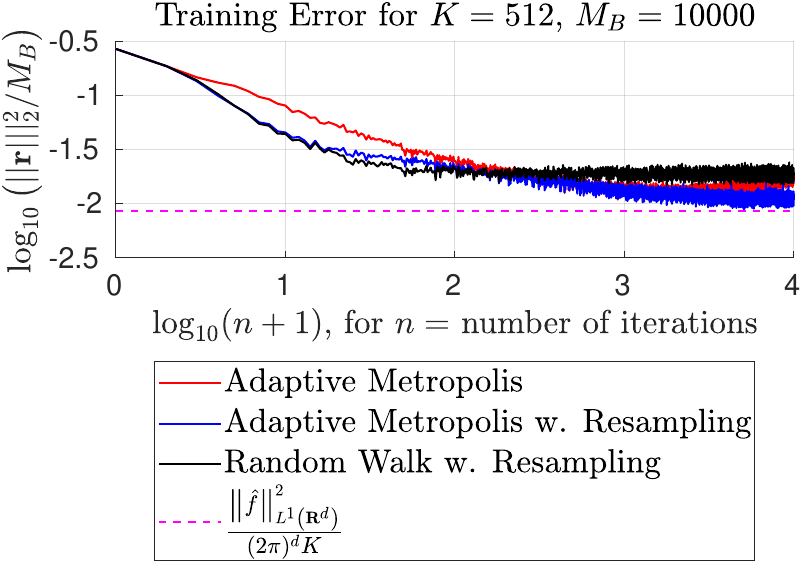}
	   \hspace{0.02\linewidth}
	   \includegraphics[width=0.48\linewidth]{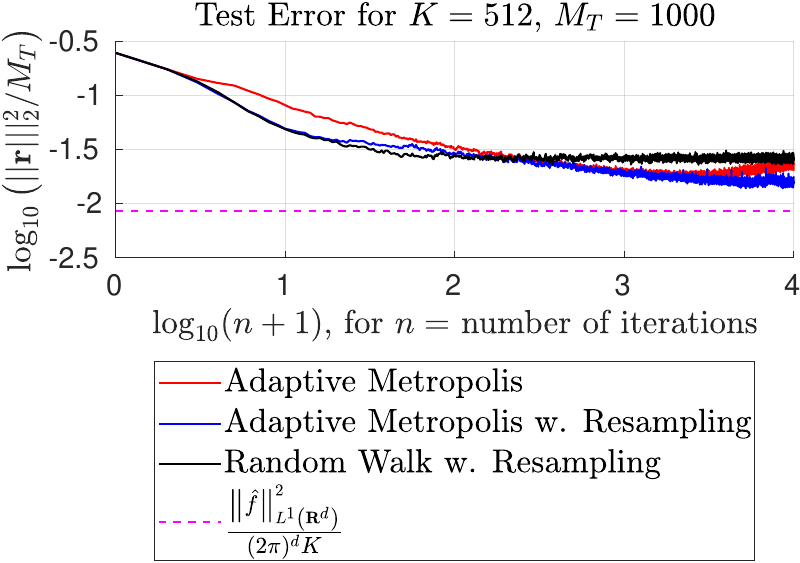}
    \end{center}
	
	\vspace{5mm}
    \begin{center}
	   \includegraphics[width=0.48\linewidth]{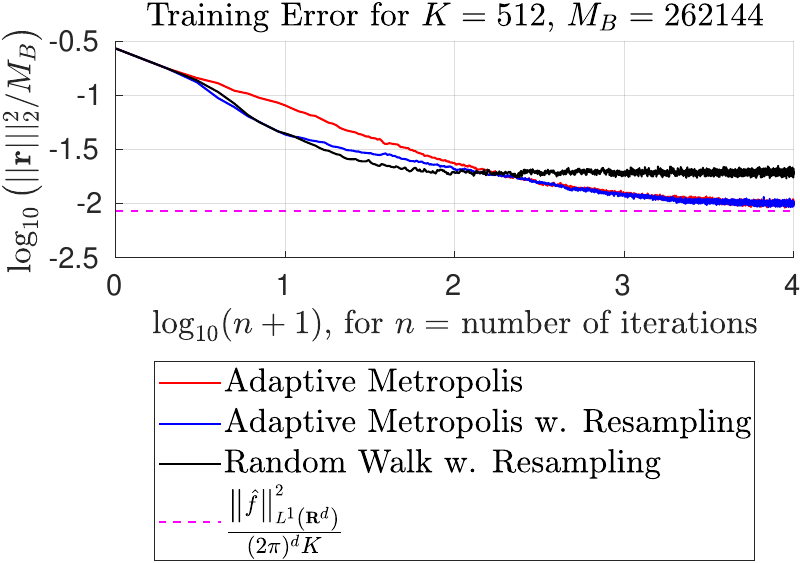}
	   \hspace{0.02\linewidth}
	   \includegraphics[width=0.48\linewidth]{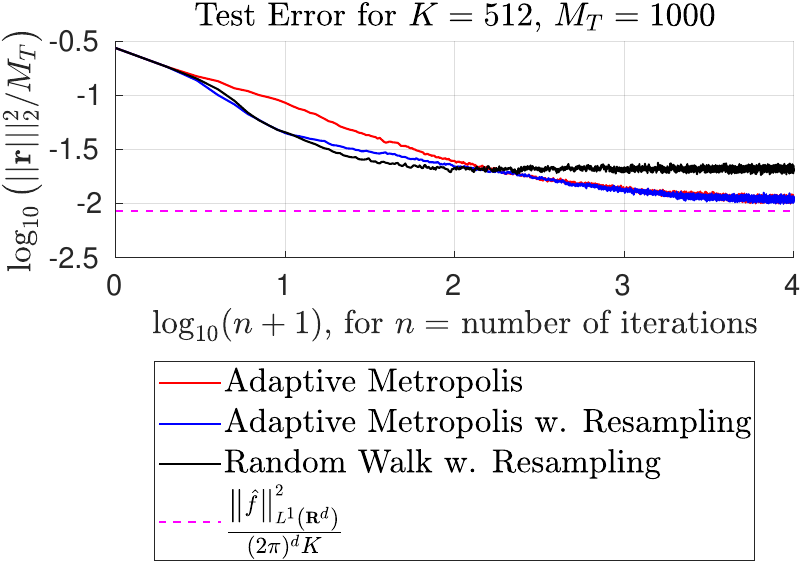}
    \end{center}
	\caption{Test~\ref{test:effect_batch} (i.e.,~\eqref{eq:reg_disc_data_set} with $B$ in~\eqref{eq:rot_mat}) illustrating sensitivity w.r.t. $M_B$.
		One realization of the stochastic algorithms is included.
		Parameters are the same as in the case of $K=512$ in Table~\ref{tab:Sigint_full}, except $M_B$, which is
		$M_B=10^3$, (top), $M_B=10^4$, (middle), and $M_B=M$, (bottom).
		}
	\label{fig:reduced_batch_size}
\end{figure}

\begin{figure}[ht]
    \begin{center}
	   \includegraphics[width=0.48\linewidth]{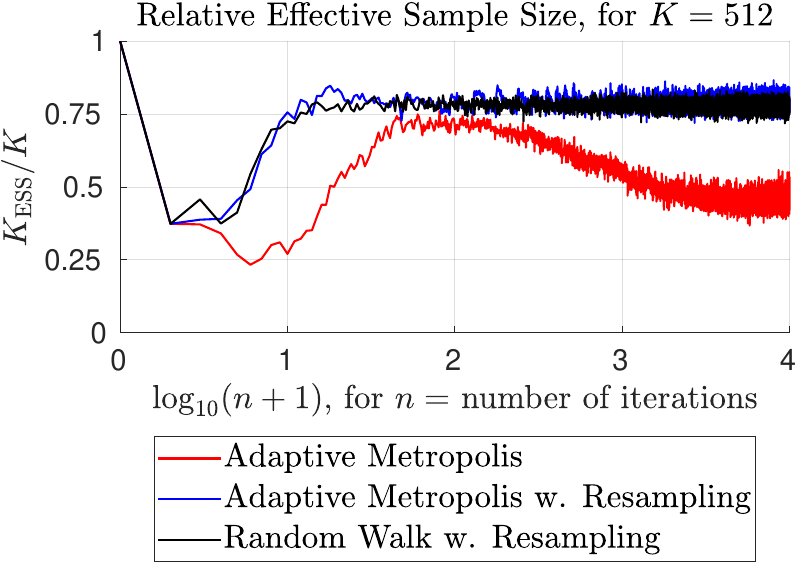}
	   \hspace{0.02\linewidth}
	   \includegraphics[width=0.48\linewidth]{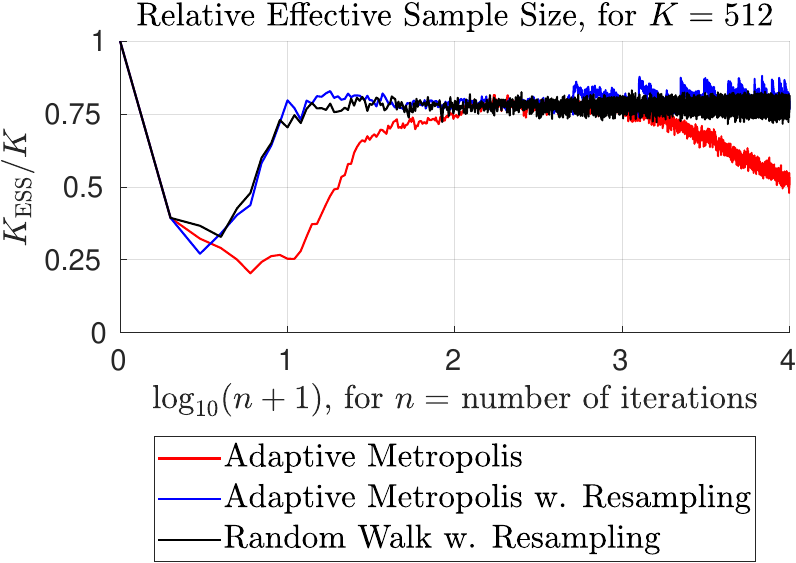}
    \end{center}

	\vspace{5mm}
    \begin{center}
	   \includegraphics[width=0.48\linewidth]{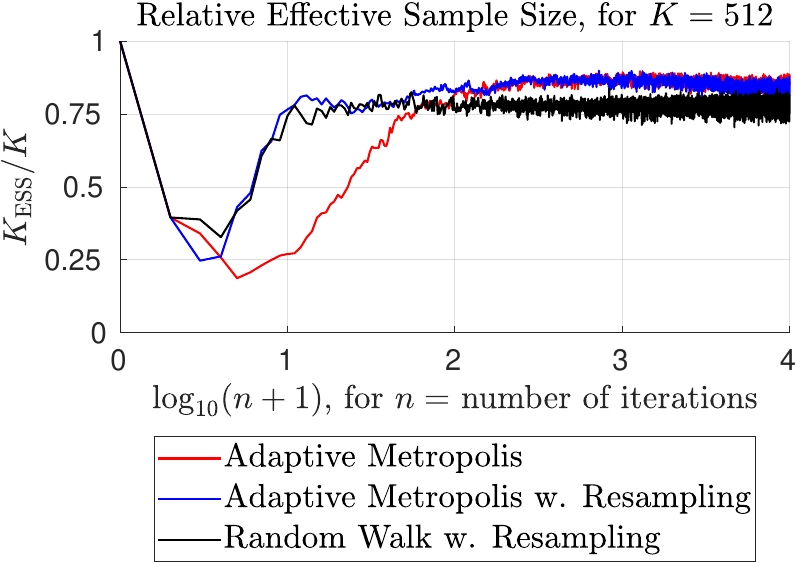}
    \end{center}
	\caption{Test~\ref{test:effect_batch} (i.e.,~\eqref{eq:reg_disc_data_set} with $B$ in~\eqref{eq:rot_mat}) illustrating the 
		effective sample sizes corresponding to Figure~\ref{fig:reduced_batch_size}.
		One realization of the stochastic algorithms is included.
		Parameters are the same as in the case of $K=512$ in Table~\ref{tab:Sigint_full}, except $M_B$, which is
		$M_B=10^3$, (top), $M_B=10^4$, (middle), and $M_B=M$, (bottom).
		}
	\label{fig:reduced_batch_size_ESS}
\end{figure}

\paragraph{\textbf{Test~\testnr{test:effect_batch}:} Using a batch size of $M_B<M$.} 
As noted above, including all training data in each iteration is too expensive when $K$ is large. 
Figure~\ref{fig:reduced_batch_size} shows one outcome of all three versions of Algorithm~\ref{alg:AMRS_E} in Table~\ref{tab:Alg_defs_1} for 
the case of $K=512$ in Table~\ref{tab:Sigint_full}, but with a reduced batch size in the two upper rows of the figure 
(Table~\ref{tab:reduced_batch_size}). The bottom row is the result for $K=512$ in Test~\ref{test:all_data}. 
In the top row, with a batch size of $M_B=10^3$, the training and testing errors of 
\algAM~increase 
again after only a few hundred iterations before reaching the theoretical error bound. 
The training errors of 
\algAMR~and \algRWR~appear
to stagnate around the level of the error bound, 
but the testing errors stagnate at a slightly higher level. However, the errors do not increase when resampling is applied.
Increasing the batch size to $M_B=10^4$ somewhat stabilizes \algAM,
although the errors start to increase after
a few thousand iterations. The \algAM~and \algAMR~testing errors reach
lower values than those of \algRWR~but do not reach the error bound. 
Finally, the results from Test~\ref{test:all_data} indicate that 
\algAM~and \algAMR~reach the level of 
the error bound, and none of the versions display the tendency of increasing errors in the first $10^4$ iterations.
Figure~\ref{fig:reduced_batch_size_ESS} reveals that, in this instance, a reduced batch size leads to a premature 
decrease in the ESS of \algAM. 
Resampling solves this issue in the contexts of both the adaptive Metropolis method and a simple random walk method.

\begin{table}[ht]
	\centering
	\begin{tabular}{|c|c|c|c|c|c|c|c|c|}
	\hline
	Problem & $K$ & $M$ & $N$ & $\delta$ & $\gamma$ & $M_B$ & $\lambda$ & $\omegaVector^0$ \\ 
	\hline
    \multirow{3}{*}{\eqref{eq:NN-Def},\eqref{eq:NonConvexOptProblem},\eqref{eq:reg_disc_data_set},\eqref{eq:rot_mat}} & 
	\multirow{3}{*}{512}& 
	\multirow{3}{*}{$K^2$} &
	\multirow{3}{*}{$10^4$} &
	\multirow{3}{*}{$2^{-1.75}$} &
	\multirow{3}{*}{$10$} &
    $10^3$ &
	\multirow{3}{*}{$0.1$} &
	\multirow{3}{*}{$\mathbf{0}$} \\
	 & & & & & & $10^4$ & & \\
	 & & & & & & $M$ & & \\
	\hline 
	\end{tabular}
	\caption{Algorithm parameters corresponding to Algorithm~\ref{alg:AMRS_E}, as used in Test~\ref{test:effect_batch} 
        (see Figures~\ref{fig:reduced_batch_size} and~\ref{fig:reduced_batch_size_ESS}).}
	\label{tab:reduced_batch_size}
\end{table}

This test demonstrates that including resampling improves the stability of the adaptive Metropolis method with smaller 
batch sizes, which are crucial for computational efficiency.

\paragraph{\textbf{Test~\testnr{test:effect_init}:} Non-degenerate initial distribution of frequencies.}
In Tests~\ref{test:statistics} to~\ref{test:effect_batch}, all frequencies were initialized to $\boldsymbol{0}$. 
That choice was motivated by the observation that the adaptive Metropolis method
is inefficient at concentrating the frequency distribution. This test instead initializes the 
frequencies with independent samples from the four-dimensional 
$N\left(\boldsymbol{0},10^2 \ID\right)$ distribution.
For this manufactured test problem, the initial distribution was purposely selected to be excessively spread out in three directions while being more concentrated than the ideal 
sampling distribution in the other direction. 
Figure~\ref{fig:init_normal_KDEs} displays kernel density estimates (KDEs) of the distributions of $\left[B^{-1}\omegaVector\right]_j$, 
along the direction $j=1$, where the initial sample distribution is more concentrated than the ideal one, (left column) and 
$j=2$, where it is too spread out (right column).
This test applies the parameter values in Tables~\ref{tab:Alg_defs_1} and~\ref{tab:Normal_init}.

\begin{table}[ht]
	\centering
	\begin{tabular}{|c|c|c|c|c|c|c|c|c|}
	\hline
	Problem & $K$ & $M$ & $N$ & $\delta$ & $\gamma$ & $M_B$ & $\lambda$ & $\omegaVector^0$ \\ 
	\hline
    \multirow{6}{*}{\eqref{eq:NN-Def},\eqref{eq:NonConvexOptProblem},\eqref{eq:reg_disc_data_set},\eqref{eq:rot_mat}} & 
	32 & 
	\multirow{6}{*}{$K^2$} &
	\multirow{6}{*}{$10^4$} &
	$2^{-0.75}$ & 
	\multirow{6}{*}{$10$} &
    $K^2$ &
	\multirow{6}{*}{$0.1$} &
	\multirow{6}{*}{$N\left(\boldsymbol{0},10^2 \ID\right)$} \\
	 & $64$ & & & $2^{-1}$ & & $K^2$ & & \\ 
	 & $128$ & & & $2^{-1.25}$ & & $10^4$ & & \\ 
	 & $256$ & & & $2^{-1.5}$ & & $10^4$ & & \\ 
	 & $512$ & & & $2^{-1.75}$ & & $10^4$ & & \\ 
	 & $1024$ & & & $2^{-2}$ & & $10^4$ & & \\ 
	\hline
	\end{tabular}
	\caption{Algorithm parameters corresponding to Algorithm~\ref{alg:AMRS_E}, 
        as used in Test~\ref{test:effect_init} (see Figures~\ref{fig:init_normal} and~\ref{fig:init_normal_KDEs}). 
        The choices of $M_B$ and $\omegaVector^0$ differ from Table~\ref{tab:Sigint_full}. }
	\label{tab:Normal_init}
\end{table}

\begin{figure}[ht]
    \begin{center}
      \includegraphics[width=0.48\linewidth]{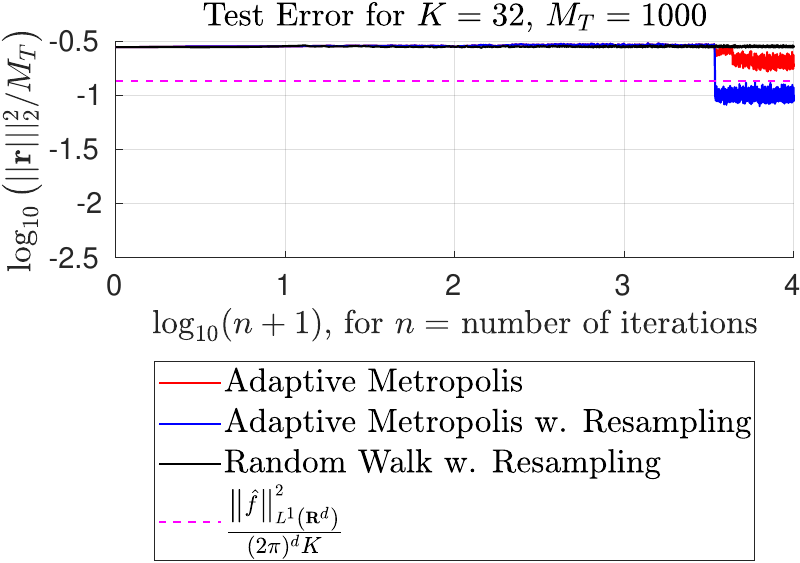}
	   \hspace{0.02\linewidth}
	   \includegraphics[width=0.48\linewidth]{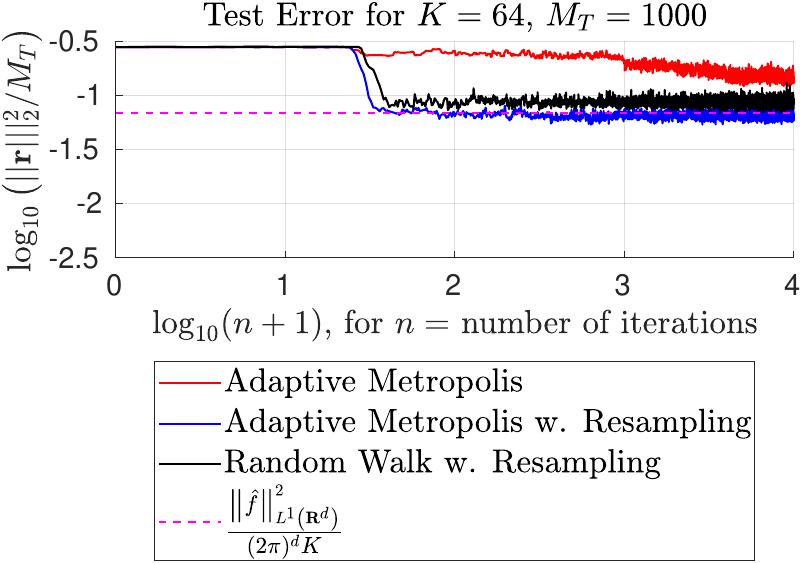}
    \end{center}

	\vspace{5mm}
    \begin{center}
      \includegraphics[width=0.48\linewidth]{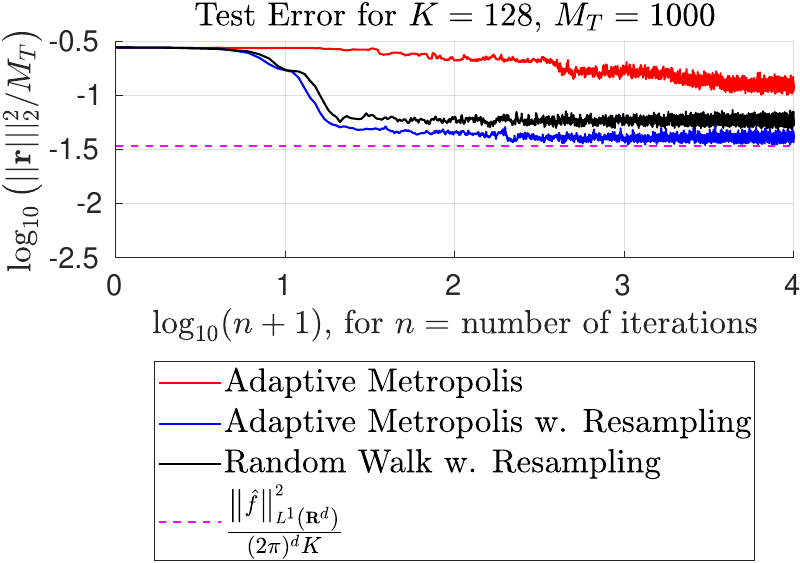}
	   \hspace{0.02\linewidth}
	   \includegraphics[width=0.48\linewidth]{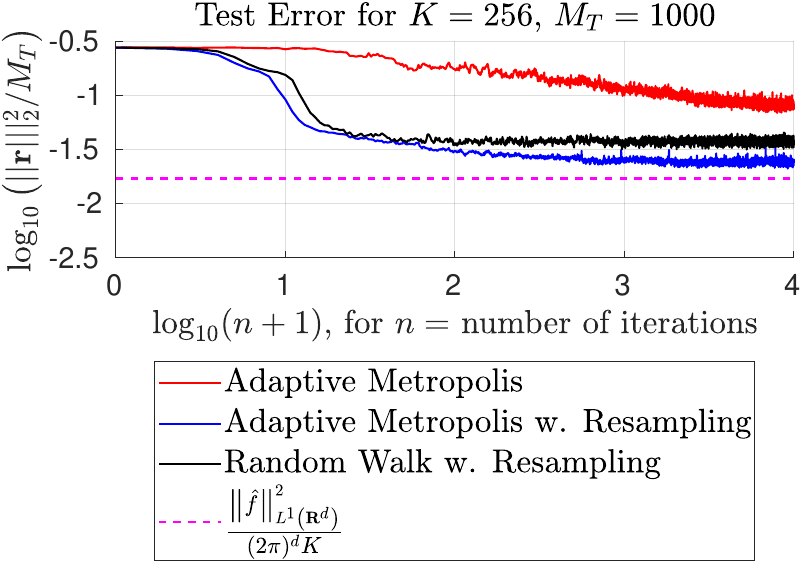}
    \end{center}
	
	\vspace{5mm}
    \begin{center}
	   \includegraphics[width=0.48\linewidth]{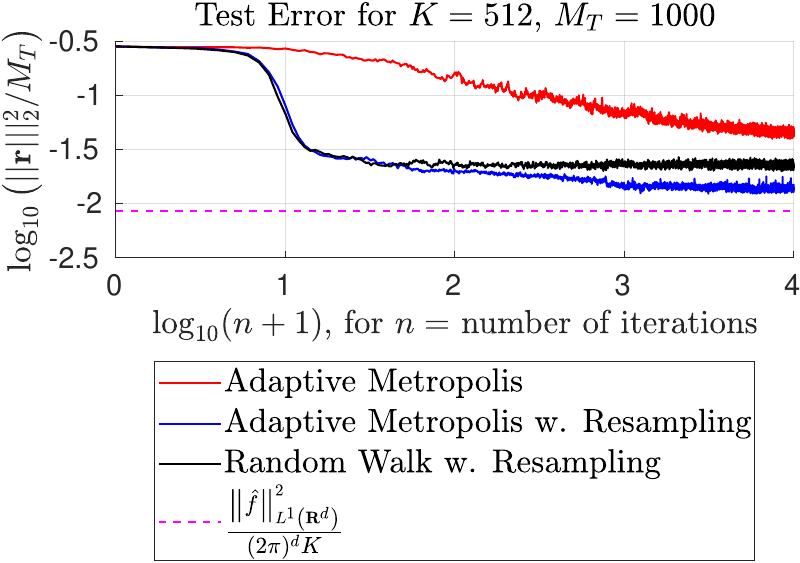}
	   \hspace{0.02\linewidth}
	   \includegraphics[width=0.48\linewidth]{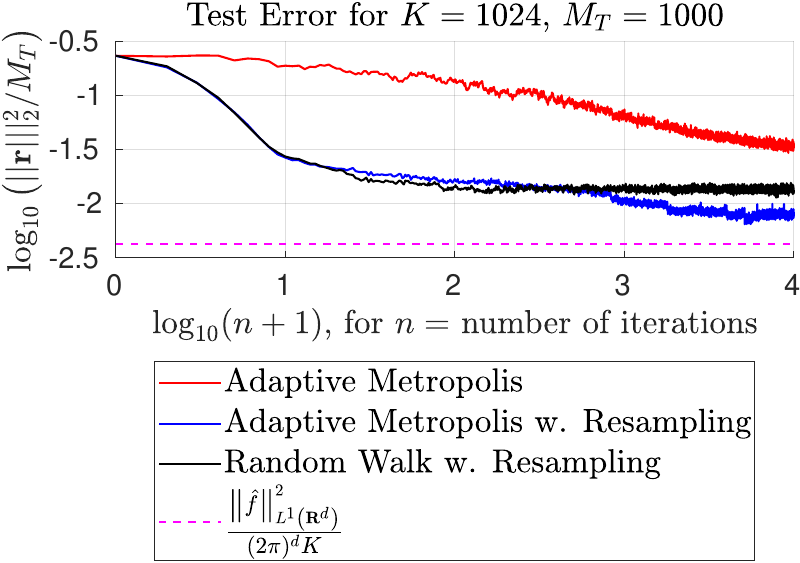}
    \end{center}
	\caption{Test~\ref{test:effect_init} 
        (i.e.,~\eqref{eq:reg_disc_data_set} with $B$ in~\eqref{eq:rot_mat}) illustrating sensitivity w.r.t. the initial frequency distribution.
		One realization of the stochastic algorithms is included for each $K$.
        Parameters are in Table~\ref{tab:Normal_init}.
		}
	\label{fig:init_normal}
\end{figure}

\begin{figure}[ht]
    \begin{center}
	   \includegraphics[width=0.48\linewidth]{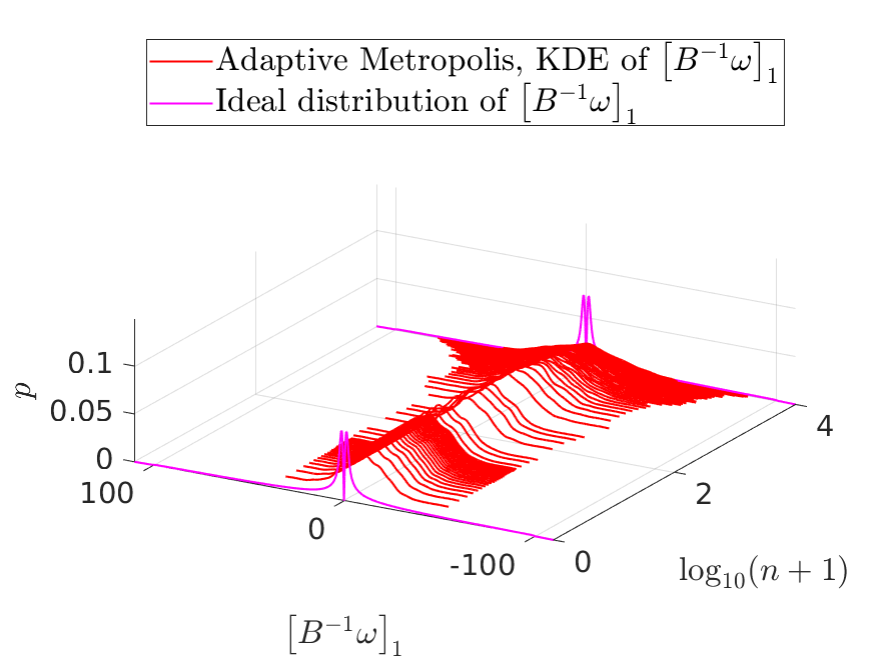}
	   \hspace{0.02\linewidth}
	   \includegraphics[width=0.48\linewidth]{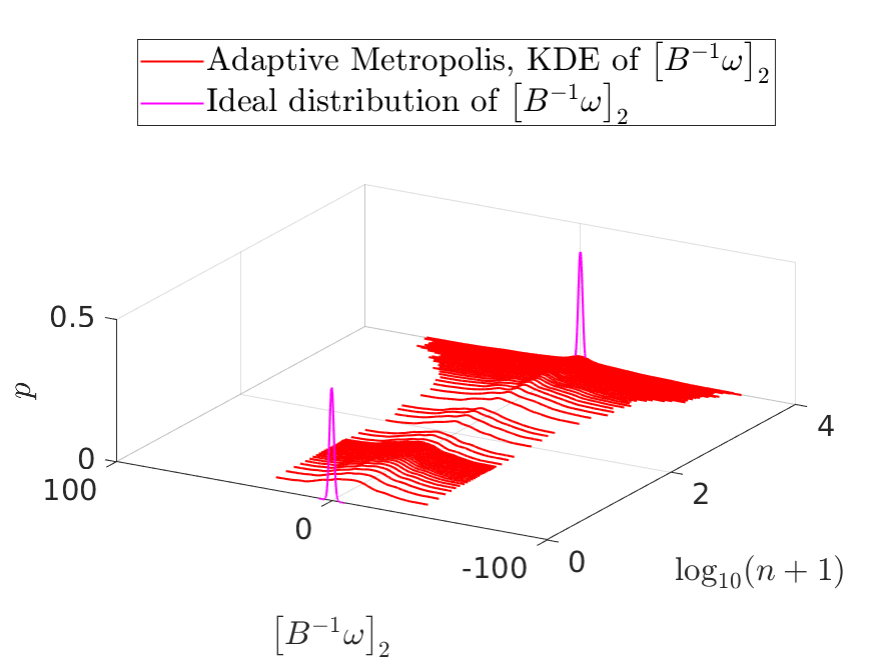}
    \end{center}
    \noindent\rule{\textwidth}{0.4pt}
    \begin{center}
	   \includegraphics[width=0.48\linewidth]{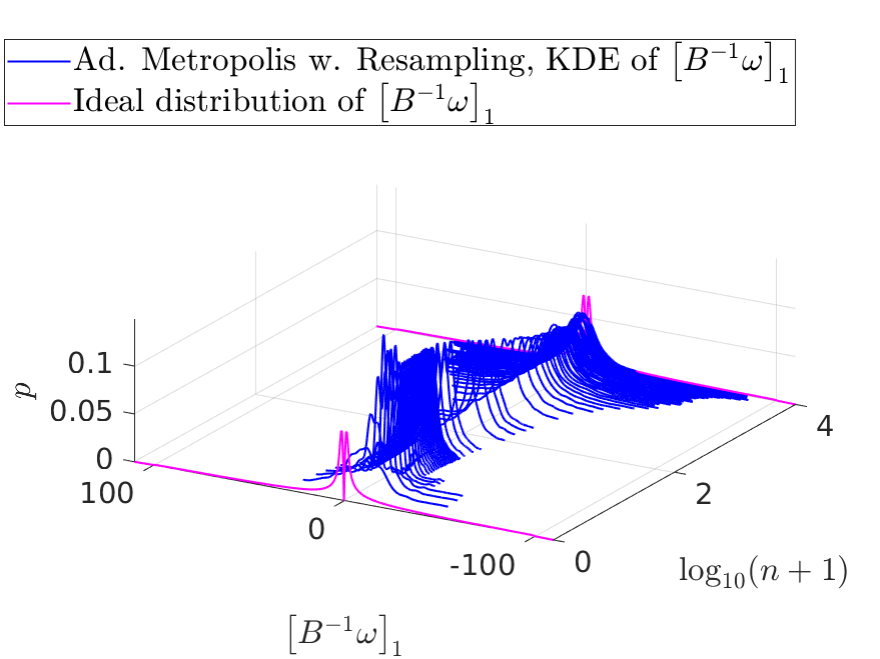}
	   \hspace{0.02\linewidth}
	   \includegraphics[width=0.48\linewidth]{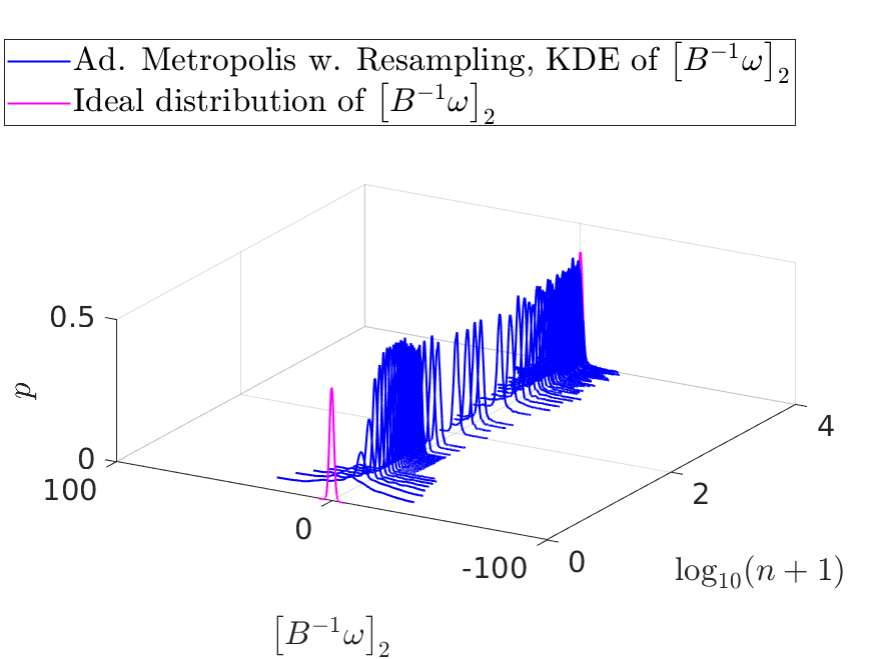}
    \end{center}
    \noindent\rule{\textwidth}{0.4pt}
    \begin{center}
	   \includegraphics[width=0.48\linewidth]{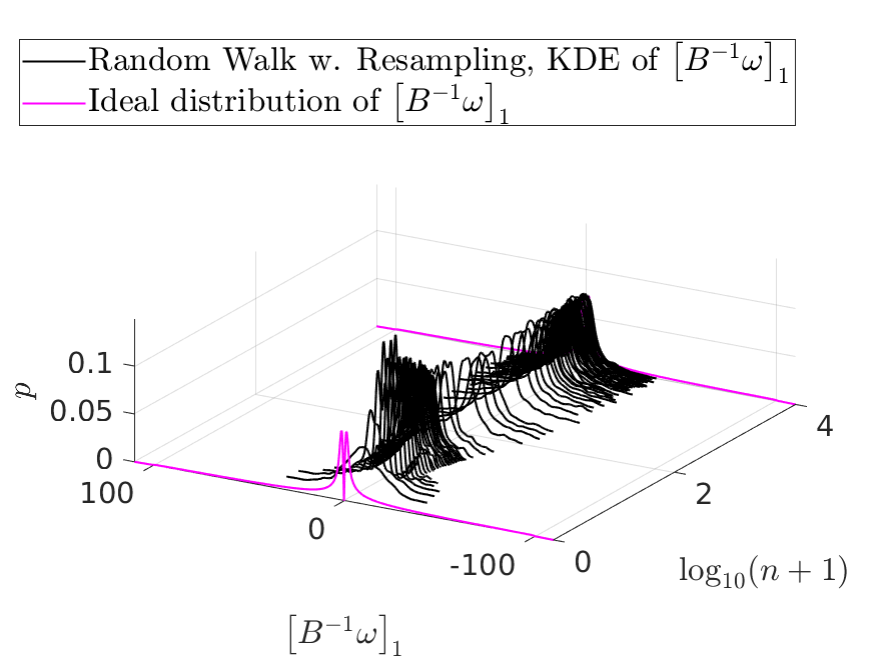}
	   \hspace{0.02\linewidth}
	   \includegraphics[width=0.48\linewidth]{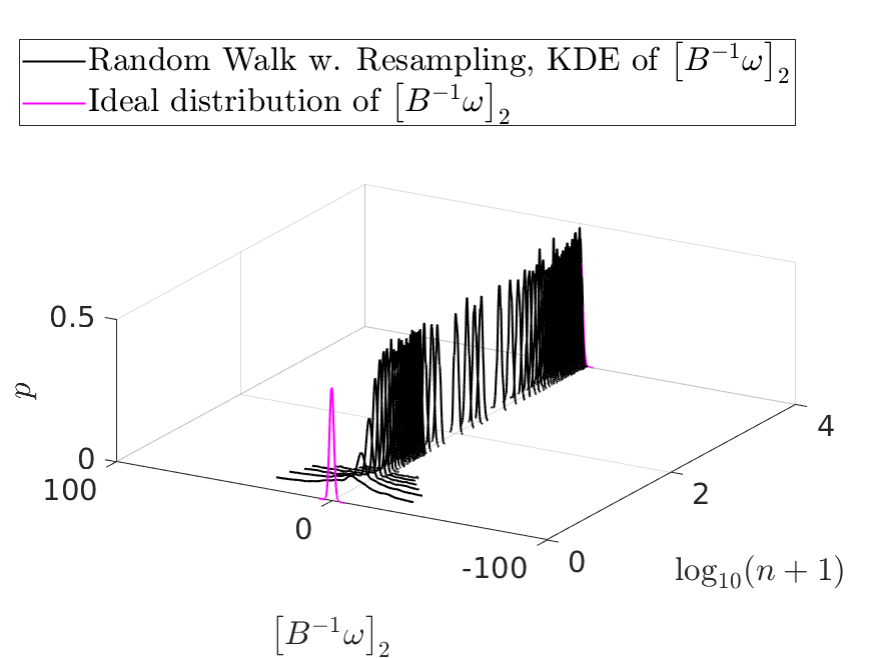}
    \end{center}
	\caption{Test~\ref{test:effect_init}, the case of $K=512$ in Figure~\ref{fig:init_normal}, illustrating the iterative evolution of the frequency distribution. 
	The KDEs, 
    $p$, of the distributions of $\left[B^{-1}\omegaVector\right]_j$, $j=1,2$, are shown for some iterations $n$. 
    Left:
    Distributions along the direction depending on the Sinint function, where the ideal distribution 
	decays more slowly than a Normal until around $1/a=100$. 
    Right:
    Distributions along one of the orthogonal directions, where the ideal distribution is the standard normal. 
	The rows present adaptive Metropolis method without resampling (top) and with resampling (middle), and random walk with resampling (bottom).
	}
	\label{fig:init_normal_KDEs}
\end{figure}

Due to the suboptimal initialization of frequencies the random behavior of the algorithm is more pronounced, particularly 
for the smaller values of $K$. Therefore, which algorithm first demonstrates a sudden decrease in the testing error in a single run, 
as in Figure~\ref{fig:init_normal} is not important. Instead, the purpose of the figure is 
to illustrate the sharpness of the decline in the error value when resampling is included in the algorithm. Once the 
resampling-based versions start to concentrate the sampling distribution, the errors decay much faster over the following iterations.
The failure of the Adaptive Metropolis' method to concentrate sampling distributions is illustrated in the top row of 
Figure~\ref{fig:init_normal_KDEs}, where KDEs of its sample frequencies are compared to the target 
distribution, $p^\ast$, which is known in this test problem. For comparison, the corresponding results of two resampling-based
versions are depicted in the bottom two rows.

This test illustrates the effect resampling has in mitigating unsuccessful initial frequency distributions.

\subsubsection{Algorithm~\ref{alg:AMRS_E} as a Pretraining Algorithm}
\label{sec:num_exp_pre_training}

One suggested use of Algorithm~\ref{alg:AMRS_E} is to pretrain a two-layer neural network before training it with the Adam optimizer~\cite{KingmaBa-ADAM-2017} 
or any other gradient-based optimizer. We first applied training using Algorithm~\ref{alg:AMRS_E} until the validation error stagnated and then switched to training with the Adam optimizer. 
In this work we used the implementation of Adam provided in TensorFlow 2.0.0.
This pretraining concept is similar to the pretraining algorithm presented in~\cite{kammonen2022deepVSshallow}. The difference here is that we did not train deep residual neural networks  
and we applied Algorithm~\ref{alg:AMRS_E} for the pretraining instead of the ARFF algorithm. When training using the Adam optimizer we employed the mean squared error (MSE) loss without an L2-regularizer.

Test~\ref{test:pre_train} concerns~\eqref{eq:reg_disc_data_set}, but this time with 
the trivial rotation matrix: 
\begin{align}
	\label{eq:rot_mat_II}
	B & = \ID =
	\begin{bmatrix} 
		1 &   0 &  0 &  0 \\
        0 &   1 &  0 &  0 \\
        0 &   0 &  1 &  0 \\
        0 &   0 &  0 &  1
	\end{bmatrix}.
\end{align}
Furthermore the complex exponential activation function was replaced by the cosine activation function in this experiment.
Thus, the neural network is defined as the parameterized function 
$\beta_{\boldsymbol{\theta}}: \R^d \rightarrow \R$, given by
\begin{equation}
    \label{eq:NN-Def_cos}
    \beta_{\boldsymbol{\theta}}(\xVector) = 
    \sum_{k = 1}^{K} \ampl_{k}\cos(\check{\omegaVector}_k^T\check{\xVector}),
\end{equation}
with the extended frequency and data vectors $\check{\omegaVector}_k := (\omegaVector_k, b_k)^T$ and $\check{\xVector}_m := (\xVector^T_m, 1)^T$.
With the change to a real-valued activation function, real-valued amplitudes $\ampl_k$ suffice to approximate
real-valued target functions. The bias $b_k$ must be added to approximate non-even functions. 
Thus, the number of degrees of freedom in the minimization problem~\eqref{eq:NonConvexOptProblem} remains the same, whereas the 
parameter space is changed to $\boldsymbol{\Theta}_{\ampl} \times\boldsymbol{\Theta}_{\check{\omegaVector}}=\R^K\times\R^{(d+1)\times K}$ (from $\C^K\times\R^{d\times K}$).
The linear system~\eqref{eq:num_normal_eq} is changed to using the matrix $\BARS\in\R^{M_B\times K}$ with elements 
$\BARS_{j,k} = \cos(\check{\omegaVector}_k^T\check{\xVector}_{m_j})$ when the neural network~\eqref{eq:NN-Def} is replaced by~\eqref{eq:NN-Def_cos}. 
With this change in~\eqref{eq:num_normal_eq}, Algorithm~\ref{alg:AMRS_E} can be employed with the cosine activation function, 
operating on the extended vectors $\check{\omegaVector}_k$ and $\check{\xVector}_m$.
The adjusted Algorithm~\ref{alg:AMRS_E} was applied, as described, as a stand-alone training algorithm 
and a pretrainer for the Adam optimizer in Test~\ref{test:pre_train} with the parameters in Tables~\ref{tab:Alg_defs_2} and~\ref{tab:pre_training}.

\begin{table}[ht]
	\centering
	\begin{tabular}{|c|l|c|c|}
	\hline
	& & $R(n)$ & $A(n)$ \\ 
	\hline
	\algAM & Adaptive Metropolis & 0 & \textbf{true} \\ 
    \algAMRmod & Adaptive Metropolis with Resampling & 1 & \textbf{true} \\ 
    \algRWR & Random Walk with Resampling & 1 & \textbf{false} \\ 
	\hline
	\end{tabular}
	\caption{Flow control parameters in Algorithm~\ref{alg:AMRS_E} defining the three versions, \algAM, \algAMRmod, and \algRWR, compared in Test~\ref{test:pre_train}.}
	\label{tab:Alg_defs_2}
\end{table}

\begin{table}[ht]
	\centering
	\begin{tabular}{|c|c|c|c|c|c|c|c|c|}
	\hline
	Problem & $K$ & $M$ & $N$ & $\delta$ & $\gamma$ & $M_B$ & $\lambda$ & $\check{\omegaVector}^0$ \\ 
	\hline
    \multirow{2}{*}{\eqref{eq:NN-Def_cos},\eqref{eq:NonConvexOptProblem},\eqref{eq:reg_disc_data_set},\eqref{eq:rot_mat_II}} &
	\multirow{2}{*}{1024} & 
	\multirow{2}{*}{$10^6$} &
	300 &
	\multirow{2}{*}{$2^{-2}$} & 
	\multirow{2}{*}{$10$} &
	\multirow{2}{*}{$10^4$} &
	\multirow{2}{*}{$0.1$} &
	\multirow{2}{*}{$\mathbf{0}\in\R^{d+1}$} \\
	 & & & $1.2\cdot10^4$ & & & & & \\
	\hline
	\end{tabular}
	\caption{Algorithm parameters corresponding to Algorithm~\ref{alg:AMRS_E}, 
        as used in Test~\ref{test:pre_train} (Figure~\ref{fig:f1}). 
        The number of iterations $N=300$ refers to the use as a pretrainer, and the higher number refers to the use as a stand-alone training algorithm.
        The choice of activation function in~\eqref{eq:NN-Def_cos} is no longer the complex exponential. 
        }
	\label{tab:pre_training}
\end{table}

\paragraph{\textbf{Test~\testnr{test:pre_train}:} Accelerating the Adam optimizer by pretraining with Algorithm~\ref{alg:AMRS_E}}

We normalized
the data-set $\{(\xRV_m, Y_m)\}_{m=1}^M$ by subtracting the mean and dividing by the standard deviation component-wise. Let
\begin{equation*}
\{(\bar \xVector_m, \bar y_m)\}_{m=1}^M
\end{equation*}
denote the normalized data.
In the experiments we used the normalized data-set $\{(\bar \xVector_m, \bar y_m)\}_{m=1}^M$ with $M = 10^6$ training data points and $10^4$ validation and testing data points each. The two-layer neural network has $K = 1024$ nodes.
As comparisons to the pretraining approach we also performed training with only the Adam optimizer and with only Algorithm~\ref{alg:AMRS_E} for various parameters. 

When training and pretraining with Algorithm~\ref{alg:AMRS_E}, we set the parameter choices to $\delta = 1/4$, $M_B = 10^4$, $\lambda = 0.1$, and $\gamma = 3d - 2 = 10$. 
When pretraining we performed $N = 300$ iterations, and when training, we performed $N = 12,000$ iterations. 
Pretraining was conducted with the resampling rule $R \equiv 1$ and the adaptive Metropolis rule \textbf{true}. The comparisons when training with Algorithm~\ref{alg:AMRS_E} and not following with the Adam optimizer 
were performed for the resampling and adaptive Metropolis rule combinations $(R, A) = (1, \textbf{false}), (0, \textbf{true}), \text{ and } (1, \textbf{true})$. 
As an example, $(R,A) = (1, \textbf{false})$ indicates that we employ resampling in each iteration but not perform the Metropolis test.

\begin{figure}[ht]
	\centering
    \includegraphics[width=0.67\linewidth]{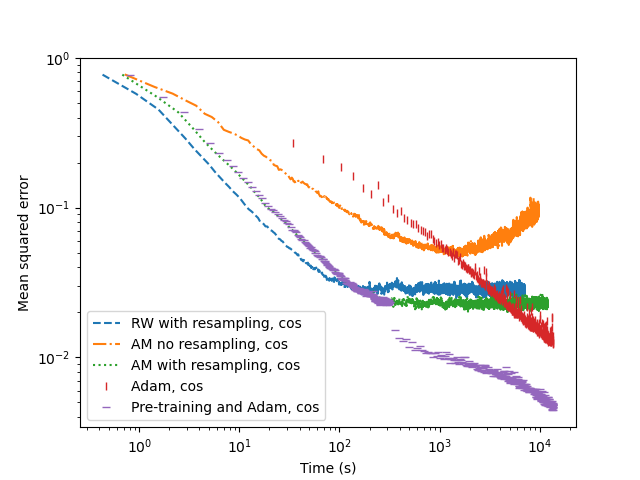}
    \caption{Test~\ref{test:pre_train} (i.e.,~\eqref{eq:reg_disc_data_set} with $B$ in~\eqref{eq:rot_mat_II}
        and the real-valued neural network~\eqref{eq:NN-Def_cos})
        illustrating Algorithm~\ref{alg:AMRS_E} as a pretrainer for the Adam optimizer. 
        The validation loss is displayed as a function of the actual time. Pretraining is conducted with Algorithm~\ref{alg:AMRS_E} with $(R,A)=(1,\textbf{true})$. 
        The dashed blue line marks the validation loss when training with Algorithm~\ref{alg:AMRS_E} is for $(R,A)=(1,\textbf{false})$, 
        the dashed and dotted orange line for $(R,A)=(0,\textbf{true})$, and 
        the dotted green line is for $(R,A)=(1,\textbf{true})$.}
        \label{fig:f1}
\end{figure}

We used learning rate $0.0005$, batch size $128$, and $400$ epochs when training with Adam alone and also when training with Adam after pre-training with Algorithm~\ref{alg:AMRS_E}.
The remaining hyperparameters for the Adam optimizer were set to the default values given in TensorFlow 2.0.0. 
Figure~\ref{fig:f1} presents the validation error with respect to the actual time.
The computations were performed on a 2019 MacBook Pro, with a 2.4 GHz, 8-core Intel Core~i9 CPU and 64 GB of 2667 MHz DDR4 RAM, using the MacOS Sonoma~14.5 operating system.

In Figure~\ref{fig:f1}, the Adam optimizer has a lower validation error than training only with Algorithm~\ref{alg:AMRS_E}, but that is not necessarily true when pretraining with Algorithm~\ref{alg:AMRS_E} before training with the Adam optimizer, 
which can also result in a faster total convergence. For an equal number of iterations, the random walk with resampling method, $(R,A) = (1, \textbf{false})$, takes less time than training with the standard ARFF algorithm, $(R,A) = (0, \textbf{true})$, 
which also takes less time than including both the Metropolis test and resampling step, $(R,A) = (1, \textbf{true})$.

\subsection{Coordinate-based Multilayer Perceptrons}
\label{sec:num_exp_image}

In image regression problems, an Multilayer Perceptron (MLP) can be trained to provide a continuous representation of an image. The MLP takes pixel coordinates, $\xVector$, and maps them to $(R,G,B)$ color data, $\boldsymbol{y}$.
The idea of passing data through an RFF layer before the MLP when performing image regression first appeared in~\cite{Tancik2020FourierFL} where they demonstrated that MLPs can learn high-frequency content using an RFF layer. In their experiments, they sampled the frequencies for the RFF layer from a normal distribution and manually varied the standard deviation. In contrast, in this paper, we followed the idea of passing data through an RFF layer, but instead of manually adjusting the optimal sampling distribution, we used Algorithm~\ref{alg:AMRS_E} to sample the frequencies.

The proposed method is simple and has the following two steps:
\begin{description}
    \item[Step 1. ] \textbf{Frequency sampling:} Train a two-layer neural network on an image with Algorithm~\ref{alg:AMRS_E} and keep the frequencies from the trained neural network.
    \item[Step 2. ] \textbf{MLP training:} Train an MLP with the Adam optimizer (for example) on the same image as in Step 1, where the first hidden layer is an RFF layer with the frequencies initialized to those frequencies received in Step 1.
\end{description}

As in Section~\ref{sec:num_exp_pre_training}, we used the cosine activation function (see the description of how  Algorithm~\ref{alg:AMRS_E} is adjusted for this in that section).
Algorithm~\ref{alg:AMRS_E} treats one-dimensional amplitudes $a\in \mathbb{R}$, but for the image regression case, the amplitudes are in $\mathbb{R}^3$. To address this discrepancy, we use the 2-norm instead of the absolute value of the amplitudes where necessary in the algorithm.

The images used in our experiments are $92$ random samples from the DIV2K dataset, see~\cite{div2k} and~\cite{Timofte_2018_CVPR_Workshops}. We normalized each pixel value to be between $0$ and $1$ and cropped each image to the centered $512\times 512$ pixels of the original images and used one-fourth for training and another fourth for testing. The training data were taken as all pixels with both coordinates as
even numbers, and the testing data were taken such that both coordinates were odd numbers. The training data for each image can be expressed as follows: $\{(\xVector_m,\boldsymbol{y}_m)\}_{m=1}^{65536}\subset[0,1]^2\times[0,1]^3$.

As for Step 1, we trained a two-layer neural network with Algorithm~\ref{alg:AMRS_E} on the corresponding training data $\{(\xVector_m, \boldsymbol{y}_m)\}_{m=1}^{M}$ for each image and set the frequencies and biases to $\{(\omegaVector_k, b_k)\}_{k=1}^K$. The parameter choices were $K = 256$, $N = 20$, $\lambda = 10^{-4}$, $\delta = 1$, and $M_B = M = 65536$. The adaptive Metropolis rule $A$ in Algorithm~\ref{alg:AMRS_E} was set to \textbf{false}, and the resampling rule $R$ was set to $1$. We initialized the weights and biases $\{(\omegaVector_k, b_k)\}_{k=1}^K$ to zeros.

In Step 2, we used the Adam optimizer to train the MLP $\beta : \mathbb{R}^{2} \to [0,1]^{3}$ defined as follows:
\begin{align*}
	\beta & = \sigma_L \circ h_L \circ \sigma_{L-1} \circ h_{L-1} \circ \dots \circ \sigma_1 \circ h_1,
\end{align*}
with the activation function applied componentwise as follows:
\begin{align*}
	[\sigma_1(\xVector)]_k & = \cos(x_k), && k=1,\dots,K,\\
	[\sigma_\ell(\xVector)]_k &= \max(0,x_k), && \ell=2,\dots,L-1,\;k=1,\dots,K\\
    [\sigma_L(\xVector)]_i &= \frac{1}{1 + e^{-x_i}}, && i=1,2,3,\\
\end{align*}
with the following affine mappings:
\begin{align*}
	h_\ell\left(\xVector;\omegaVector^\ell,\boldsymbol{b}^\ell\right) & = \omegaVector^\ell\xVector + \boldsymbol{b}^\ell, && \ell=1,\dots,L,\quad \boldsymbol{b}^L=\boldsymbol{0}\in\rset^3,\\
    \omegaVector^1 &\in\rset^{K\times 2},\\
    \omegaVector^\ell &\in\rset^{K\times K}, && \ell=2,\dots,L-1,\\
    \omegaVector^L &\in\rset^{3\times K},\\
    \boldsymbol{b}^\ell &\in\rset^{K}, && \ell=1,\dots,L-1.
\end{align*}
We initialized $\omegaVector^1$ and $\boldsymbol{b}^1$ to the corresponding values of $\{(\omegaVector_k, b_k)\}_{k=1}^K$ attained in Step 1. The weights and biases $\omegaVector^\ell, \ell=2,\dots,L$, and $\boldsymbol{b}^\ell, \ell=2,\dots,L-1$ are initialized using the Glorot normal initializer. The loss function is the mean squared loss, and we did not include any L2 regularizer when training with the Adam optimizer.

As a comparison, instead of sampling the weights and biases of the RFF layer with Algorithm~\ref{alg:AMRS_E}, we initialized them using the Glorot normal initializer. In another comparison, we replaced the RFF layer with another ReLU layer. In a third comparison, we completely removed the RFF layer. When training with the Adam optimizer, in all experiments, we set $L = 4$, $K = K_{\ell} = 256, \ell = 2,3,4$, the learning rate to $10^{-3}$, the batch size to $256,$ and the epochs $2,000$ epochs. The training was performed in TensorFlow 2.0.0 with the default values for the parameters used by the Adam optimizer.

In summary, we presented four different approaches to the image regression problem.
\begin{enumerate}
    \item Sample the frequencies and biases with Algorithm~\ref{alg:AMRS_E} for the RFF layer in an MLP and train the MLP with the Adam optimizer. Besides the RFF layer, the MLP has three hidden ReLU layers.
    \item Use the same neural network structure and Adam optimizer training as in the first approach, but instead of sampling the frequencies and biases using Algorithm~\ref{alg:AMRS_E}, the RFF layer is initialized as any other layer of the MLP with Glorot.
    \item A three-hidden-layer MLP trained using the Adam optimizer is employed without an RFF layer.
    \item A four-hidden-layer MLP trained using the Adam optimizer is employed without an RFF layer.
\end{enumerate}

The mean squared error between the MLP image approximation and the original image is denoted as MSE. Additionally, we define $\text{MAX}_I$ as the maximum pixel intensity value present in the original image.
Then, the peak signal-to-noise ratio (PSNR) is defined as $\text{PSNR} := 10\log_{10}\left(\frac{\text{MAX}_I^2}{MSE}\right)$. After training, we computed the PSNR for the neural networks evaluated in the $65,536$ testing points. Table~\ref{tab:psnr_summary} summarizes the mean, max, and min PSNR values,  with the standard deviations of the PSNR values for each approach using $92$ images. 
We concluded that, for the experiments, the first approach gives the largest PSNR, considering the mean, max and min values.

\begin{figure}[!tbp]
  \begin{subfigure}[t]{0.49\textwidth}
	\centering
	\includegraphics[width=0.85\linewidth]{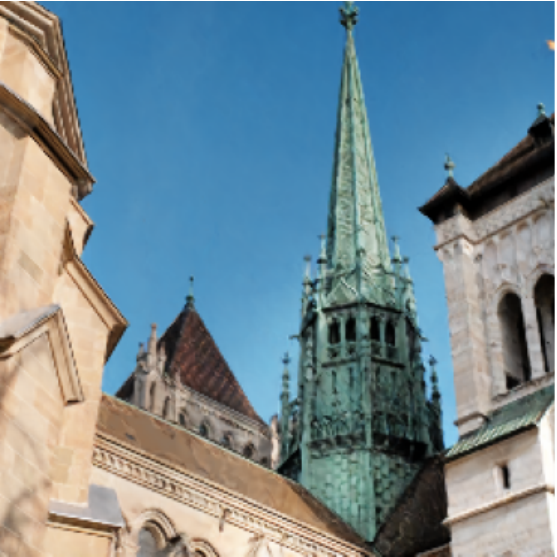}%
  \caption{RFF layer initialized using Algorithm~\ref{alg:AMRS_E} with three hidden ReLU layers.}
  \end{subfigure}
  \hfill
  \begin{subfigure}[t]{0.49\textwidth}
	\centering
	\includegraphics[width=0.85\linewidth]{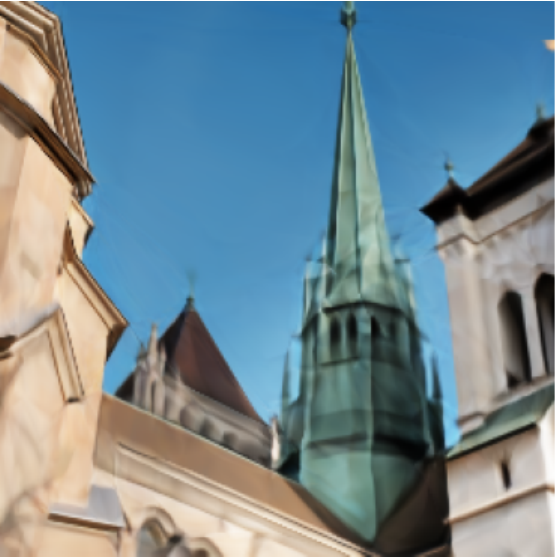}
  \caption{No RFF layer with four hidden ReLU layers.}
  \end{subfigure}
  \begin{subfigure}[b]{0.49\textwidth}
	\centering
	\includegraphics[width=0.85\linewidth]{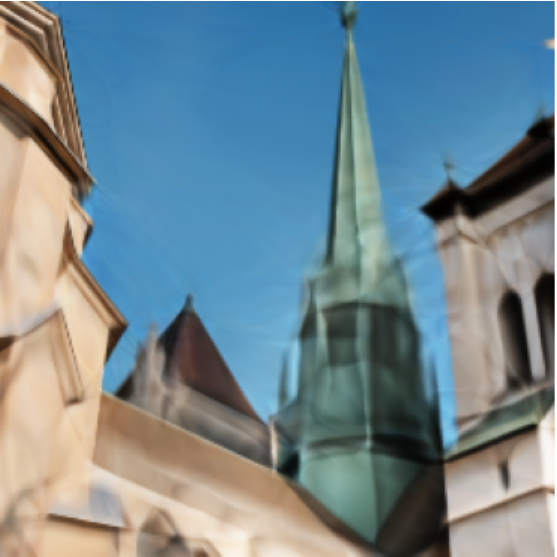}%
  \caption{No RFF layer with three hidden ReLU layers.}
  \end{subfigure}
  \hfill
  \begin{subfigure}[b]{0.49\textwidth}
	\centering
	\includegraphics[width=0.85\linewidth]{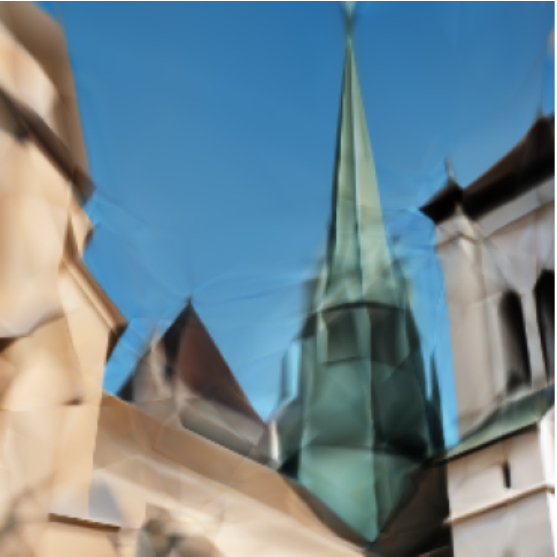}%
  \caption{RFF layer initialized using Xavier.\\~}
  \end{subfigure}
  \caption{Neural network representations of an image from the DIV2K dataset.}
  \label{fig:grid_image_regression_0756}
\end{figure}

\begin{figure}[!tbp]
  \begin{subfigure}[t]{0.49\textwidth}
	\centering
	\includegraphics[width=\linewidth]{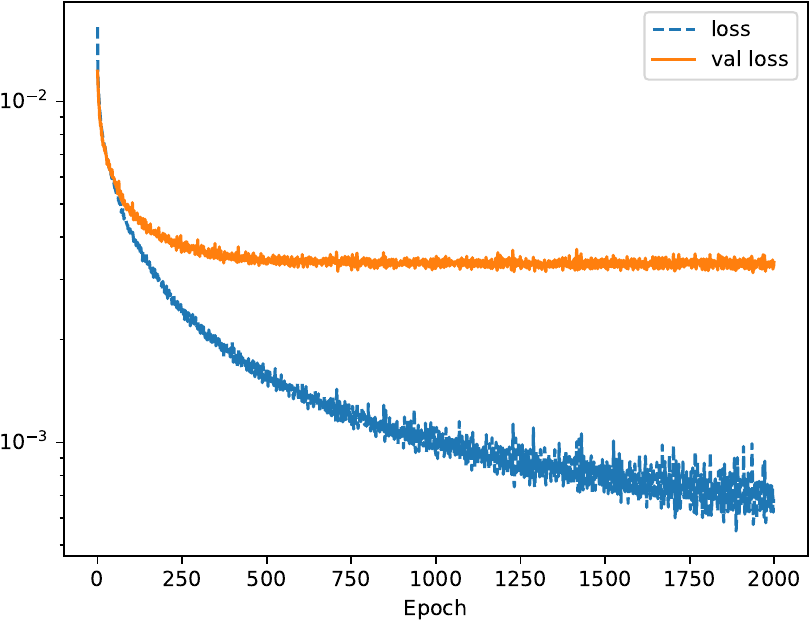}%
    \caption{RFF layer initialized using Algorithm~\ref{alg:AMRS_E} with three hidden ReLU layers.}
  \end{subfigure}
  \hfill
  \begin{subfigure}[t]{0.49\textwidth}
	\centering
	\includegraphics[width=\linewidth]{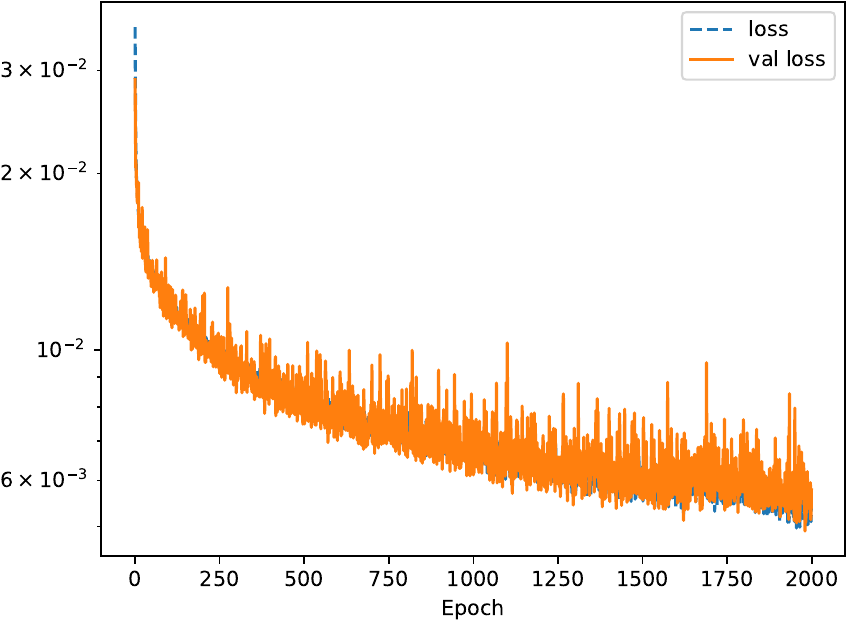}
    \caption{No RFF layer with four hidden ReLU layers.}
  \end{subfigure}
  \begin{subfigure}[b]{0.49\textwidth}
	\centering
	\includegraphics[width=\linewidth]{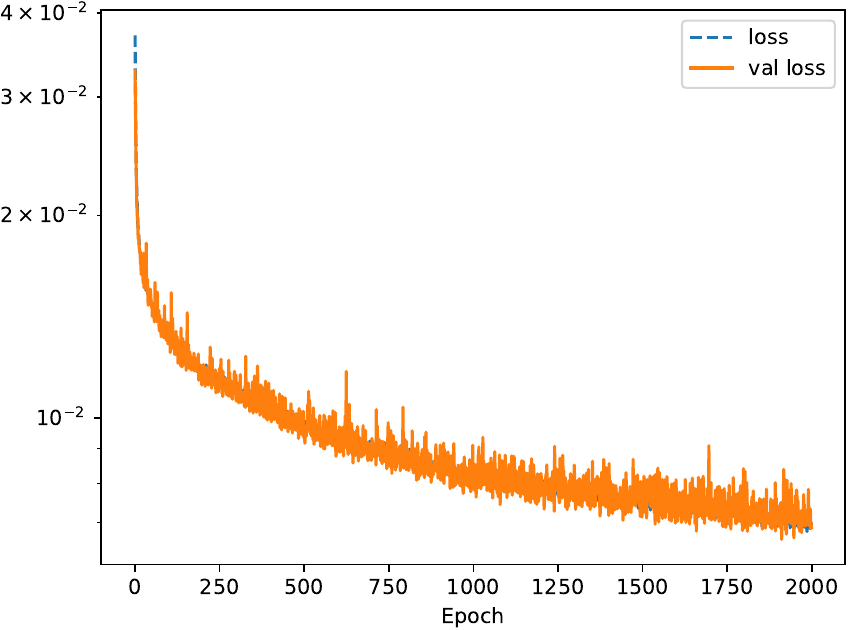}%
    \caption{No RFF layer with three hidden ReLU layers.}
  \end{subfigure}
  \hfill
  \begin{subfigure}[b]{0.49\textwidth}
	\centering
	\includegraphics[width=\linewidth]{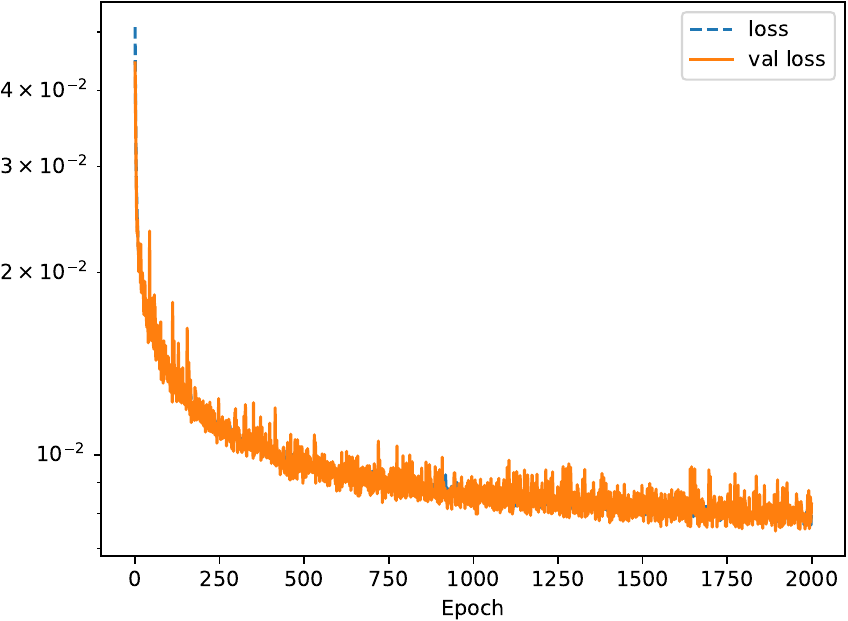}%
    \caption{RFF layer initialized using Xavier.\\~}
  \end{subfigure}
  \caption{Training and validation errors with respect to the number of epochs when training on an image from the DIV2K dataset for different neural network structures and/or initialization. Figure~\ref{fig:grid_image_regression_0756} present the resulting images.}
  \label{fig:grid_image_regression_error_0756}
\end{figure}

\begin{figure}[!tbp]
  \begin{subfigure}[t]{0.49\textwidth}
	\centering
	\includegraphics[width=0.85\linewidth]{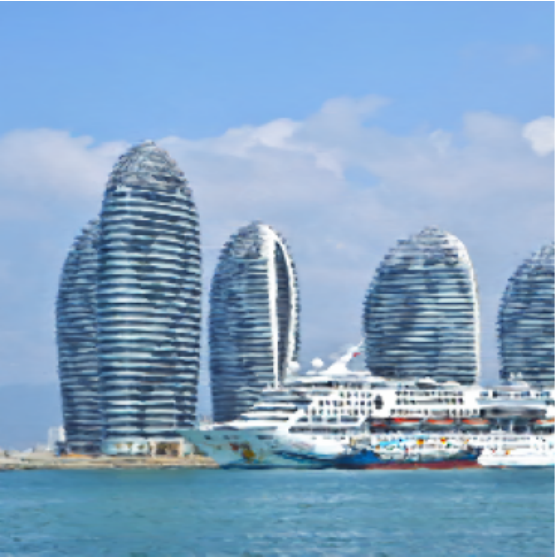}%
    \caption{RFF layer initialized using Algorithm~\ref{alg:AMRS_E} with three hidden ReLU layers.}
  \end{subfigure}
  \hfill
  \begin{subfigure}[t]{0.49\textwidth}
	\centering
	\includegraphics[width=0.85\linewidth]{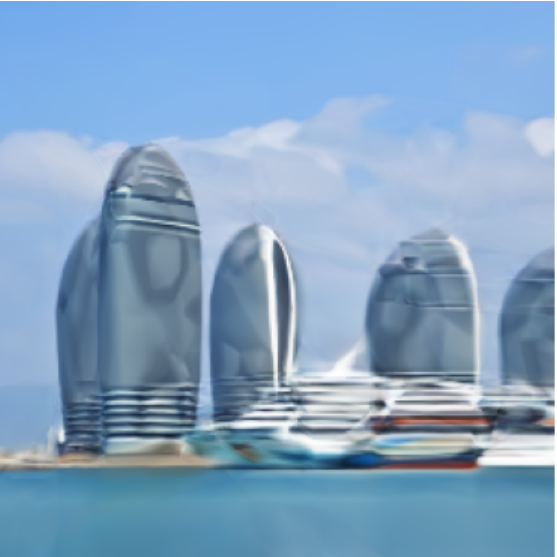}
    \caption{No RFF layer with four hidden ReLU layers.}
  \end{subfigure}
  \begin{subfigure}[b]{0.49\textwidth}
	\centering
	\includegraphics[width=0.85\linewidth]{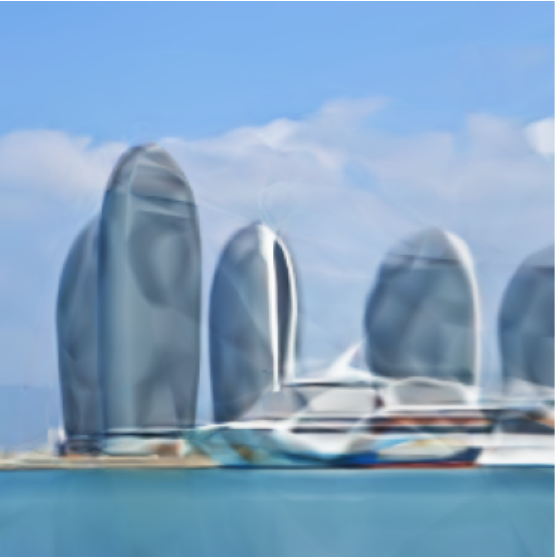}%
    \caption{No RFF layer with three hidden ReLU layers.}
  \end{subfigure}
  \hfill
  \begin{subfigure}[b]{0.49\textwidth}
	\centering
	\includegraphics[width=0.85\linewidth]{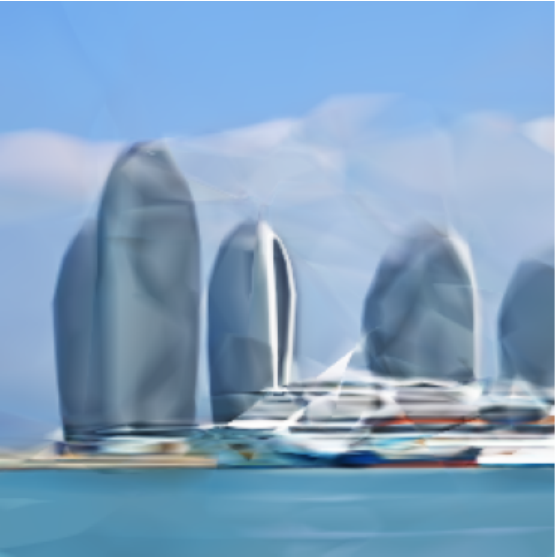}%
    \caption{RFF layer initialized using Xavier.\\~}
  \end{subfigure}
  \caption{Neural network representations of an image from the DIV2K dataset.}
  \label{fig:grid_image_regression_0080}
\end{figure}

\begin{figure}[!tbp]
  \begin{subfigure}[t]{0.49\textwidth}
	\centering
	\includegraphics[width=\linewidth]{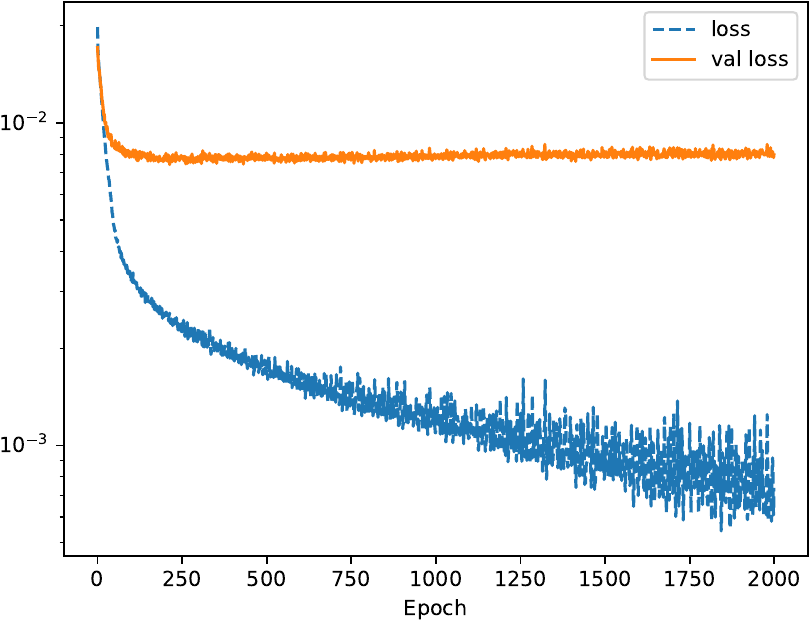}%
  \caption{RFF layer initialized using Algorithm~\ref{alg:AMRS_E} with three hidden ReLU layers.}
  \end{subfigure}
  \hfill
  \begin{subfigure}[t]{0.49\textwidth}
	\centering
	\includegraphics[width=\linewidth]{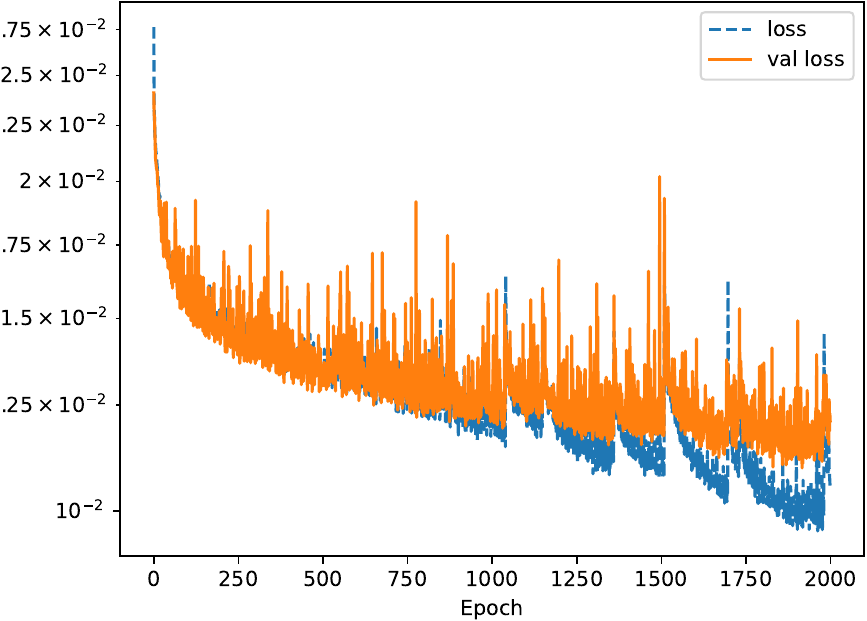}
  \caption{No RFF layer with four hidden ReLU layers.}
  \end{subfigure}
  \begin{subfigure}[b]{0.49\textwidth}
	\centering
	\includegraphics[width=\linewidth]{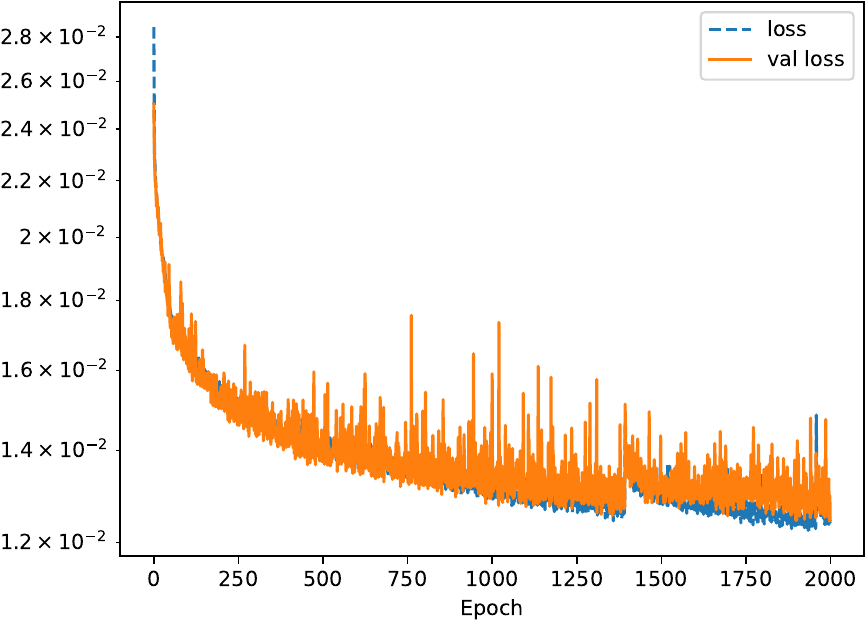}%
  \caption{No RFF layer with three hidden ReLU layers.}
  \end{subfigure}
  \hfill
  \begin{subfigure}[b]{0.49\textwidth}
	\centering
	\includegraphics[width=\linewidth]{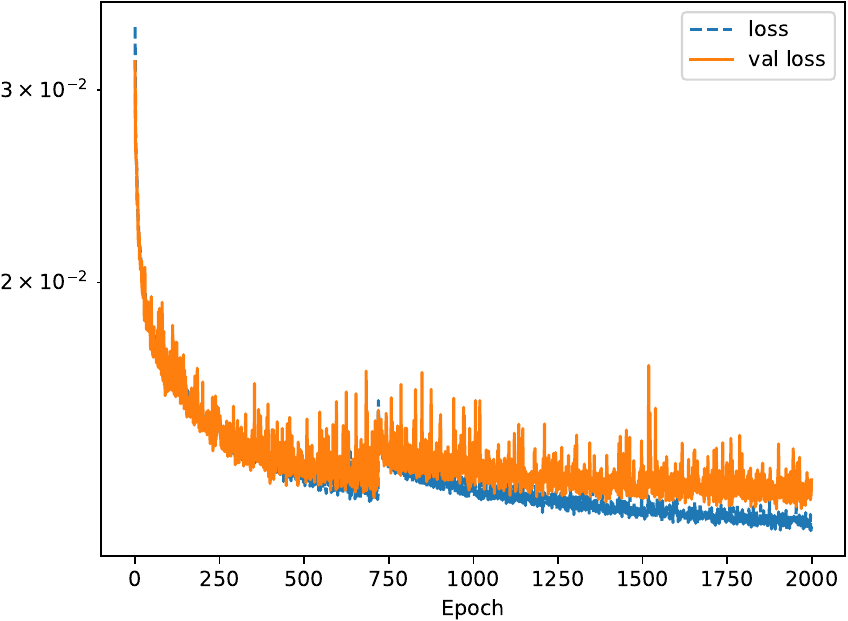}%
  \caption{RFF layer initialized using Xavier.\\~}
  \end{subfigure}
  \caption{Training and validation errors with respect to the number of epochs when training on an image from the DIV2K dataset for different neural network structures and/or initialization; Figure~\ref{fig:grid_image_regression_0080} presents the resulting images.}
  \label{fig:grid_image_regression_error_0080}
\end{figure}

\begin{figure}[!tbp]
  \begin{subfigure}[t]{0.49\textwidth}
	\centering
	\includegraphics[width=0.85\linewidth]{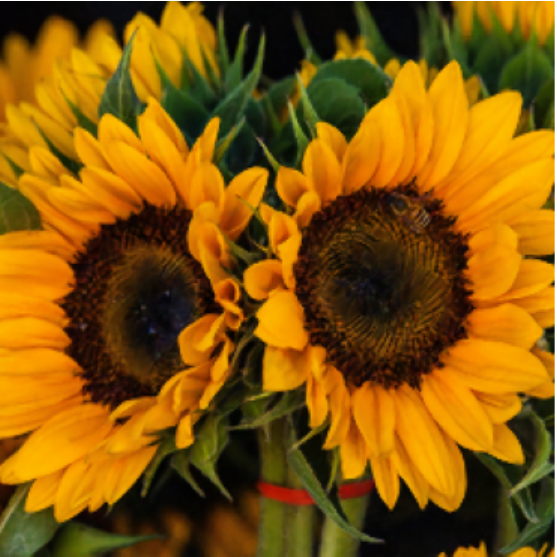}%
    \caption{RFF layer initialized with Algorithm~\ref{alg:AMRS_E} with three hidden ReLU layers.}
  \end{subfigure}
   \hfill
  \begin{subfigure}[t]{0.49\textwidth}
	\centering
	\includegraphics[width=0.85\linewidth]{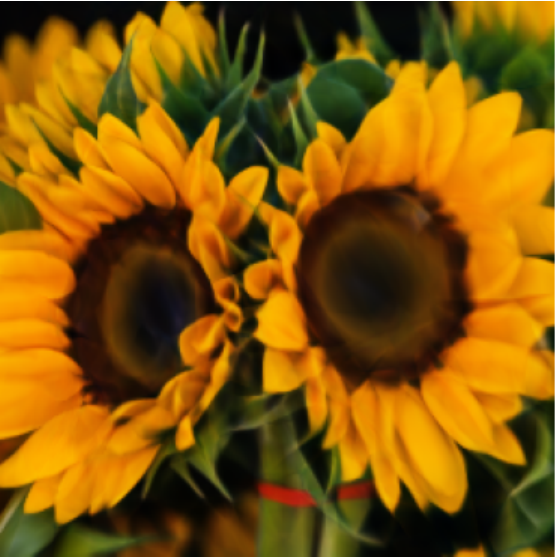}
    \caption{No RFF layer with four hidden ReLU layers.}
  \end{subfigure}
  \begin{subfigure}[b]{0.49\textwidth}
	\centering
	\includegraphics[width=0.85\linewidth]{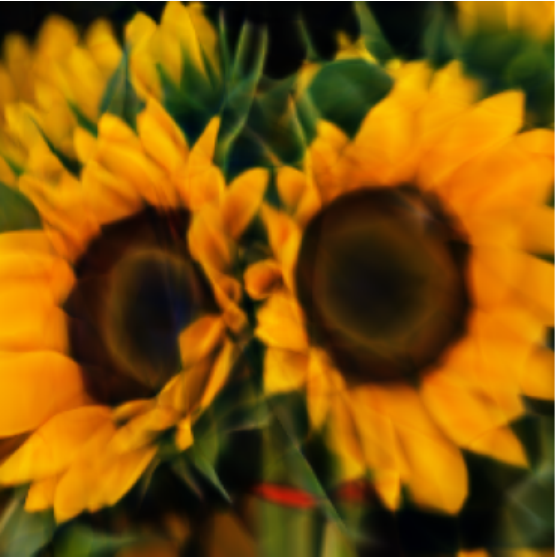}%
    \caption{No RFF layer with three hidden ReLU layers.}
  \end{subfigure}
  \hfill
  \begin{subfigure}[b]{0.49\textwidth}
	\centering
	\includegraphics[width=0.85\linewidth]{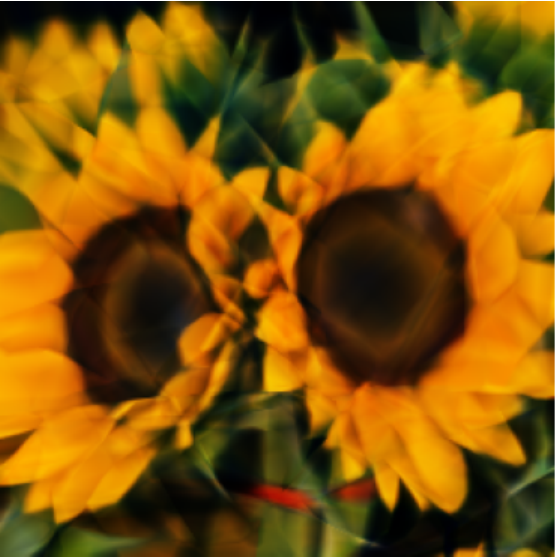}%
    \caption{RFF layer initialized using Xavier.\\~}
  \end{subfigure}
  \caption{Neural network representations of an image from the DIV2K dataset.}
  \label{fig:grid_image_regression_0746}
\end{figure}

\begin{figure}[!tbp]
  \begin{subfigure}[t]{0.49\textwidth}
	\centering
	\includegraphics[width=\linewidth]{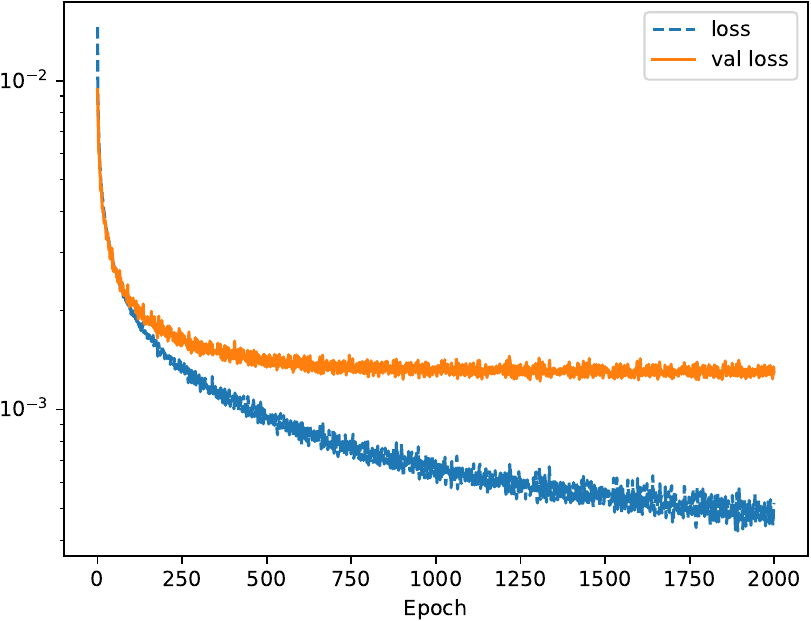}%
    \caption{RFF layer initialized using Algorithm~\ref{alg:AMRS_E} with three hidden ReLU layers.}
  \end{subfigure}
  \hfill
  \begin{subfigure}[t]{0.49\textwidth}
	\centering
	\includegraphics[width=\linewidth]{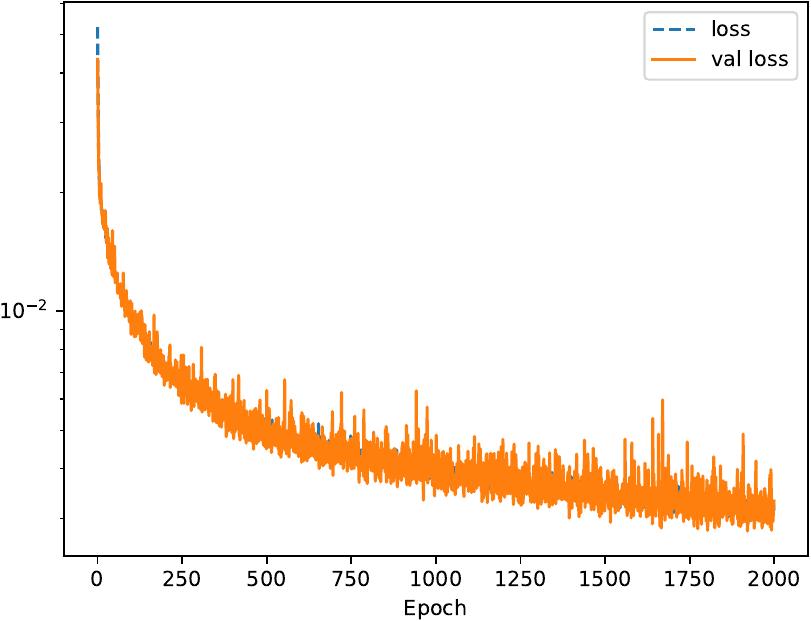}
    \caption{No RFF layer with four hidden ReLU layers.}
  \end{subfigure}
  \begin{subfigure}[b]{0.49\textwidth}
	\centering
	\includegraphics[width=\linewidth]{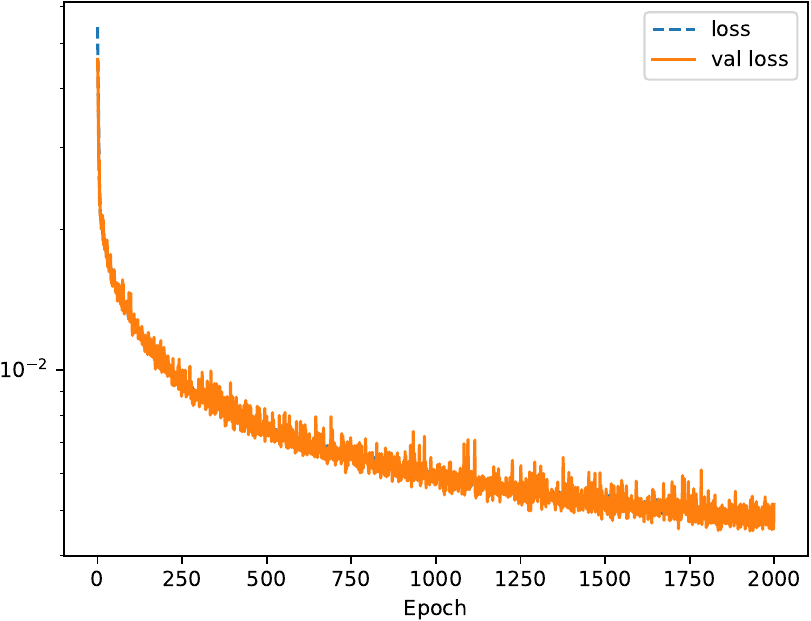}%
    \caption{No RFF layer with three hidden ReLU layers.}
  \end{subfigure}
  \hfill
  \begin{subfigure}[b]{0.49\textwidth}
	\centering
	\includegraphics[width=\linewidth]{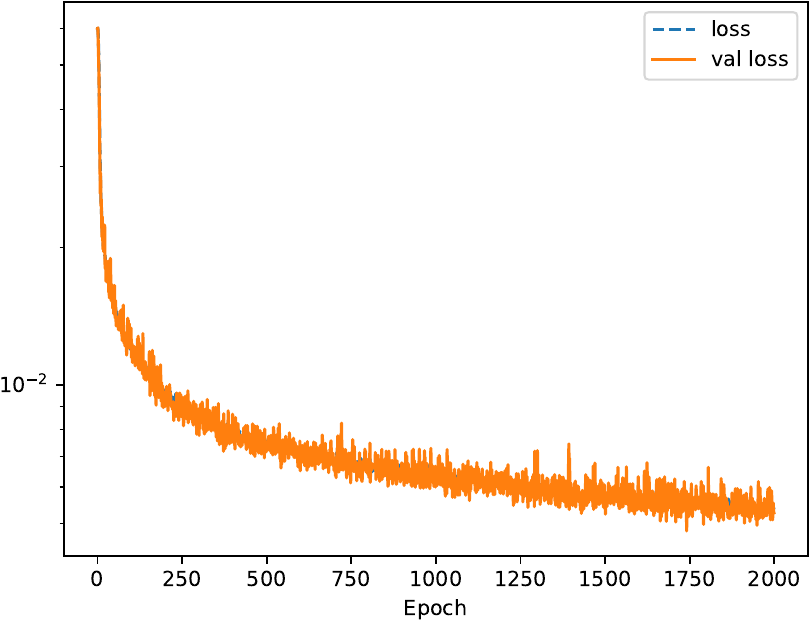}%
    \caption{RFF layer initialized using Xavier.\\~}
  \end{subfigure}
  \caption{Training and validation errors with respect to the number of epochs when training on an image from the DIV2K dataset for different neural network structures and/or initialization; Figure~\ref{fig:grid_image_regression_0746} present the resulting images.}
  \label{fig:grid_image_regression_error_0746}
\end{figure}

\begin{figure}[!tbp]
  \begin{subfigure}[t]{0.49\textwidth}
	\centering
	\includegraphics[width=0.85\linewidth]{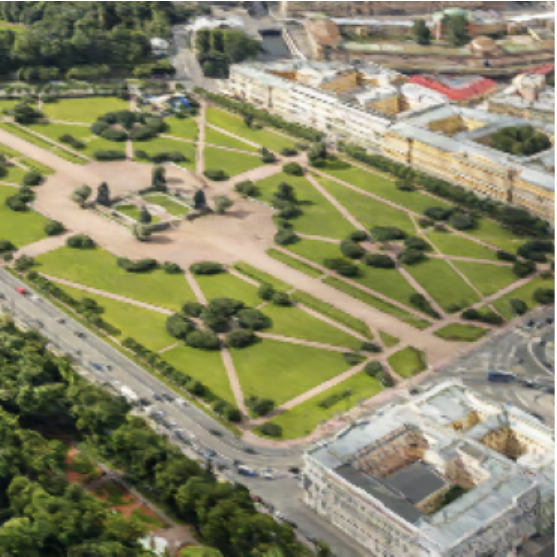}%
  	\caption{RFF layer initialized using Algorithm~\ref{alg:AMRS_E} with three hidden ReLU layers.}
  \end{subfigure}
  \hfill
  \begin{subfigure}[t]{0.49\textwidth}
	\centering
	\includegraphics[width=0.85\linewidth]{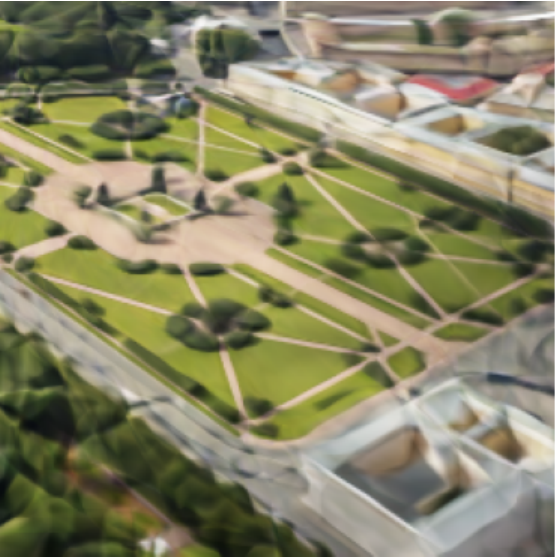}
    \caption{No RFF layer with four hidden ReLU layers.}
  \end{subfigure}
  \begin{subfigure}[b]{0.49\textwidth}
	\centering
	\includegraphics[width=0.85\linewidth]{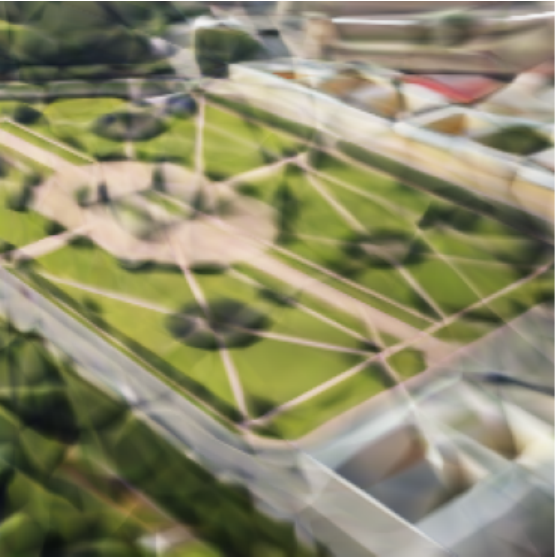}%
    \caption{No RFF layer with three hidden ReLU layers.}
  \end{subfigure}
  \hfill
  \begin{subfigure}[b]{0.5\textwidth}
	\centering
	\includegraphics[width=0.85\linewidth]{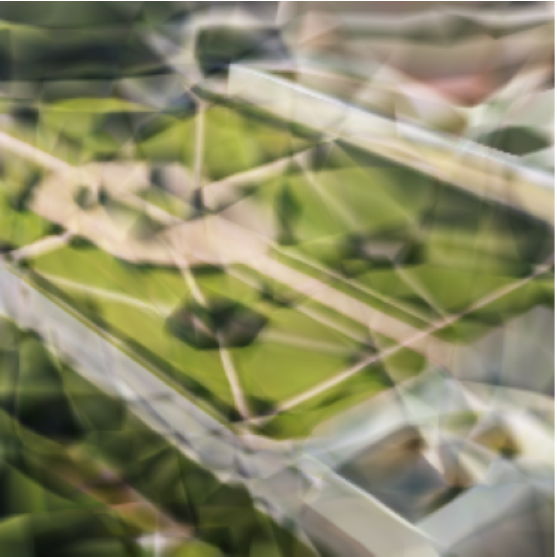}%
    \caption{RFF layer initialized using Xavier.\\~}
  \end{subfigure}
  \begin{center}
    \begin{subfigure}[b]{0.49\textwidth}
	  \centering
	  \includegraphics[width=\linewidth]{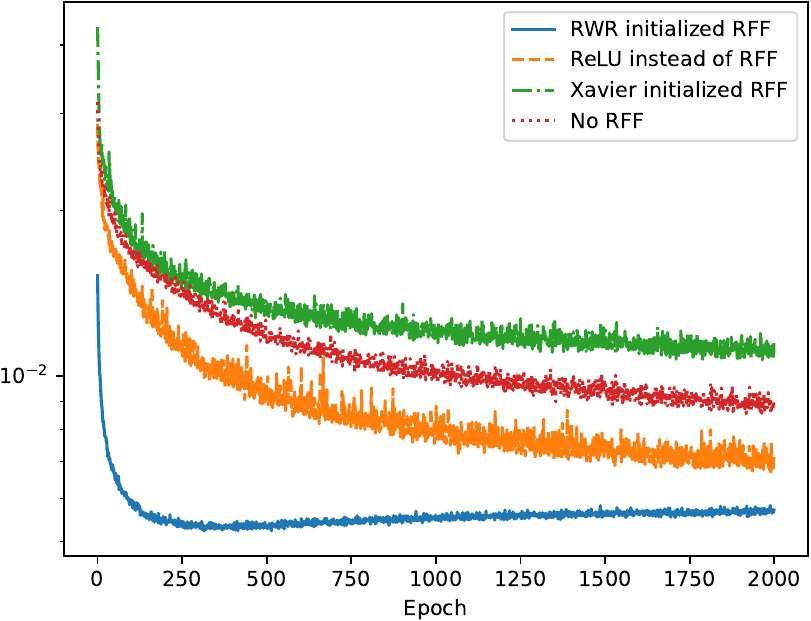}%
      \caption{Validation error with respect to epochs for various training algorithms}
    \end{subfigure}
  \end{center}
  \caption{Neural network representations of an image from the DIV2K dataset.}
  \label{fig:grid_image_regression_0078}
\end{figure}

\begin{figure}[!tbp]
  \begin{subfigure}[t]{0.49\textwidth}
	\centering
	\includegraphics[width=\linewidth]{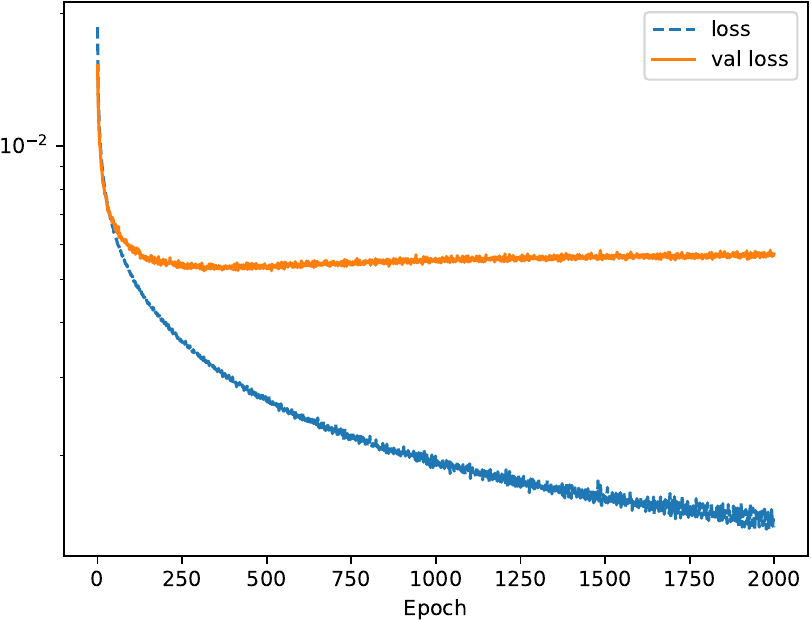}%
    \caption{RFF layer initialized using Algorithm~\ref{alg:AMRS_E} with three hidden ReLU layers.}
  \end{subfigure}
  \hfill
  \begin{subfigure}[t]{0.49\textwidth}
	\centering
	\includegraphics[width=\linewidth]{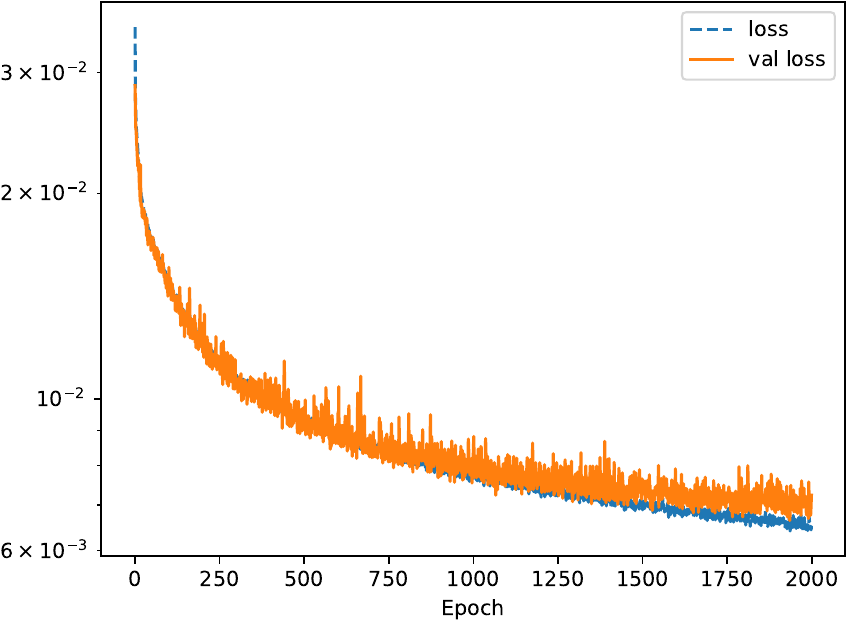}
    \caption{No RFF layer with four hidden ReLU layers.}
  \end{subfigure}
  \begin{subfigure}[b]{0.49\textwidth}
	\centering
	\includegraphics[width=\linewidth]{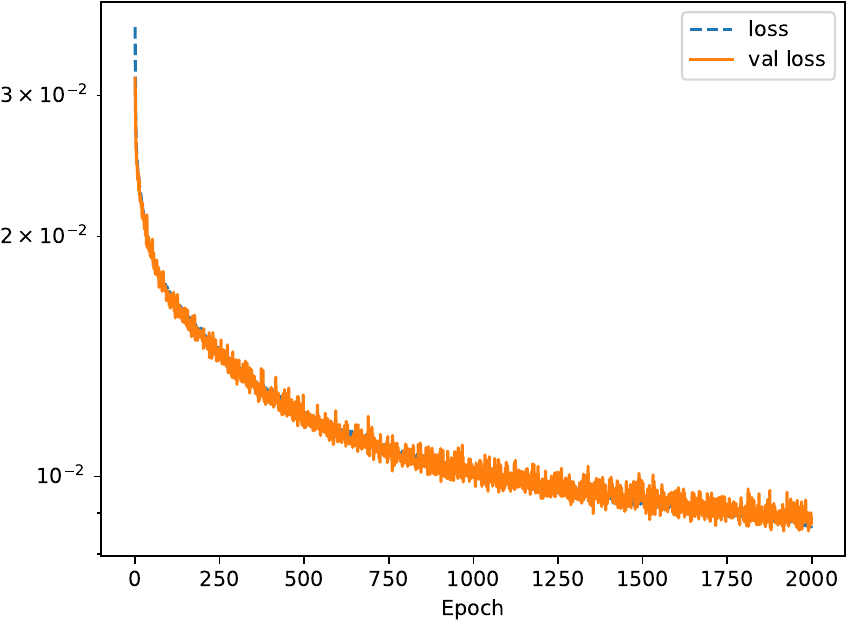}%
    \caption{No RFF layer with three hidden ReLU layers.}
  \end{subfigure}
  \hfill
  \begin{subfigure}[b]{0.5\textwidth}
	\centering
	\includegraphics[width=\linewidth]{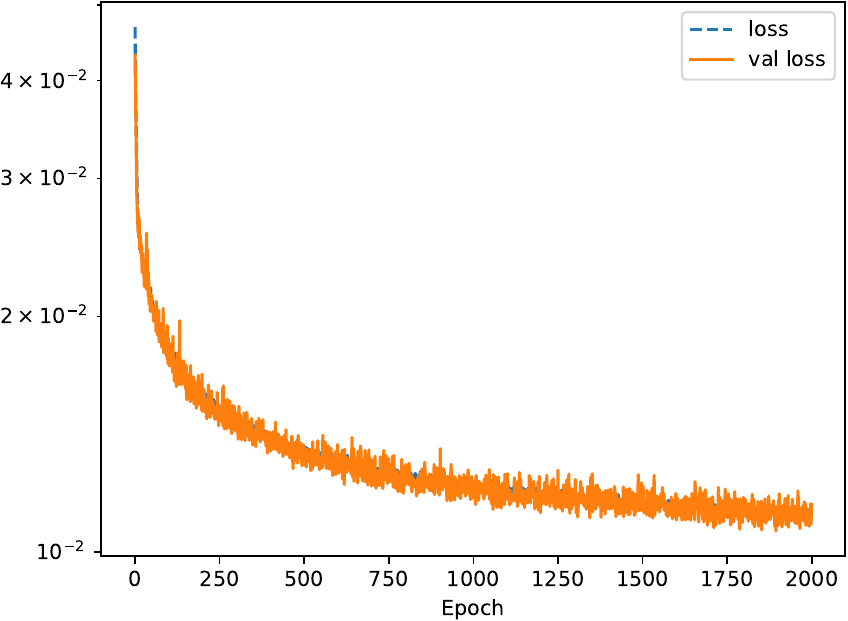}%
    \caption{RFF layer initialized using Xavier.\\~}
  \end{subfigure}
  \caption{Training and validation errors with respect to the number of epochs when training on an image from the DIV2K dataset for different neural network structures and/or initialization; Figure~\ref{fig:grid_image_regression_0078} presents the resulting images.}
  \label{fig:grid_image_regression_error_0078}
\end{figure}

\begin{figure}[!tbp]
  \begin{subfigure}[t]{0.49\textwidth}
	\centering
	\includegraphics[width=0.85\linewidth]{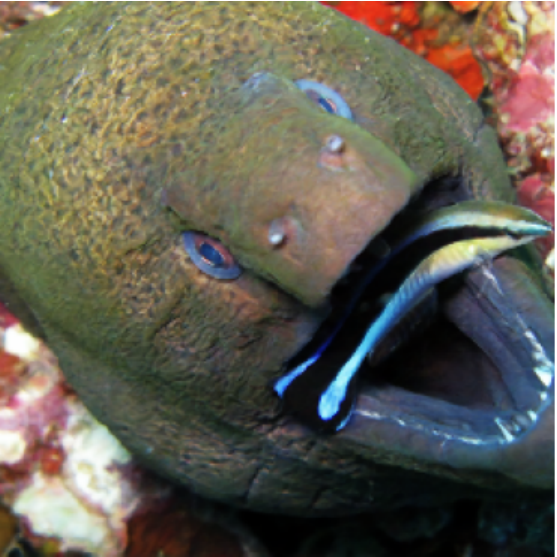}%
    \caption{RFF layer initialized using Algorithm~\ref{alg:AMRS_E} with three hidden ReLU layers.}
  \end{subfigure}
  \hfill
  \begin{subfigure}[t]{0.49\textwidth}
	\centering
	\includegraphics[width=0.85\linewidth]{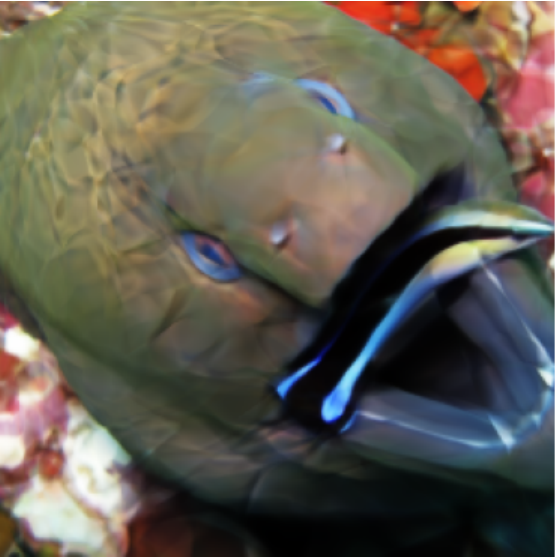}
    \caption{No RFF layer with four hidden ReLU layers.}
  \end{subfigure}
  \begin{subfigure}[b]{0.49\textwidth}
	\centering
	\includegraphics[width=0.85\linewidth]{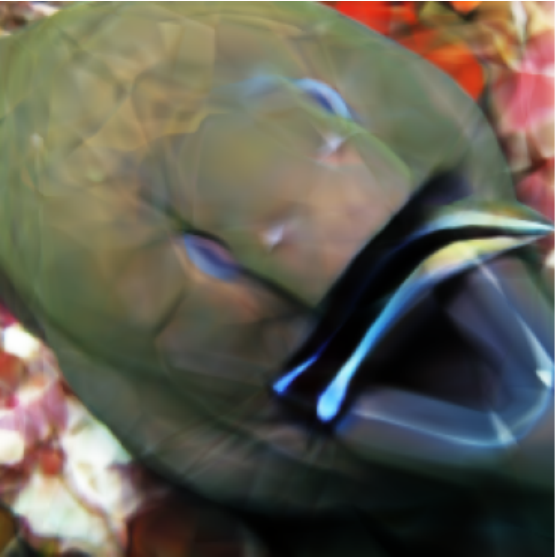}%
    \caption{No RFF layer with three hidden ReLU layers.}
  \end{subfigure}
  \hfill
  \begin{subfigure}[b]{0.49\textwidth}
	\centering
	\includegraphics[width=0.85\linewidth]{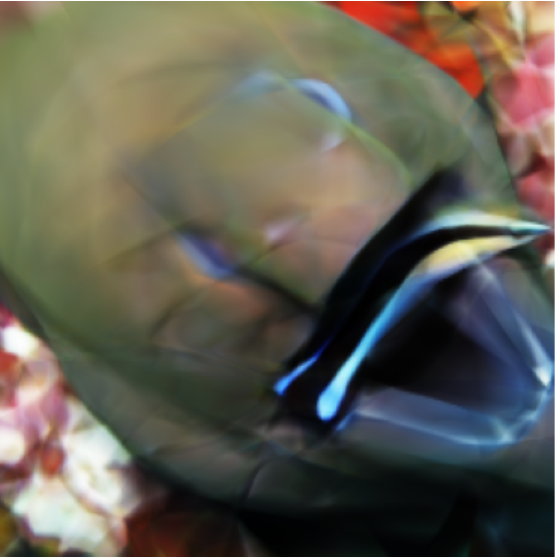}%
    \caption{RFF layer initialized using Xavier.\\~}
  \end{subfigure}
  \hfill
  \begin{center}
	\begin{subfigure}[b]{0.49\textwidth}
      \includegraphics[width=1.0\linewidth]{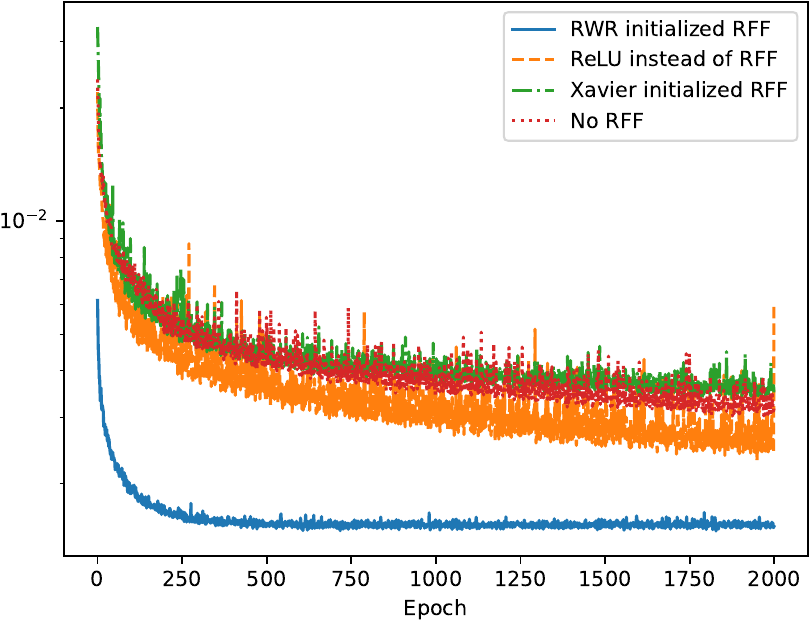}%
      \caption{Validation error with respect to epochs for various training algorithms}
    \end{subfigure}
  \end{center}
  \caption{Neural network representations of an image from the DIV2K dataset.}
  \label{fig:grid_image_regression_0098}
\end{figure}

\begin{figure}[!tbp]
  \begin{subfigure}[t]{0.49\textwidth}
	\centering
	\includegraphics[width=\linewidth]{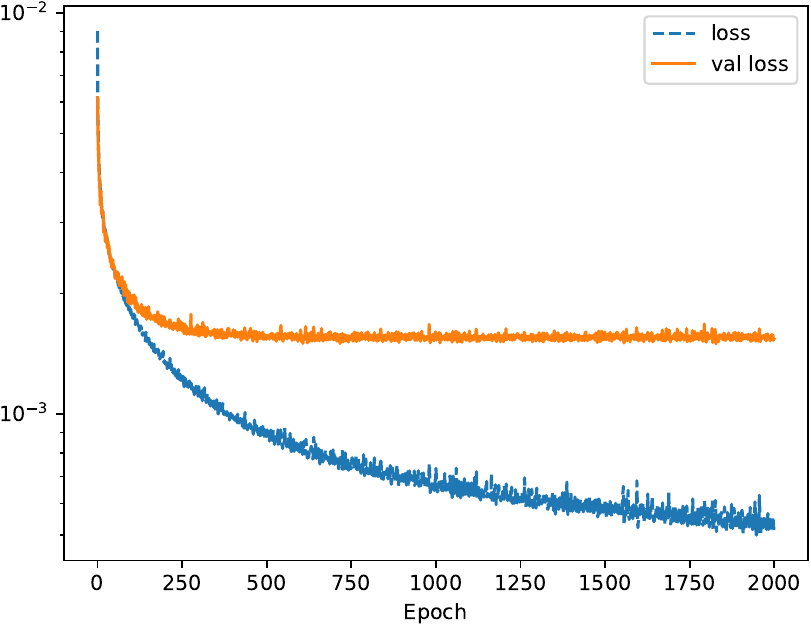}%
    \caption{RFF layer initialized using Algorithm~\ref{alg:AMRS_E} with three hidden ReLU layers.}
  \end{subfigure}
  \hfill
  \begin{subfigure}[t]{0.49\textwidth}
	\centering
	\includegraphics[width=\linewidth]{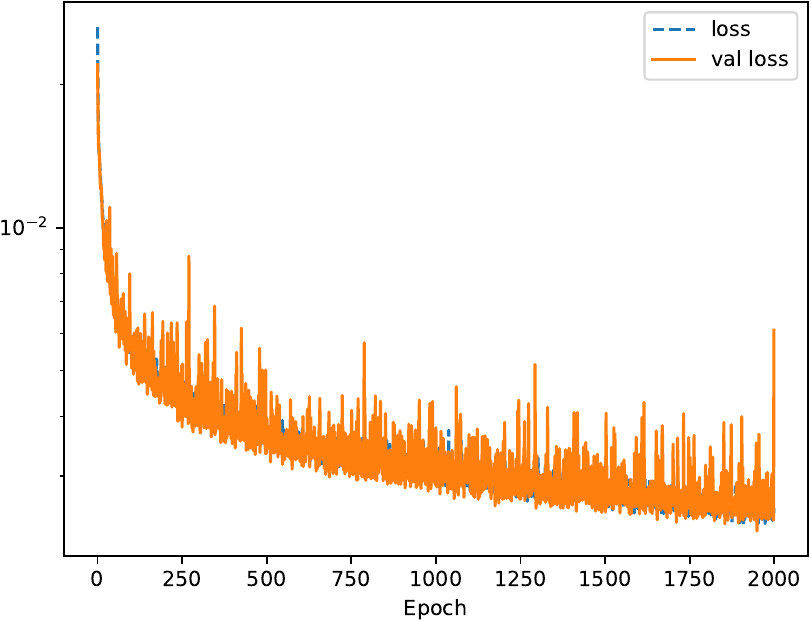}
    \caption{No RFF layer with four hidden ReLU layers.}
  \end{subfigure}
  \begin{subfigure}[b]{0.49\textwidth}
	\centering
	\includegraphics[width=\linewidth]{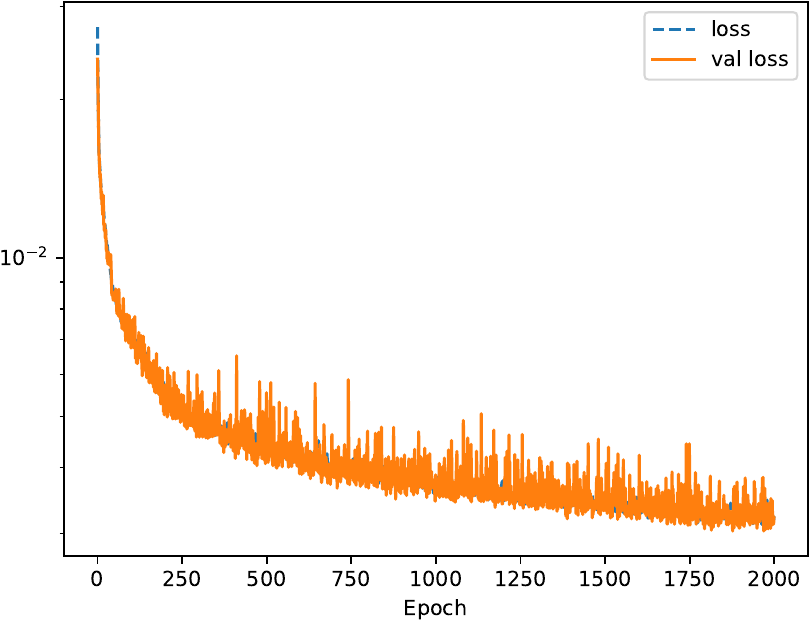}%
    \caption{No RFF layer with three hidden ReLU layers.}
  \end{subfigure}
  \hfill
  \begin{subfigure}[b]{0.49\textwidth}
	\centering
	\includegraphics[width=\linewidth]{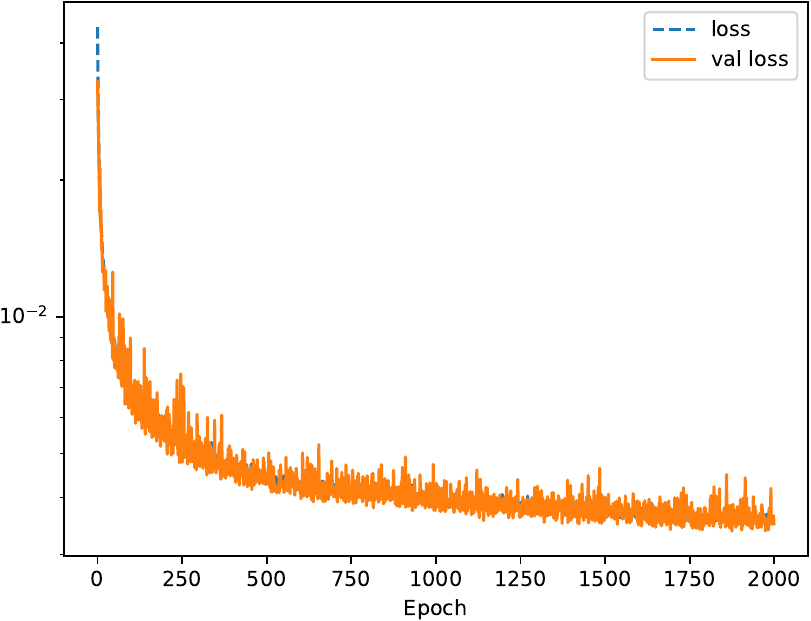}%
    \caption{RFF layer initialized using Xavier.\\~}
  \end{subfigure}
  \caption{Training and validation errors with respect to the number of epochs when training on an image from the DIV2K dataset for different neural network structures and/or initialization. Figure~\ref{fig:grid_image_regression_0098} presents the resulting images.}
  \label{fig:grid_image_regression_error_0098}
\end{figure}

\begin{figure}[!tbp]
  \begin{subfigure}[t]{0.49\textwidth}
	\centering
	\includegraphics[width=0.85\linewidth]{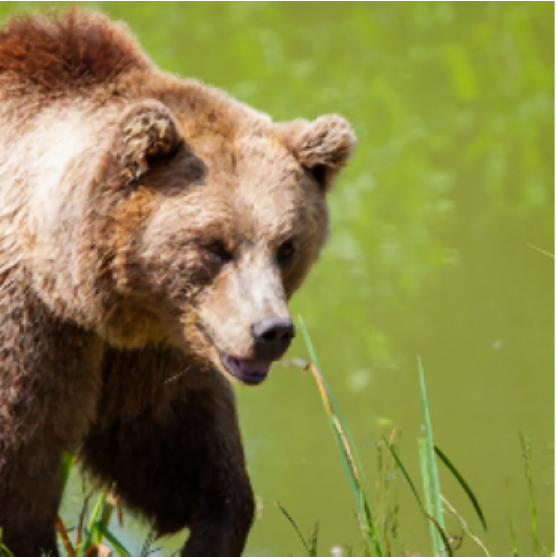}%
    \caption{RFF layer initialized using Algorithm~\ref{alg:AMRS_E} with three hidden ReLU layers.}
  \end{subfigure}
  \hfill
  \begin{subfigure}[t]{0.49\textwidth}
	\centering
	\includegraphics[width=0.85\linewidth]{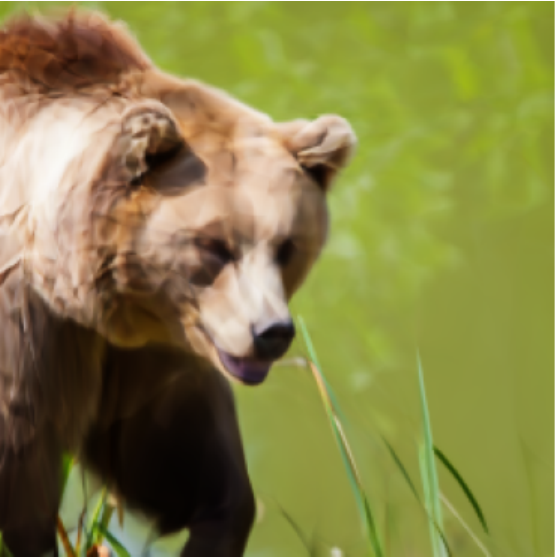}
    \caption{No RFF layer with four hidden ReLU layers.}
  \end{subfigure}
  \begin{subfigure}[b]{0.49\textwidth}
	\centering
	\includegraphics[width=0.85\linewidth]{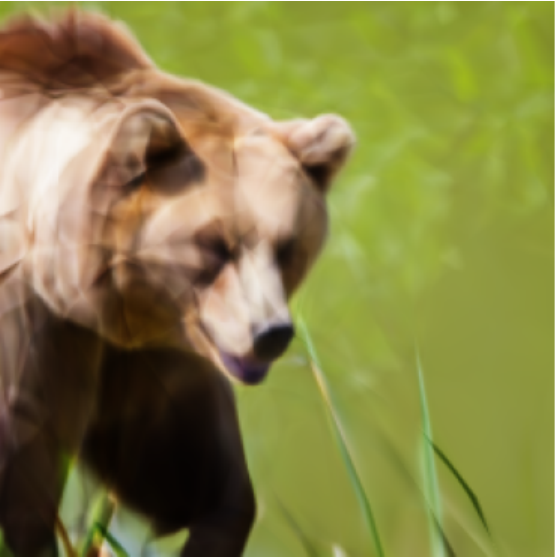}%
    \caption{No RFF layer with three hidden ReLU layers.}
  \end{subfigure}
  \hfill
  \begin{subfigure}[b]{0.49\textwidth}
	\centering
	\includegraphics[width=0.85\linewidth]{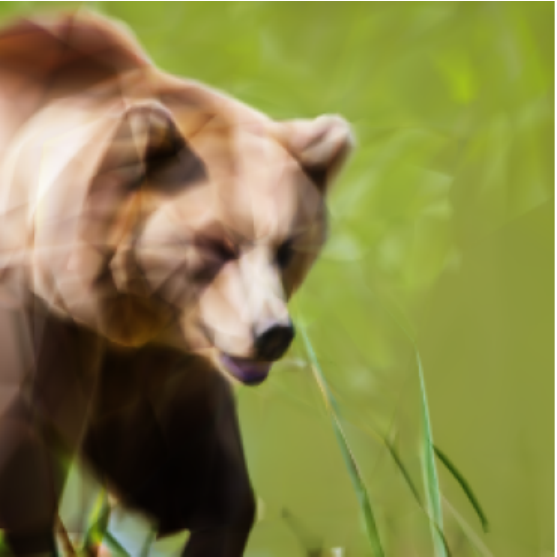}%
    \caption{RFF layer initialized using Xavier.\\~}
  \end{subfigure}
  \begin{center}
    \begin{subfigure}[b]{0.49\textwidth}
	  \includegraphics[width=1.0\linewidth]{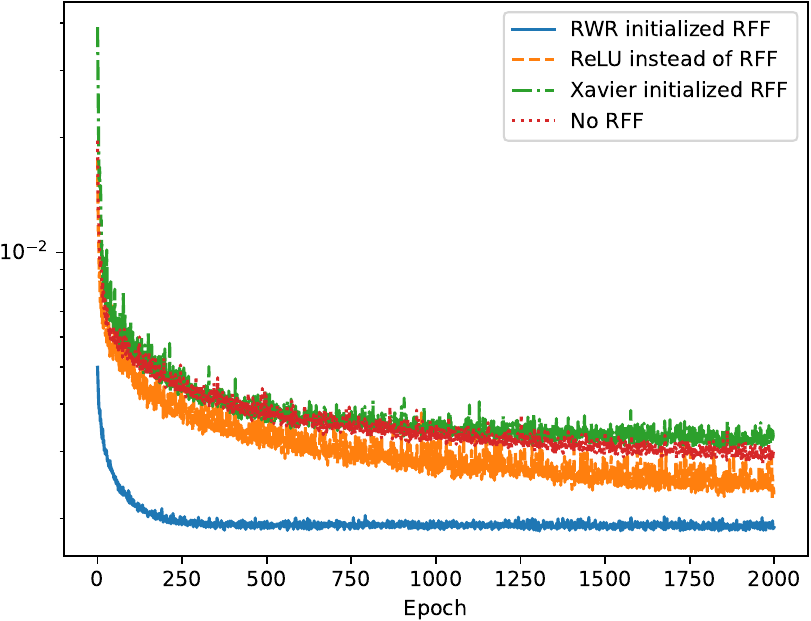}%
      \caption{Validation error with respect to epochs for various training algorithms}
    \end{subfigure}
  \end{center}
  \caption{Neural network representations of an image from the DIV2K dataset.}
  \label{fig:grid_image_regression_0336}
\end{figure}

\begin{figure}[!tbp]
  \begin{subfigure}[t]{0.49\textwidth}
	\includegraphics[width=\linewidth]{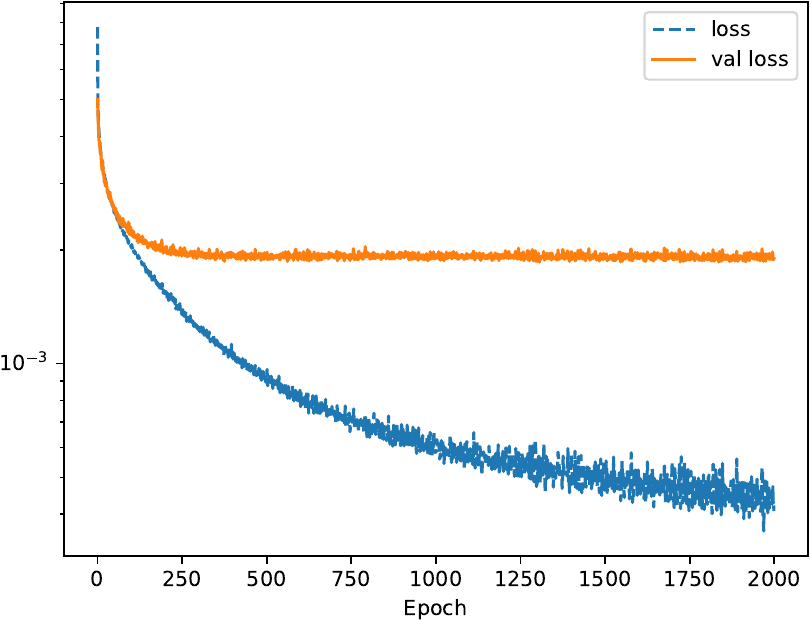}%
    \caption{RFF layer initialized using Algorithm~\ref{alg:AMRS_E} with three hidden ReLU layers.}
  \end{subfigure}
  \hfill
  \begin{subfigure}[t]{0.49\textwidth}
	\includegraphics[width=\linewidth]{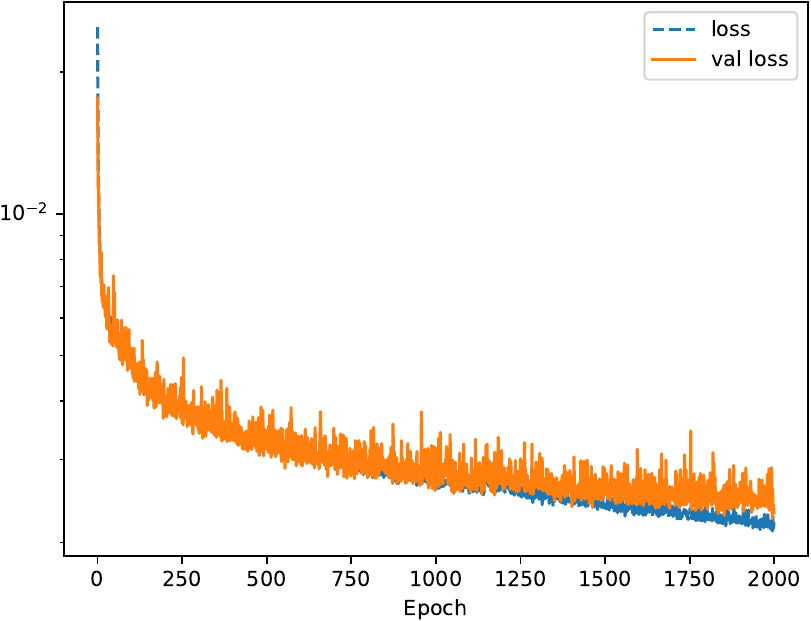}
    \caption{No RFF layer with four hidden ReLU layers.}
  \end{subfigure}
  \begin{subfigure}[b]{0.49\textwidth}
	\includegraphics[width=\linewidth]{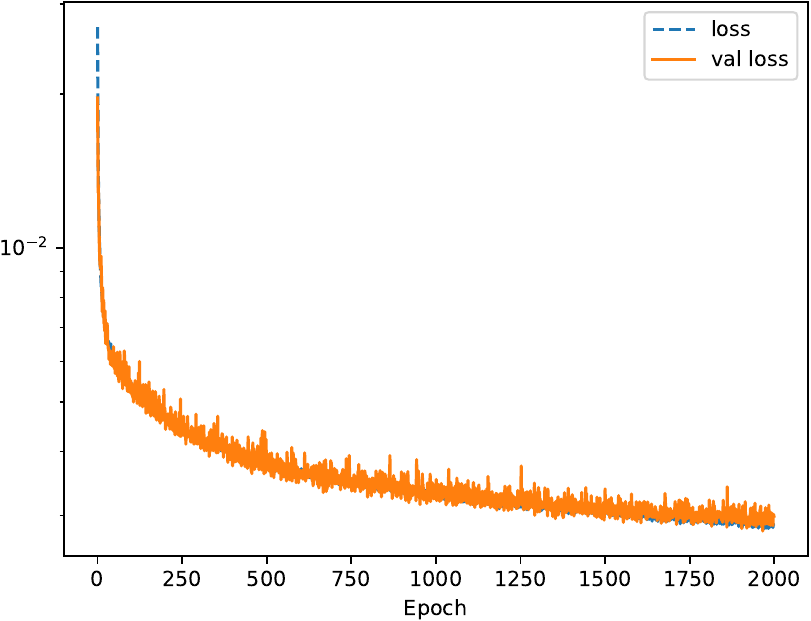}%
    \caption{No RFF layer with three hidden ReLU layers.}
  \end{subfigure}
  \hfill
  \begin{subfigure}[b]{0.49\textwidth}
  	\includegraphics[width=\linewidth]{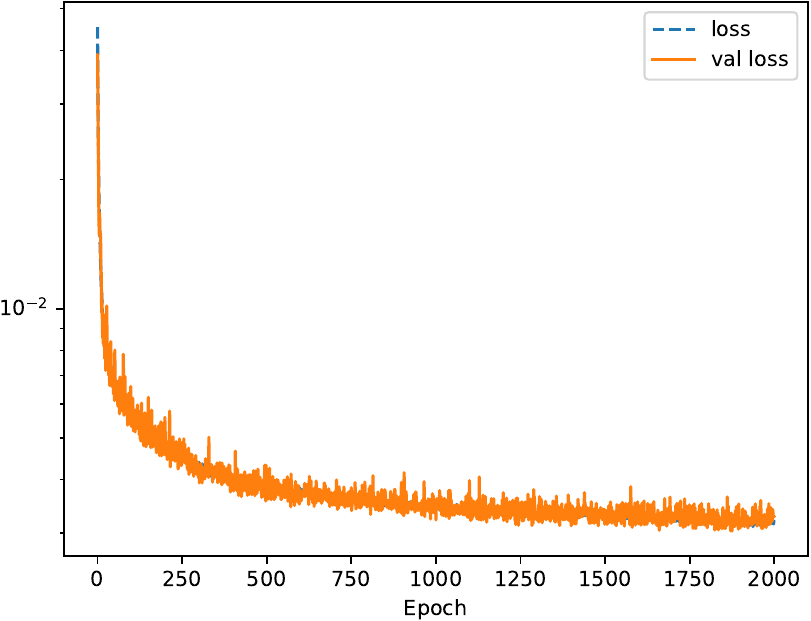}%
    \caption{RFF layer initialized using Xavier.\\~}
  \end{subfigure}
  \caption{Training and validation errors with respect to the number of epochs when training on an image from the DIV2K dataset for different neural network structures and/or initialization. Figure~\ref{fig:grid_image_regression_0336} presents the resulting images.}
  \label{fig:grid_image_regression_error_0336}
\end{figure}

Figures~\ref{fig:grid_image_regression_0756}, \ref{fig:grid_image_regression_0080}, \ref{fig:grid_image_regression_0746}, \ref{fig:grid_image_regression_0078}, \ref{fig:grid_image_regression_0098}, and~\ref{fig:grid_image_regression_0336} present images generated by evaluating the trained neural networks in the testing points. 
Figures~\ref{fig:grid_image_regression_error_0756}, \ref{fig:grid_image_regression_error_0080}, \ref{fig:grid_image_regression_error_0746}, \ref{fig:grid_image_regression_error_0078}, \ref{fig:grid_image_regression_error_0098}, and~\ref{fig:grid_image_regression_error_0336} depict the corresponding training and validation errors with respect to epochs during Adam optimizer training. Figures~\ref{fig:grid_image_regression_0078}, \ref{fig:grid_image_regression_0098}, and~\ref{fig:grid_image_regression_0336} display the validation errors for each approach in one figure. Sampling the frequencies and biases with Algorithm~\ref{alg:AMRS_E} converges the fastest during the Adam optimizer iterations of the compared approaches.

\begin{table}[]
	\centering
\begin{tabular}{|c|c|c|c|c|}
\hline
Approach & 1 & 2 & 3 & 4 \\ \hline
Mean                        & $25.49$  & $21.88$ & $22.53$ & $23.41$ \\ \hline
Standard deviation & $4.1$       & $3.58$   & $3.44$   & $3.39$ \\ \hline
Max                          & $37.34$  & $33.11$  & $33.83$ & $34.41$  \\ \hline
Min                          & $15.77$    & $10.09$ & $14.55$  & $15.19$ \\ \hline
\end{tabular}
\caption{The tabulated values demonstrate the impact of a Random Fourier Features (RFF) layer on the Peak Signal-to-Noise Ratio (PSNR) and these results are based on 92 randomly sampled images.
		The four approaches were the following:\\
		1 used one RFF layer initialized using Algorithm~\ref{alg:AMRS_E} and three hidden ReLU layers,\\
		2 used one RFF layer initialized using Glorot and three hidden ReLU layers,\\
		3 used no RFF layer but three hidden ReLU layers, and\\ 
		4 used no RFF layer but four hidden ReLU layers.\\
		In each approach, the last layer of the neural network was the Sigmoid activation function. 
}\label{tab:psnr_summary}
\end{table}


\section{Conclusions and Future Work}
We demonstrated that Algorithm~\ref{alg:AMRS_E} improves the stability of the ARFF algorithm with respect to the batch size, $M_B$, 
the initial frequency distribution, and the value of the parameter $\gamma$ in ARFF by occasionally resampling based on the probability mass 
$\mathbf{\check{p}}=\left\vert\BFBH\right\vert/\norm{\BFBH}_1$.
Algorithm~\ref{alg:AMRS_E} with resampling in every iteration can also omit the Metropolis step, reducing the computational cost per iteration, 
compared to ARFF, and entirely removing the hyper-parameter $\gamma$.
In a function regression test with a target function previously used to study ARFF, we experimentally demonstrated that Algorithm~\ref{alg:AMRS_E} 
with resampling typically reduces the training and testing errors faster than ARFF in the initial stage.
One example revealed that using Algorithm~\ref{alg:AMRS_E} as a pretrainer for the
Adam optimizer can reduce the cost compared to running it to a comparable error level without pretraining.
We also applied Algorithm~\ref{alg:AMRS_E} to sample frequencies for the RFF layer of 
coordinate-based MLPs in a simple image regression problem, automating a task that may 
otherwise be performed by manually tuning the parameters.

In the present work, we presented Algorithm~\ref{alg:AMRS_E} without convergence analysis. 
There exists an ongoing, more theoretical work where the convergence of Algorithm~\ref{alg:AMRS_E}
is analyzed in the case of a random walk with resampling in every iteration. 
%

\section*{Acknowledgement}
We thank Professor Anders Szepessy for his valued insights during research discussions and his much-appreciated feedback on earlier drafts of this paper. 
This publication is based upon work supported by the King Abdullah University of Science and Technology (KAUST) Office of Sponsored Research (OSR) under Award No. OSR-2019-CRG8-4033.
We acknowledge the financial support provided by the Alexander von Humboldt Foundation. 
\clearpage
\printbibliography
\end{document}